\documentclass[letterpaper,journal]{IEEEtran}
\usepackage{algorithmic}
\usepackage{algorithm}
\usepackage{array}
\usepackage[font=normalsize]{subfig}
\usepackage{textcomp}
\usepackage{stfloats}
\usepackage{url}
\usepackage{verbatim}
\usepackage{graphicx}
\usepackage[colorlinks=true, 
linkcolor = blue, 
citecolor = green,
urlcolor = blue]{hyperref}
\usepackage{orcidlink}
\usepackage{booktabs}
\usepackage{multirow}
\usepackage{makecell}
\usepackage{amsmath,mathtools,amssymb}
\usepackage{soul}
\usepackage{color}
\soulregister\ref7 
\soulregister\citealt7 

\newsavebox{\DiffTRtablebox}
\newcommand{\fitTable}[1]{%
  \sbox{\DiffTRtablebox}{#1}%
  \typeout{DiffTR table \thetable: natural=\the\wd\DiffTRtablebox,
    limit=\the\linewidth, font=\csname f@size\endcsname pt}%
  \ifdim\wd\DiffTRtablebox>\linewidth
    \resizebox{\linewidth}{!}{\usebox{\DiffTRtablebox}}%
  \else
    \usebox{\DiffTRtablebox}%
  \fi
}

\usepackage{microtype}
\usepackage{placeins}
\AtBeginEnvironment{figure*}{\captionsetup{font=small}}
\usepackage{cite}

\begin{document}

\title{Shared-Factor-Conditioned Diffusion for Low-Rank Tensor Recovery}
\author{
Zhengyun Cheng\orcidlink{0000-0002-1003-4381},
Guanwen Zhang\orcidlink{0000-0001-6036-4074},
and Wei Zhou\orcidlink{0000-0001-9715-6957}
\thanks{Corresponding authors: Guanwen Zhang and Wei Zhou.}
\thanks{Zhengyun Cheng, Guanwen Zhang, Wei Zhou are with the School of Electronics and Information, Northwestern Polytechnical University, Xi'an 710129, China. (e-mail: chengzy@mail.nwpu.edu.cn; guanwen.zh@nwpu.edu.cn; zhouwei@nwpu.edu.cn)}
}
\maketitle

\begin{abstract}
Tensor recovery spans heterogeneous data structures and observation processes, calling for a representation that can adapt across modalities and tasks. We propose DiffTR, an unsupervised framework that unifies tensor completion, denoising, and hyperspectral super-resolution through coordinate-conditioned diffusion. Shared factors indexed by tensor coordinates condition a compact score network, and score matching jointly learns the factors and an implicit tensor representation from the available observations. Scalar and vector-valued outputs accommodate individual entries and entire tubes, enabling a common formulation for tabular tensors, images, and videos. 
The framework separates representation learning from task-specific observation modeling: reverse diffusion recovers missing values; estimation via Tweedie's formula, soft thresholding, and optional spatial consistency support mixed-noise denoising; and measurement-guided sampling transfers a prior learned from low-resolution hyperspectral observations to high-resolution reconstruction, followed by post-sampling refinement with data consistency and spectral subspace regularization. This separation makes the representation compatible with different corruption models and permits task-specific constraints to be incorporated within the same recovery framework.
Experiments across multiple tensor modalities and recovery tasks demonstrate competitive reconstruction quality and the flexibility of a unified conditional generative representation. 
The code is available at \url{https://github.com/CZY-Code/DiffTR}
\end{abstract}


\section{Introduction}
Tensor recovery estimates an underlying tensor from incomplete, corrupted, or indirect measurements. This problem arises in sparse multiway tables, continuous-time records, RGB images, multispectral images, hyperspectral images, and videos. Applications span completion \cite{zhe2016distributed,fang2021streaming}, denoising, spectral super-resolution, factor analysis \cite{chen2022factor}, sequence modeling \cite{miller2021tensor}, model compression \cite{novikov2015tensorizing,dai2023deep}, and generative modeling \cite{glasser2019expressive,kuznetsov2019prior}. A general recovery framework must accommodate both heterogeneous structure and heterogeneous observations. Sparse tables encode interactions among discrete entities, whereas visual tensors exhibit spatial, spectral, and temporal correlations. Missing entries, missing tubes, mixed noise, and resolution loss impose different constraints on these structures. We use MSI and HSI to denote multispectral and hyperspectral images, respectively.

Classical low-rank tensor recovery combines a tensor representation, an observation model, and an optimization procedure. CP \cite{kolda2009tensor}, Tucker \cite{zhou2015efficient}, tensor train \cite{oseledets2011tensor}, tensor ring \cite{zhao2016tensor}, and tensor networks such as fully connected and tensor wheel decompositions \cite{zheng2021fully,wu2022tensor} prescribe how latent factors or cores are contracted into tensor entries. Structure search improves the representation by adapting network topology and mode assignments \cite{li2020evolutionary,li2022permutation}, but the resulting model still relies on an explicit contraction. Many methods based on implicit neural representations (INRs) extend tensor models with fixed contraction structures to continuous domains \cite{luo2023low,cheng2026cppruner}. Their effectiveness therefore depends on matching the contraction structure to the data, which can be challenging when spatial, spectral, temporal, and entity-level dependencies differ across modalities.

CoSTCo and POND address the restrictions of multilinear interactions through convolutional networks and Gaussian processes with neural kernels, respectively \cite{liu2019costco,tillinghast2020probabilistic}. Energy-based tensor models learn entry--factor compatibility without prescribed contractions, but primarily operate on scalar entries \cite{tao2023undirected,cheng2025score}. Extending these formulations to joint tube recovery and indirect measurements requires corresponding treatment of vector-valued outputs and observation constraints. Building on these advances, we seek a common conditional diffusion framework that accommodates heterogeneous tensor modalities, scalar entries and vector-valued tubes, and task-specific observation models across completion, denoising, and HSI super-resolution.

We introduce \emph{DiffTR}, a diffusion framework conditioned on coordinate-indexed tensor factors. For each mode, an $R$-dimensional factor vector is associated with every index or coordinate, either retrieved from a factor matrix for discrete modes or generated by an implicit neural representation for continuous modes. Atoms that share an index within the same mode reuse the corresponding factor vector, so that their observations jointly inform its learning. The resulting factor vectors are concatenated to condition a score network. Here, an atom refers to a scalar entry or a vector-valued tube, indexed by its positions across the modes. Denoising score matching jointly learns the factors and the conditional score, and reverse diffusion recovers the target atoms. We define the representation rank $R$ as the feature dimension of each factor vector. By learning interactions among factors, the score network provides a flexible alternative to prescribed multilinear contractions.

DiffTR adopts a unified coordinate-conditioned diffusion formulation for completion, denoising, and HSI super-resolution, with learning and recovery procedures tailored to the available observations. For completion, it learns from observed entries or tubes and generates the missing ones. For denoising, it alternates score learning, clean-value estimation via Tweedie's formula, and sparse-noise separation, with optional spatial consistency enforcement. For HSI super-resolution, it learns from the upsampled LR-HSI, guides reverse diffusion with HR-MSI measurements, and applies post-sampling refinement to improve data consistency. Each tensor is trained on its own observations, and the number of diffusion steps can be adjusted according to its structure and recovery requirements.

The main contributions are summarized as follows:
\begin{itemize}
	\item We formulate tensor recovery as conditional diffusion over scalar entries or vector-valued tubes, where shared mode-wise factors condition a score network that learns factor interactions.
	\item We develop an extensible, unsupervised framework that learns from each tensor's own observations and supports completion, denoising, and HSI super-resolution across tabular, image, and video tensors via task-specific observation models.
	\item We evaluate DiffTR across tensor modalities and recovery tasks, demonstrating competitive reconstruction quality and analyzing the effects of key components and computational cost.
\end{itemize}

\section{Related Work}
\subsection{Tensor Completion}
Tensor completion exploits dependencies among modes to infer missing values. Classical approaches use CP, Tucker, and tensor-network factorizations \cite{kolda2009tensor,zhao2016tensor,zheng2021fully}, unfolding-based trace norms \cite{liu2013tensorcompletion}, or tensor-tensor-product models \cite{zhou2018tensorcompletion}. Recent work combines low-rank Tucker factor priors with spatial--spectral graph regularization to capture global correlations and nonlinear manifold structure in HSI completion \cite{tian2026manifold}. Bayesian and nonparametric methods incorporate rank adaptation, nonlinear interactions, and uncertainty \cite{zhao2015bayesian,zhe2016distributed,tillinghast2020probabilistic,tillinghast2022nonparametric}, while continuous-time models capture temporal variation through splines or neural dynamics \cite{zhang2021dynamic,wang2022nonparametric,li2022decomposing}. Neural methods learn factor-to-value mappings, continuous tensor functions, frequency-aware representations, and deep latent parameterizations \cite{liu2019costco,luo2023low,zhou2025frequency,zhou2025dtr,cheng2026cppruner}. EnergyTD and ScoreTR further introduce implicit distributional modeling \cite{tao2023undirected,cheng2025score}.

Existing formulations typically target specific output units or coordinate domains, making it difficult to jointly support scalar entries, vector-valued tubes, and side information. Prescribed contractions can also restrict expressiveness when their structure poorly matches the dependencies in different tensor modalities. DiffTR addresses these limitations by supporting scalar and tube-valued atoms within a unified conditional diffusion formulation. Mode-wise factors are retrieved for discrete indices or generated by INRs for continuous coordinates, including timestamps. A conditional score network learns their interactions without prescribing a multilinear contraction.

\subsection{Tensor Denoising}
Tensor denoising combines structural priors with noise models. Tensor robust principal component analysis separates low-rank structure from sparse outliers \cite{lu2020trpca}, while Schatten-$p$ regularization provides a nonconvex low-rank prior \cite{fan2023euclideannorminduced}. A concave singular-value penalty combined with gradient tensors has been proposed to balance low-rankness and smoothness in sparse-noise removal \cite{yang2026singular}. Learned factorizations and continuous tensor functions provide flexible representations \cite{luo2022hlrtf,luo2023low}, while total variation (TV) and coordinate-Jacobian regularization promote spatial smoothness \cite{cheng2026cppruner}. Score matching offers a distributional approach by estimating gradients of noise-perturbed log densities \cite{vincent2011connection,song2019generative}.

The effectiveness of these methods depends on how well their structural priors and noise assumptions match the data. Mixed-noise recovery requires coordinated treatment of dense noise and sparse outliers, since fitting corrupted values directly can embed noise in the learned representation. DiffTR alternates score learning, clean-value estimation via Tweedie's formula, and sparse-noise separation by soft thresholding. An optional spatial-consistency loss penalizes score differences at nearby coordinates, incorporating local smoothness into the same conditional representation.

\subsection{Hyperspectral Image Super-Resolution}
HSI super-resolution combines the spectral information of an LR-HSI with the spatial information of an HR-MSI. Coupled tensor methods share latent structure between the two observations \cite{kanatsoulis2018hyperspectral,xu2024coupled}. Other approaches employ generalized tensor nuclear norms \cite{dian2024hyperspectral}, nonlinear transforms \cite{xu2025nonlinear}, or self-supervised decomposition networks \cite{chen2025learning}. Fang et al.\ estimate spatial coefficients in an LR-HSI-derived spectral subspace using a spatial-frequency network and data-fidelity refinement \cite{fang2026subspace}. Diffusion and score-based models offer generative priors \cite{ho2020denoising,song2021scorebased}, while diffusion posterior sampling uses measurements to guide reverse diffusion \cite{chung2023diffusion}.

Generative priors describe plausible HSI structure, while reconstruction must also remain consistent with both the LR-HSI and HR-MSI observations. Spatial blur and downsampling couple neighboring locations, requiring observation constraints that account for the full spatial and spectral degradation operators. DiffTR learns a conditional HSI score from the upsampled LR-HSI and uses HR-MSI measurements to guide reverse diffusion. Post-sampling refinement combines data consistency under both full observation operators with spectral subspace regularization, extending the common tensor recovery formulation to HSI super-resolution.

\section{Methodology}
\begin{figure*}[t]
	\centering
	\includegraphics[width=\textwidth]{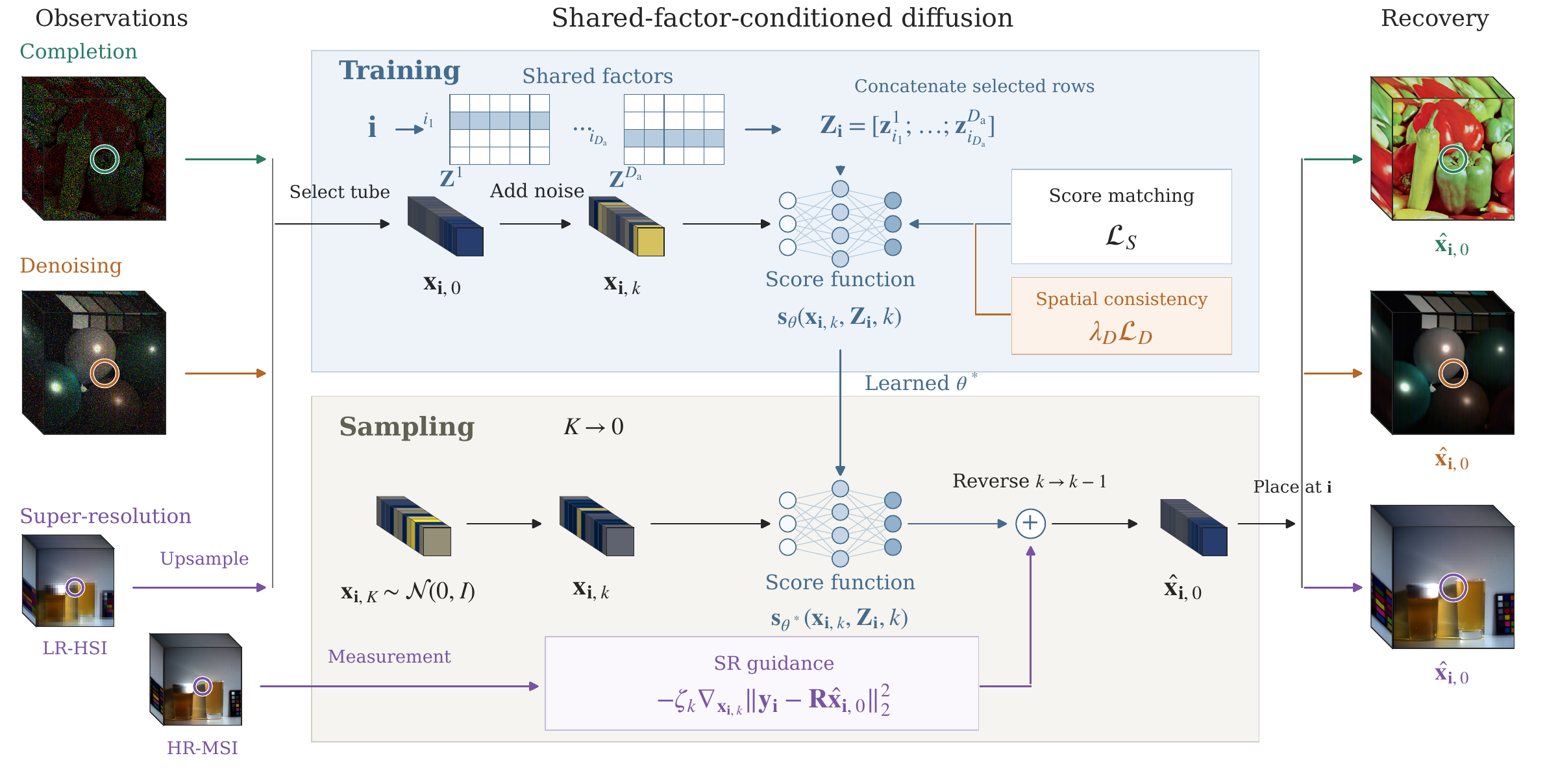}
	\caption{\textbf{Overview of the DiffTR framework}. Coordinate-indexed shared factors condition a score network for tensor completion, denoising, and HSI super-resolution. Green, orange, and purple denote completion, denoising constraints, and super-resolution guidance, respectively. Matched circles indicate the same spatial location.}
	\label{fig:difftr_overview_unified}
\end{figure*}

Figure~\ref{fig:difftr_overview_unified} illustrates the DiffTR framework, which uses coordinate-indexed shared factors to condition diffusion over scalar entries or vector-valued tubes. Task-specific learning and recovery procedures extend this common formulation to tensor completion, denoising, and HSI super-resolution.

Following standard tensor notation \cite{kolda2009tensor}, we denote scalars, vectors, matrices, and tensors by lowercase, bold lowercase, bold uppercase, and calligraphic uppercase letters, respectively. Let $\mathcal{X}\in\mathbb{R}^{I_1\times\cdots\times I_D}$ be a $D$th-order tensor. An atom $\mathbf{x}_{\mathbf{i}}\in\mathbb{R}^{C}$ is indexed by $\mathbf{i}=(i_1,\ldots,i_{D_{\mathrm a}})$, where $D_{\mathrm a}$ is the number of coordinate modes. Scalar entries have $D_{\mathrm a}=D$ and $C=1$; tubes have $D_{\mathrm a}=D-1$ and $C=I_D$, with the retained mode placed last by permutation if needed. The set of observed atom coordinates is $\Omega$, and $\bar{\Omega}$ is its complement.

\subsection{Preliminaries}
Denoising diffusion probabilistic models (DDPMs) \cite{ho2020denoising} define a fixed forward Markov chain that gradually perturbs a clean sample $\mathbf{x}_0$ with Gaussian noise:
\begin{equation}
q(\mathbf{x}_k\mid\mathbf{x}_{k-1})=\mathcal{N}\!\left(\mathbf{x}_k;\sqrt{\alpha_k}\mathbf{x}_{k-1},\beta_k\mathbf{I}\right),
\end{equation}
where $\alpha_k=1-\beta_k$, $k\in\{1,\ldots,K\}$, and $\{\beta_k\}_{k=1}^{K}$ is a prescribed variance schedule. With $\bar{\alpha}_k=\prod_{s=1}^{k}\alpha_s$, the marginal at any noise level is available in closed form:
\begin{equation}
\label{equ:forwardGuassian}
q(\mathbf{x}_k\mid\mathbf{x}_0)=\mathcal{N}\!\left(\mathbf{x}_k;\sqrt{\bar{\alpha}_k}\mathbf{x}_0,(1-\bar{\alpha}_k)\mathbf{I}\right).
\end{equation}
For a sufficiently long schedule, $q(\mathbf{x}_K)$ approaches a standard Gaussian. The reverse process is parameterized as
\begin{equation}
p_\theta(\mathbf{x}_{k-1}\mid\mathbf{x}_k)=\mathcal{N}\!\left(\mathbf{x}_{k-1};\boldsymbol{\mu}_\theta(\mathbf{x}_k,k),\tilde{\beta}_k\mathbf{I}\right),
\end{equation}
where $\tilde{\beta}_k$ is the reverse variance and
\begin{equation}
\boldsymbol{\mu}_\theta(\mathbf{x}_k,k)=\frac{1}{\sqrt{\alpha_k}}\left(\mathbf{x}_k+\beta_k\mathbf{s}_\theta(\mathbf{x}_k,k)\right).
\end{equation}
The network $\mathbf{s}_\theta$ approximates $\nabla_{\mathbf{x}_k}\log q_k(\mathbf{x}_k)$. Denoising score matching trains it against the tractable conditional score induced by Eq.~\eqref{equ:forwardGuassian} \cite{vincent2011connection,song2019generative}. Starting with $\mathbf{x}_K\sim\mathcal{N}(\mathbf{0},\mathbf{I})$, ancestral sampling applies
\begin{equation}
\label{equ:backwardSDE}
\mathbf{x}_{k-1}=\frac{1}{\sqrt{\alpha_k}}\left(\mathbf{x}_k+\beta_k\mathbf{s}_\theta(\mathbf{x}_k,k)\right)+\sqrt{\tilde{\beta}_k}\boldsymbol{\epsilon},
\end{equation}
where $\boldsymbol{\epsilon}\sim\mathcal{N}(\mathbf{0},\mathbf{I})$ for $k>1$ and is zero at the final step.

\subsection{Shared-Factor-Conditioned Diffusion}
\label{sec:shared_factor_diffusion}
Let $\mathcal{Y}$ be an incomplete, noisy, or degraded observation of the target tensor $\mathcal{X}$. For each coordinate mode $d\in\{1,\ldots,D_{\mathrm a}\}$, DiffTR uses a factor matrix $\mathbf{Z}^{d}\in\mathbb{R}^{I_d\times R}$. Throughout this paper, the representation rank $R$ denotes the feature dimension of a shared factor. For atom coordinate $\mathbf{i}$, the condition
\begin{equation}
\mathbf{Z}_{\mathbf{i}}=[\mathbf{z}^{1}_{i_1};\ldots;\mathbf{z}^{D_{\mathrm a}}_{i_{D_{\mathrm a}}}]\in\mathbb{R}^{D_{\mathrm a}R}
\end{equation}
concatenates the selected factor rows. Atoms sharing index $i_d$ use the same factor $\mathbf{z}^{d}_{i_d}$, coupling their representations across the tensor. The score network learns the mapping from this condition to atom values. For visual tensors, sinusoidal coordinate features \cite{tancik2020fourier} provide a continuous coordinate-based condition, written as $\mathbf{Z}_{\mathbf{i}}=\gamma_\phi(\mathbf{i})$. For sparse tabular tensors, the factor rows are optimized directly.

A conventional tensor decomposition makes $\mathcal{X}$ a deterministic contraction of its factors. DiffTR instead models the conditional distribution of each scalar or tube-valued atom. At diffusion level $k$,
\begin{equation}
\nabla_{\mathbf{x}_{\mathbf{i},k}}\log p_k(\mathbf{x}_{\mathbf{i},k}\mid\mathbf{Z}_{\mathbf{i}})=\nabla_{\mathbf{x}_{\mathbf{i},k}}\log p_k(\mathbf{x}_{\mathbf{i},k},\mathbf{Z}_{\mathbf{i}}),
\end{equation}
where the equality holds because $p(\mathbf{Z}_{\mathbf{i}})$ is constant with respect to $\mathbf{x}_{\mathbf{i},k}$. We approximate this conditional score with $\mathbf{s}_{\theta}(\mathbf{x}_{\mathbf{i},k},\mathbf{Z}_{\mathbf{i}},k)$.

For completion, training samples the observed atoms. For denoising, it uses the current clean-atom estimates obtained by the alternating procedure below; for super-resolution, it uses the upsampled LR-HSI. Conditioned on a training atom $\mathbf{x}_{\mathbf{i},0}$, the forward perturbation is independent of $\mathbf{Z}_{\mathbf{i}}$. Eq.~\eqref{equ:forwardGuassian} therefore provides the score-matching target
\begin{equation}
\nabla_{\mathbf{x}_{\mathbf{i},k}}\log q(\mathbf{x}_{\mathbf{i},k}\mid\mathbf{x}_{\mathbf{i},0})=-\frac{\mathbf{x}_{\mathbf{i},k}-\sqrt{\bar{\alpha}_k}\mathbf{x}_{\mathbf{i},0}}{1-\bar{\alpha}_k}.
\end{equation}
We jointly optimize the score network and shared factors using the score-matching loss $\mathcal{L}_S$:
\begin{equation}
\label{equ:DSMLoss}
\begin{aligned}
&\mathcal{L}_S(\theta,\mathcal{Z};\mathcal{X}_{\Omega}) =\frac{1}{2}\mathbb{E}_{\substack{\mathbf{i}\sim\Omega,\,k,\\\mathbf{x}_{\mathbf{i},k}\sim q(\cdot\mid\mathbf{x}_{\mathbf{i},0})}} \Big[\omega(k)\\ &\quad \big\|\mathbf{s}_{\theta}(\mathbf{x}_{\mathbf{i},k},\mathbf{Z}_{\mathbf{i}},k)-\nabla_{\mathbf{x}_{\mathbf{i},k}}\log q(\mathbf{x}_{\mathbf{i},k}\mid\mathbf{x}_{\mathbf{i},0})\big\|_2^2\Big].
\end{aligned}
\end{equation}
where $\omega(k)$ balances the noise levels. The network separately embeds the noisy atom, shared-factor condition, and diffusion index; concatenates the features; and maps them to an output with the atom dimension:
\begin{equation}
\mathbf{s}_{\theta}=g_{\theta_o}([g_{\theta_x}(\mathbf{x}_{\mathbf{i},k});g_{\theta_z}(\mathbf{Z}_{\mathbf{i}});g_{\theta_k}(k)]).
\end{equation}
Scalar atoms are used for entry-wise completion, whereas vector atoms are used for tube-wise completion and joint spectral recovery. Algorithm~\ref{algo:posteriorSampling} summarizes conditional reverse-diffusion sampling at the target coordinates.

\begin{algorithm}[t]
\caption{Reverse-diffusion sampling for tensor recovery.}
\label{algo:posteriorSampling}
\renewcommand{\algorithmicrequire}{\textbf{Input:}}
\renewcommand{\algorithmicensure}{\textbf{Output:}}
\begin{algorithmic}[1]
\REQUIRE Trained $\theta^*$ and $\mathcal{Z}$; schedule $\{\alpha_k,\tilde{\beta}_k\}_{k=1}^{K}$; target coordinates $\mathcal{I}$.
\FOR{$\mathbf{i}\in\mathcal{I}$}
\STATE Initialize $\mathbf{x}_{\mathbf{i},K}\sim\mathcal{N}(\mathbf{0},\mathbf{I})$.
\FOR{$k\gets K$ to $1$}
\STATE Draw $\boldsymbol{\epsilon}\sim\mathcal{N}(\mathbf{0},\mathbf{I})$ if $k>1$; otherwise set $\boldsymbol{\epsilon}=\mathbf{0}$.
\STATE $\mathbf{x}_{\mathbf{i},k-1}\gets\frac{1}{\sqrt{\alpha_k}}(\mathbf{x}_{\mathbf{i},k}+\beta_k\mathbf{s}_{\theta^*}(\mathbf{x}_{\mathbf{i},k},\mathbf{Z}_{\mathbf{i}},k))+\sqrt{\tilde{\beta}_k}\boldsymbol{\epsilon}$.
\ENDFOR
\ENDFOR
\STATE Fold the recovered atoms and retain observations where required by the task.
\ENSURE Recovered tensor $\hat{\mathcal{X}}$.
\end{algorithmic}
\end{algorithm}
For completion, $\mathcal{I}=\bar{\Omega}$ and sampled atoms are combined with the observations in $\Omega$. For denoising and super-resolution, the reverse process is modified as described next.

\subsection{Sparse-Noise Separation and Spatial Consistency}
\label{sec:denoising_method}
For denoising, DiffTR learns the tensor's implicit representation by score matching while separating corruption from the training targets. We write $\mathcal{Y}=\mathcal{X}+\mathcal{S}+\mathcal{E}$, where $\mathcal{S}$ represents sparse or structured outliers and $\mathcal{E}$ represents dense residual noise. Conditioned on the shared factors,
\begin{equation}
p_\theta(\mathcal{X},\mathcal{S}\mid\mathcal{Y},\mathcal{Z})\propto p(\mathcal{Y}\mid\mathcal{X},\mathcal{S})p(\mathcal{S})p_\theta(\mathcal{X}\mid\mathcal{Z}).
\end{equation}
A Gaussian data term models dense residual noise, and a Laplace prior $p(\mathcal{S})\propto\exp(-\lambda\|\mathcal{S}\|_1)$ promotes sparse corruption. We alternate clean-atom estimation, sparse-noise separation, and score-model updates. Tweedie's formula expresses the clean estimate in terms of the learned score:
\begin{equation}
\hat{\mathbf{x}}_{\mathbf{i},0}\approx\frac{1}{\sqrt{\bar{\alpha}_k}}\left(\mathbf{x}_{\mathbf{i},k}+(1-\bar{\alpha}_k)\mathbf{s}_{\theta}(\mathbf{x}_{\mathbf{i},k},\mathbf{Z}_{\mathbf{i}},k)\right),
\end{equation}
and the sparse update is
\begin{equation}
\label{eq:sparse_noise_update}
\mathcal{S}\leftarrow\operatorname{Soft}_{\lambda/2}(\mathcal{Y}-\hat{\mathcal{X}}),\quad \operatorname{Soft}_{\tau}(u)=\operatorname{sgn}(u)\max(|u|-\tau,0).
\end{equation}

Spatial consistency offers a coordinate-based alternative to TV regularization. For tensors with local spatial correlations, we perturb an atom's coordinate and penalize the resulting change in its conditional score. This optional regularizer introduces spatial smoothness through mini-batches while preserving the shared recovery formulation. The loss is
\begin{equation}
\label{equ:smoothReg}
\begin{aligned}
&\mathcal{L}_D(\theta,\phi) =\mathbb{E}_{\substack{\boldsymbol{\varepsilon}\sim\mathcal{N}(\mathbf{0},\sigma_c^2\mathbf{I}),\\k,\mathbf{i}}} \Big[\\ &\quad \big\|\mathbf{s}_\theta(\mathbf{x}_{\mathbf{i},k},\gamma_\phi(\mathbf{i}+\boldsymbol{\varepsilon}),k)-\mathbf{s}_\theta(\mathbf{x}_{\mathbf{i},k},\gamma_\phi(\mathbf{i}),k)\big\|_2^2\Big].
\end{aligned}
\end{equation}
where $\sigma_c$ sets the coordinate neighborhood. Gradients propagate through both score evaluations by default; Section~\ref{sec:denoise_component_ablation} compares this choice with a stop-gradient target. The denoising objective is $\mathcal{L}_S+\lambda_D\mathcal{L}_D$, with $\lambda_D$ controlling the spatial constraint and allowing it to be disabled for tasks that do not require local continuity. Algorithm~\ref{algo:tensorDenoising} summarizes the block coordinate descent (BCD) procedure and subsequent tensor recovery.

\begin{algorithm}[t]
\caption{Block coordinate descent for tensor denoising.}
\label{algo:tensorDenoising}
\renewcommand{\algorithmicrequire}{\textbf{Input:}}
\renewcommand{\algorithmicensure}{\textbf{Output:}}
\begin{algorithmic}[1]
\REQUIRE Noisy tensor $\mathcal{Y}$; $\lambda$; diffusion schedule; outer iterations $T$.
\STATE Initialize the model parameters.
\FOR{$t\gets1$ to $T$}
\STATE Sample $\mathbf{i}$, $k$, and Gaussian noise $\boldsymbol\epsilon$; perturb $\mathbf y_{\mathbf i}$ using Eq.~\eqref{equ:forwardGuassian} and predict $\hat{\mathbf{x}}^{(t)}_{\mathbf{i},0}$ without gradients.
\STATE $\mathbf{s}^{(t)}_{\mathbf{i}}\gets\operatorname{Soft}_{\lambda/2}(\mathbf{y}_{\mathbf{i}}-\hat{\mathbf{x}}^{(t)}_{\mathbf{i},0})$.
\STATE Set $\mathbf{x}^{(t)}_{\mathbf{i},0}=\mathbf{y}_{\mathbf{i}}-\mathbf{s}^{(t)}_{\mathbf{i}}$ and perturb it using the same $k$ and $\boldsymbol\epsilon$.
\STATE Take a gradient step on $\mathcal{L}_S+\lambda_D\mathcal{L}_D$ with the clean target fixed.
\ENDFOR
\STATE Recover $\hat{\mathcal{X}}$ by reverse diffusion and compute the total noise residual $\hat{\mathcal{N}}=\mathcal{Y}-\hat{\mathcal{X}}$.
\ENSURE Denoised tensor $\hat{\mathcal{X}}$ and total residual $\hat{\mathcal{N}}$ (sparse and dense noise).
\end{algorithmic}
\end{algorithm}

The same coordinate-conditioned network also supports likelihood-based fitting of a deterministic clean estimate. A Gaussian--outlier mixture accounts for clipping, with noise parameters learned from the observations. Optional proximal refinement balances the noise-scaled likelihood against a quadratic penalty on corrections within an estimated spectral subspace. A two-stage variant first fits a deterministic estimate, then trains diffusion on this fixed target and applies proximal guidance during denoising diffusion implicit model (DDIM) sampling. Section~\ref{sec:noise_aware_ablation} compares these choices; Appendix~\ref{app:noise_aware_optimization} gives their objectives and optimization steps.

\subsection{Likelihood-Guided Sampling for HSI Super-Resolution}
Let $\mathcal{X}\in\mathbb{R}^{I_H\times I_W\times I_C}$ be the target HR-HSI. The LR-HSI $\mathcal{X}_{\mathrm{LR}}\in\mathbb{R}^{I_h\times I_w\times I_C}$ preserves spectral resolution, whereas the HR-MSI $\mathcal{Y}\in\mathbb{R}^{I_H\times I_W\times I_c}$ preserves spatial resolution, with $I_h<I_H$, $I_w<I_W$, and $I_c<I_C$. We use
\begin{align}
\mathcal{X}_{\mathrm{LR}}&=\mathcal{X}\times_1\mathbf{P}_1\times_2\mathbf{P}_2+\mathcal{E}_{\mathcal{X}},\\
\mathcal{Y}&=\mathcal{X}\times_3\mathbf{R}+\mathcal{E}_{\mathcal{Y}},
\end{align}
where $\mathbf{P}_1$ and $\mathbf{P}_2$ model spatial degradation and $\mathbf{R}\in\mathbb{R}^{I_c\times I_C}$ is the spectral response. At spatial coordinate $\mathbf{i}$,
\begin{equation}
\label{equ:degradation}
\mathbf{y}_{\mathbf{i}}=\mathbf{R}\mathbf{x}_{\mathbf{i}}+\boldsymbol{\epsilon},
\end{equation}
with spectral signature $\mathbf{x}_{\mathbf{i}}\in\mathbb{R}^{I_C}$ treated as a tube-valued atom.

We spatially upsample $\mathcal{X}_{\mathrm{LR}}$ and use it to learn the conditional HSI score. During reconstruction, Bayes' rule gives
\begin{equation}
\begin{aligned}
&\nabla_{\mathbf{x}_{\mathbf{i},k}}\log p(\mathbf{x}_{\mathbf{i},k}\mid\mathbf{Z}_{\mathbf{i}},\mathbf{y}_{\mathbf{i}})\\ &\quad =\nabla_{\mathbf{x}_{\mathbf{i},k}}\log p(\mathbf{x}_{\mathbf{i},k}\mid\mathbf{Z}_{\mathbf{i}})+\nabla_{\mathbf{x}_{\mathbf{i},k}}\log p(\mathbf{y}_{\mathbf{i}}\mid\mathbf{x}_{\mathbf{i},k}).
\end{aligned}
\end{equation}
The first term is approximated by $\mathbf{s}_{\theta^*}$. Following diffusion posterior sampling \cite{chung2023diffusion}, the second is evaluated at the denoised estimate:
\begin{equation}
\label{equ:likelihoodAppro}
p(\mathbf{y}_{\mathbf{i}}\mid\mathbf{x}_{\mathbf{i},k})\approx p(\mathbf{y}_{\mathbf{i}}\mid\hat{\mathbf{x}}_{\mathbf{i},0}),
\end{equation}
where $\hat{\mathbf{x}}_{\mathbf{i},0}=\frac{1}{\sqrt{\bar{\alpha}_k}}\left(\mathbf{x}_{\mathbf{i},k}+(1-\bar{\alpha}_k)\mathbf{s}_{\theta^*}\right)$. Under Gaussian measurement noise,
\begin{equation}
\nabla_{\mathbf{x}_{\mathbf{i},k}}\log p(\mathbf{y}_{\mathbf{i}}\mid\hat{\mathbf{x}}_{\mathbf{i},0})\propto-\nabla_{\mathbf{x}_{\mathbf{i},k}}\|\mathbf{y}_{\mathbf{i}}-\mathbf{R}\hat{\mathbf{x}}_{\mathbf{i},0}\|_2^2.
\end{equation}
The LR-HSI supplies the learned prior, and the HR-MSI guides reverse diffusion through the pixelwise spectral residual in Algorithm~\ref{algo:superResolution}, producing an initial HR-HSI estimate $\widehat{\mathcal X}$. We use guidance weights $\zeta_k=g\bar\alpha_k$, where $g$ controls the guidance strength. The main experiments use DDPM sampling; Section~\ref{sec:sampling_depth} examines deterministic DDIM updates ($\eta=0$) with the same likelihood correction to vary the number of reverse steps at fixed model weights.

For observations with spatial blur, post-sampling refinement combines data consistency under both full observation operators with spectral subspace regularization. Let $\mathcal U\in\mathbb R^{I_H\times I_W\times I_C}$ denote the candidate HR-HSI. We obtain $\mathcal X^{+}=\arg\min_{\mathcal U}\mathcal J_{\tau,\gamma}(\mathcal U;\widehat{\mathcal X})$, where
\begin{equation}
\label{eq:sr_terminal_proximal}
\begin{split}
\mathcal J_{\tau,\gamma}(\mathcal U;\widehat{\mathcal X})={}&\tfrac12\|\mathcal U-\widehat{\mathcal X}\|_F^2\\
&+\tfrac{\tau}{2}\sum_{q\in\{L,M\}}\frac{\|A_q(\mathcal U)-\mathcal Y_q\|_F^2}{b_q}\\
&+\tfrac{\gamma}{2}\|\mathbf U_{\mathrm{sp}}(I-BB^{\mathsf T})\|_F^2.
\end{split}
\end{equation}
Here, $A_L$ and $A_M$ are the full LR-HSI and HR-MSI observation operators, including spatial degradation; $\mathcal Y_L=\mathcal X_{\mathrm{LR}}$ and $\mathcal Y_M=\mathcal Y$. The scalars $b_q>0$ normalize the operators. Each row of $\mathbf U_{\mathrm{sp}}\in\mathbb R^{I_HI_W\times I_C}$ is a pixel spectrum, and $B$ is an orthonormal spectral basis estimated from the LR-HSI. The three terms preserve the sampled estimate, fit the observations, and penalize spectral components outside this subspace, respectively. We solve this quadratic problem by conjugate gradients (CG) and then clip $\mathcal X^{+}$ to $[0,1]$. Appendix~\ref{app:sr_refinement} gives the derivation and solver settings.

\begin{algorithm}[t]
\caption{Overview of likelihood-guided diffusion for HSI super-resolution.}
\label{algo:superResolution}
\renewcommand{\algorithmicrequire}{\textbf{Input:}}
\renewcommand{\algorithmicensure}{\textbf{Output:}}
\begin{algorithmic}[1]
\REQUIRE Trained $\theta^*$ and $\mathcal{Z}$; HR-MSI $\mathcal{Y}$; $\mathbf{R}$; guidance weights $\{\zeta_k\}_{k=1}^{K}$.
\FOR{each HR spatial coordinate $\mathbf{i}$}
\STATE Initialize $\mathbf{x}_{\mathbf{i},K}\sim\mathcal{N}(\mathbf{0},\mathbf{I})$.
\FOR{$k\gets K$ to $1$}
\STATE Compute $\hat{\mathbf{x}}_{\mathbf{i},0}$ and take a reverse step conditioned on $\mathbf{Z}_{\mathbf{i}}$ to $\mathbf{x}'_{\mathbf{i},k-1}$.
\STATE $\mathbf{x}_{\mathbf{i},k-1}\gets\mathbf{x}'_{\mathbf{i},k-1}-\zeta_k\nabla_{\mathbf{x}_{\mathbf{i},k}}\|\mathbf{y}_{\mathbf{i}}-\mathbf{R}\hat{\mathbf{x}}_{\mathbf{i},0}\|_2^2$.
\ENDFOR
\ENDFOR
\STATE Fold $\{\mathbf{x}_{\mathbf{i},0}\}$ into $\hat{\mathcal{X}}$.
\ENSURE Initial HR-HSI estimate $\hat{\mathcal{X}}$.
\end{algorithmic}
\end{algorithm}

\section{Experiments}
We evaluate DiffTR on tabular completion, visual inpainting, mixed-noise denoising, and HSI super-resolution. The experiments measure recovery accuracy across modalities and examine how coordinate conditioning, task-specific observation models, and computational budget affect performance.

\subsection{Tensor Completion}

\begin{table*}[t]
	\centering
	\small
	\caption{\textbf{Quantitative comparison of sparse tensor completion}. Results are obtained by five-fold cross-validation on Alog and ACC. Lower RMSE and MAE indicate better recovery.}
	\label{tab:sparseTC}
	\fitTable{
	\begin{tabular}{c|cccc|cccc}
		\toprule
		\multicolumn{1}{c}{\textbf{Metric}} &\multicolumn{4}{c}{\textbf{RMSE}} & \multicolumn{4}{c}{\textbf{MAE}} \\
		\midrule
		\textbf{Method} &Rank 3 &Rank 5 &Rank 8 &Rank 10 &Rank 3 &Rank 5 &Rank 8 &Rank 10 \\
		\midrule
		\multicolumn{9}{c}{\textit{Alog} $(200\times 100\times 200)$}\\
		\midrule
		CP-WOPT \cite{bader2008efficient}
		&1.486 ± 0.282 &1.386 ± 0.043 &1.228 ± 0.063 &1.355 ± 0.079 
		&0.694 ± 0.098 &0.664 ± 0.018 &0.610 ± 0.027 &0.658 ± 0.026 \\
		GPTF \cite{zhe2016distributed}
		&0.911 ± 0.008 &0.867 ± 0.008 &0.878 ± 0.009 &0.884 ± 0.009 
		&0.511 ± 0.005 &0.494 ± 0.004 &0.530 ± 0.004 &0.554 ± 0.006 \\
		HGP-GPTF \cite{tillinghast2022nonparametric}
		&0.896 ± 0.011 &0.867 ± 0.009 &0.850 ± 0.011 &0.844 ± 0.006 
		&0.479 ± 0.007 &0.473 ± 0.003 &0.474 ± 0.004 &0.480 ± 0.004 \\
		POND \cite{tillinghast2020probabilistic}
		&0.885 ± 0.010 &0.871 ± 0.013 &0.858 ± 0.009 &0.857 ± 0.011 
		&0.463 ± 0.004 &0.454 ± 0.005 &0.444 ± 0.005 &0.443 ± 0.006\\
		CosTCo \cite{liu2019costco}
		&0.999 ± 0.007 &0.936 ± 0.017 &0.930 ± 0.024 &0.909 ± 0.014 
		&0.523 ± 0.006 &0.481 ± 0.007 &0.514 ± 0.031 &0.481 ± 0.008\\
		EnergyTD \cite{tao2023undirected}
		&0.864 ± 0.011 &0.835 ± 0.011 &0.840 ± 0.013 &0.833 ± 0.016 
		&0.450 ± 0.006 &0.433 ± 0.006 &0.424 ± 0.005 &0.409 ± 0.004 \\
		ScoreTR \cite{cheng2025score}
		&\underline{0.845 ± 0.013} &\underline{0.824 ± 0.008} &\underline{0.813 ± 0.015} &\underline{0.811 ± 0.012} 
		&\textbf{0.409 ± 0.010} &\underline{0.397 ± 0.008} &\underline{0.391 ± 0.007} &\underline{0.387 ± 0.005} \\
		\textbf{DiffTR}
		&\textbf{0.834 ± 0.008} &\textbf{0.803 ± 0.010} &\textbf{0.792 ± 0.009} &\textbf{0.785 ± 0.010} 
		&\underline{0.423 ± 0.004} &\textbf{0.395 ± 0.007} &\textbf{0.386 ± 0.006} &\textbf{0.369 ± 0.006} \\
		\midrule
		\multicolumn{9}{c}{\textit{Acc} $(3k\times 150\times 30k)$}\\
		\midrule
		CP-WOPT \cite{bader2008efficient}
		&0.533 ± 0.039 &0.592 ± 0.037 &0.603 ± 0.028 &0.589 ± 0.022 
		&0.138 ± 0.004 &0.147 ± 0.005 &0.148 ± 0.003 &0.147 ± 0.004\\
		GPTF \cite{zhe2016distributed}
		&0.367 ± 0.001 &0.357 ± 0.001 &0.359 ± 0.001 &0.368 ± 0.001 
		&0.152 ± 0.002 &0.150 ± 0.001 &0.167 ± 0.002 &0.182 ± 0.001 \\
		HGP-GPTF \cite{tillinghast2022nonparametric}
		&0.355 ± 0.001 &0.344 ± 0.001 &0.341 ± 0.001 &0.338 ± 0.001 
		&0.125 ± 0.003 &0.129 ± 0.001 &0.139 ± 0.000 &0.145 ± 0.002 \\
		CosTCo \cite{liu2019costco}
		&0.385 ± 0.003 &0.376 ± 0.018 &0.363 ± 0.004 &0.348 ± 0.002 
		&0.117 ± 0.004 &0.137 ± 0.020 &0.107 ± 0.004 &0.101 ± 0.004\\
		EnergyTD \cite{tao2023undirected}
		&0.348 ± 0.005 &0.336 ± 0.004 &0.328 ± 0.003 &0.328 ± 0.003 
		&0.110 ± 0.008 &0.101 ± 0.006 &0.094 ± 0.006 &0.101 ± 0.009\\
		ScoreTR \cite{cheng2025score}
		&\underline{0.325 ± 0.006} &\underline{0.319 ± 0.004} &\underline{0.322 ± 0.003} &\underline{0.320 ± 0.003} 
		&\underline{0.076 ± 0.010} &\underline{0.073 ± 0.006} &\underline{0.073 ± 0.007} &\underline{0.074 ± 0.006} \\
		\textbf{DiffTR}
		&\textbf{0.323 ± 0.003} &\textbf{0.314 ± 0.005} &\textbf{0.297 ± 0.004} &\textbf{0.292 ± 0.005}
		&\textbf{0.073 ± 0.009} &\textbf{0.071 ± 0.007} &\textbf{0.066 ± 0.007} &\textbf{0.065 ± 0.003} \\
		\bottomrule
	\end{tabular}}
\end{table*}

\begin{table*}[t]
	\centering
	\small
	\caption{\textbf{Quantitative comparison of continuous-time tensor completion}. Results are obtained by five-fold cross-validation on Air and Click. Lower RMSE and MAE indicate better recovery.}
	\label{tab:continuousTC}
	\fitTable{
	\begin{tabular}{c|cccc|cccc}
		\toprule
		\multicolumn{1}{c}{Metric} &\multicolumn{4}{c}{RMSE} & \multicolumn{4}{c}{MAE} \\
		\midrule
		Method &Rank 3 &Rank 5 &Rank 8 &Rank 10 &Rank 3 &Rank 5 &Rank 8 &Rank 10 \\
		\midrule
		\multicolumn{9}{c}{\textit{Air} $(12\times 6 \times T)$}\\
		\midrule
		CTCP \cite{zhang2021dynamic}
		&1.020 ± 0.002 &1.022 ± 0.002 &1.022 ± 0.002 &1.022 ± 0.002 
		&0.784 ± 0.002 &0.785 ± 0.002 &0.787 ± 0.002 &0.787 ± 0.002\\
		CTGP \cite{zhe2016distributed}
		&0.475 ± 0.000 &0.463 ± 0.000 &0.459 ± 0.000 &0.458 ± 0.000 
		&0.318 ± 0.000 &0.304 ± 0.000 &0.301 ± 0.000 &0.299 ± 0.000\\
		CTNN \cite{liu2019costco}
		&1.013 ± 0.001 &1.005 ± 0.005 &0.999 ± 0.009 &1.013 ± 0.002 
		&0.780 ± 0.001 &0.777 ± 0.003 &0.776 ± 0.003 &0.780 ± 0.001\\
		NNDTN \cite{wang2022nonparametric}
		&0.377 ± 0.004 &0.364 ± 0.002 &0.334 ± 0.004 &0.328 ± 0.004 
		&0.247 ± 0.003 &0.239 ± 0.002 &0.217 ± 0.003 &0.212 ± 0.004\\
		NONFAT \cite{wang2022nonparametric}
		&0.339 ± 0.002 &0.335 ± 0.002 &0.351 ± 0.005 &0.342 ± 0.002 
		&0.224 ± 0.002 &0.219 ± 0.001 &0.228 ± 0.003 &0.223 ± 0.001\\
		THIS-ODE \cite{li2022decomposing} 
		&0.569 ± 0.001 &0.566 ± 0.004 &0.542 ± 0.005 &0.541 ± 0.002 
		&0.415 ± 0.002 &0.409 ± 0.004 &0.395 ± 0.004 &0.391 ± 0.001\\
		EnergyTD \cite{tao2023undirected}
		&0.302 ± 0.008 &0.291 ± 0.006 &0.300 ± 0.012 &0.283 ± 0.004 
		&0.184 ± 0.006 &0.177 ± 0.003 &0.172 ± 0.006 &0.184 ± 0.003\\
		ScoreTR \cite{cheng2025score}
		&\underline{0.242 ± 0.003} &\underline{0.259 ± 0.005} &\underline{0.259 ± 0.009} &\underline{0.240 ± 0.003} 
		&\underline{0.156 ± 0.004} &\underline{0.170 ± 0.003} &\underline{0.168 ± 0.006} &\underline{0.154 ± 0.003}\\
		\textbf{DiffTR}
		&\textbf{0.235 ± 0.005} &\textbf{0.231 ± 0.004} &\textbf{0.230 ± 0.008} &\textbf{0.229 ± 0.002}
		&\textbf{0.145 ± 0.004} &\textbf{0.143 ± 0.005} &\textbf{0.143 ± 0.005} &\textbf{0.140 ± 0.004}\\
		\midrule
		\multicolumn{9}{c}{\textit{Click} $(7\times 2842\times 4127 \times T)$}\\
		\midrule
		CTCP \cite{zhang2021dynamic}
		&2.063 ± 0.009 &2.020 ± 0.025 &2.068 ± 0.012 &2.009 ± 0.023 
		&1.000 ± 0.009 &0.977 ± 0.021 &1.005 ± 0.010 &0.969 ± 0.012\\
		CTGP \cite{zhe2016distributed}
		&1.424 ± 0.002 &1.423 ± 0.004 &1.404 ± 0.004 &1.392 ± 0.002 
		&0.880 ± 0.003 &0.877 ± 0.003 &0.856 ± 0.002 &0.849 ± 0.001\\
		CTNN \cite{liu2019costco}
		&1.820 ± 0.005 &1.820 ± 0.005 &1.820 ± 0.005 &1.820 ± 0.005 
		&1.077 ± 0.027 &1.053 ± 0.012 &1.083 ± 0.016 &1.071 ± 0.024\\
		NNDTN \cite{wang2022nonparametric}
		&1.418 ± 0.005 &1.409 ± 0.004 &1.407 ± 0.002 &1.410 ± 0.004 
		&0.858 ± 0.002 &0.856 ± 0.002 &0.859 ± 0.003 &0.863 ± 0.006\\
		NONFAT \cite{wang2022nonparametric}
		&1.400 ± 0.008 &1.411 ± 0.006 &1.365 ± 0.004 &1.351 ± 0.002 
		&0.853 ± 0.004 &0.873 ± 0.004 &0.832 ± 0.004 &0.812 ± 0.002\\
		THIS-ODE \cite{li2022decomposing}
		&1.421 ± 0.004 &1.413 ± 0.002 &1.408 ± 0.002 &1.395 ± 0.003 
		&0.836 ± 0.004 &0.836 ± 0.003 &0.832 ± 0.002 &0.829 ± 0.003\\
		EnergyTD \cite{tao2023undirected}
		&1.396 ± 0.003 &1.385 ± 0.003 &1.356 ± 0.001 &1.357 ± 0.001 
		&0.777 ± 0.003 &0.775 ± 0.003 &0.772 ± 0.002 &0.773 ± 0.001\\
		ScoreTR \cite{cheng2025score}
		&\underline{1.374 ± 0.003} &\underline{1.368 ± 0.002} &\underline{1.355 ± 0.001} &\underline{1.346 ± 0.003} 
		&\underline{0.746 ± 0.005} &\underline{0.749 ± 0.004} &\underline{0.744 ± 0.004} &\underline{0.747 ± 0.003}\\
		\textbf{DiffTR}
		&\textbf{1.342 ± 0.005} &\textbf{1.338 ± 0.003} &\textbf{1.337 ± 0.002} &\textbf{1.340 ± 0.002}
		&\textbf{0.730 ± 0.003} &\textbf{0.723 ± 0.003} &\textbf{0.722 ± 0.005} &\textbf{0.723 ± 0.005} \\
		\bottomrule
	\end{tabular}}
\end{table*}

\subsubsection{Sparse Tensor Completion}
We evaluate DiffTR on two sparsely observed tensors \cite{zhe2015scalable}. \textit{Alog} is a file-access log with dimensions $200$ users $\times100$ actions $\times200$ resources and approximately $0.33\%$ nonzero entries. \textit{ACC} is constructed from a code-repository management system and has dimensions $3\mathrm{k}$ users $\times150$ actions $\times30\mathrm{k}$ resources, with approximately $0.009\%$ nonzero entries. We follow the split of \cite{tao2023undirected} and report five-fold cross-validation results.

The baselines cover deterministic, probabilistic, neural, and undirected generative tensor models: CP-WOPT \cite{bader2008efficient}, GPTF \cite{zhe2016distributed}, HGP-GPTF \cite{tillinghast2022nonparametric}, POND \cite{tillinghast2020probabilistic}, CoSTCo \cite{liu2019costco}, EnergyTD \cite{tao2023undirected}, and ScoreTR \cite{cheng2025score}. We evaluate factor dimensions $R\in\{3,5,8,10\}$ for all applicable methods. POND is omitted from ACC because of its reported computational and memory requirements in this setting.

Table~\ref{tab:sparseTC} shows that DiffTR achieves the lowest mean RMSE in all eight dataset--rank comparisons and the lowest mean MAE in seven; ScoreTR leads MAE on Alog at $R=3$. At $R=10$, DiffTR reduces RMSE relative to ScoreTR, the strongest baseline, by 3.2\% on Alog and 8.8\% on ACC. Its RMSE decreases as $R$ increases on both datasets, showing that the learned factor interactions effectively use additional representation capacity. These results demonstrate the effectiveness of DiffTR's shared-factor-conditioned diffusion for recovering sparsely observed multiway data.
\subsubsection{Continuous-Time Tensor Completion}
We next consider two timestamped tensors. \textit{Air} is the Beijing air-quality dataset \cite{zhang2017cautionary}, containing approximately $10^4$ observations over $12$ sites and $6$ pollutants. \textit{Click} is an advertising click-through tensor \cite{wang2022nonparametric} with $7$ banner positions, $2842$ site domains, and $4127$ mobile applications, and contains approximately $5\times10^4$ timestamped observations. We follow the split of \cite{wang2022nonparametric} and report five-fold cross-validation results.
We compare DiffTR with CTCP \cite{zhang2021dynamic}, CTGP \cite{zhe2016distributed}, CTNN \cite{liu2019costco}, NNDTN and NONFAT \cite{wang2022nonparametric}, THIS-ODE \cite{li2022decomposing}, EnergyTD \cite{tao2023undirected}, and ScoreTR \cite{cheng2025score}. These baselines span spline-based factor trajectories, Gaussian processes, recurrent and ordinary-differential-equation dynamics, and undirected generative models.

Table~\ref{tab:continuousTC} shows that DiffTR achieves the lowest mean RMSE and MAE on both datasets at every tested factor dimension. On Air at $R=5$, it reduces RMSE and MAE relative to ScoreTR by 10.8\% and 15.9\%, respectively. Its RMSE remains within 0.229--0.235 on Air and 1.337--1.342 on Click, indicating limited sensitivity to the tested factor dimensions. Together with the discrete-tensor results, these gains demonstrate that DiffTR incorporates temporal coordinates within the same conditional representation while maintaining competitive accuracy with compact factors.
\subsection{Image Recovery}
For visual inpainting and denoising, we compare with representative explicit and implicit low-rank priors: M\textsuperscript{2}DMT \cite{fan2021multi}, LRTC-ENR \cite{fan2023euclideannorminduced}, HLRTF \cite{luo2022hlrtf}, DeepTensor \cite{saragadam2024deeptensor}, LRTFR \cite{luo2023low}, CNet \cite{wang2025tensor}, DTR \cite{zhou2025dtr}, FA-INR \cite{zhou2025frequency}, and ScoreTR \cite{cheng2025score}. The comparison spans multilinear, neural-factor, implicit-coordinate, and score-based representations.
\begin{table*}[!t]
\centering
\small
\caption{\textbf{Quantitative comparison of entry-wise visual inpainting}. Results are averaged over the listed scenes. Higher PSNR and SSIM and lower NRMSE indicate better recovery.}
\label{tab:imageInpainting}
\fitTable{
\begin{tabular}{cc|ccc|ccc|ccc|ccc|ccc}
\toprule
\multicolumn{2}{c}{Sampling rate} & \multicolumn{3}{c}{0.05} & \multicolumn{3}{c}{0.1} & \multicolumn{3}{c}{0.15} & \multicolumn{3}{c}{0.2} & \multicolumn{3}{c}{0.25}\\
\midrule
Data & Method & PSNR & SSIM & NRMSE & PSNR & SSIM & NRMSE & PSNR & SSIM & NRMSE & PSNR & SSIM & NRMSE & PSNR & SSIM & NRMSE\\
\midrule
\multirow{8}{*}{\makecell[c]{Color images\\Sailboat\\House\\Peppers\\Plane\\$(512\times 512\times 3)$}} & Observed & 4.597 & 0.016 & 0.976 & 4.846 & 0.023 & 0.949 & 5.095 & 0.030 & 0.922 & 5.358 & 0.038 & 0.895 & 5.621 & 0.046 & 0.868\\
& DeepTensor \cite{saragadam2024deeptensor} & 18.47 & 0.286 & 0.182 & 21.50 & 0.484 & 0.150 & 24.53 & 0.682 & 0.118 & 26.31 & 0.717 & 0.101 & 28.09 & 0.752 & 0.084\\
& LRTFR \cite{luo2023low} & 19.84 & 0.499 & 0.180 & 23.03 & 0.597 & 0.132 & 26.22 & 0.695 & 0.084 & 27.49 & 0.741 & 0.073 & 28.76 & 0.787 & 0.062\\
& CNet \cite{wang2025tensor} & 20.57 & 0.401 & 0.162 & 22.49 & 0.540 & 0.136 & 24.41 & 0.679 & 0.110 & 25.39 & 0.711 & 0.097 & 26.37 & 0.743 & 0.084\\
& DTR \cite{zhou2025dtr} & 21.10 & 0.540 & 0.145 & 23.90 & 0.650 & 0.115 & 25.90 & 0.730 & 0.091 & 27.10 & 0.780 & 0.077 & 28.10 & 0.810 & 0.068\\
& FA-INR \cite{zhou2025frequency} & 22.20 & 0.600 & 0.127 & 24.70 & 0.700 & 0.103 & 26.70 & 0.770 & 0.082 & 27.80 & 0.810 & 0.071 & 28.70 & 0.830 & 0.063\\
& ScoreTR \cite{cheng2025score} & \underline{23.46} & \underline{0.706} & \underline{0.109} & \underline{25.34} & \underline{0.755} & \underline{0.092} & \underline{27.22} & \underline{0.804} & \underline{0.075} & \underline{28.14} & \underline{0.829} & \underline{0.068} & \underline{29.06} & \underline{0.845} & \underline{0.061}\\
& \textbf{DiffTR} & \textbf{24.74} & \textbf{0.715} & \textbf{0.099} & \textbf{26.92} & \textbf{0.779} & \textbf{0.077} & \textbf{28.19} & \textbf{0.812} & \textbf{0.067} & \textbf{28.99} & \textbf{0.832} & \textbf{0.062} & \textbf{29.61} & \textbf{0.848} & \textbf{0.057}\\
\midrule
\multirow{8}{*}{\makecell[c]{MSIs\\Beers\\Clay\\Flowers\\Balloons\\$(256\times 256\times 31)$}} & Observed & 13.71 & 0.354 & 0.976 & 13.96 & 0.386 & 0.949 & 14.21 & 0.418 & 0.922 & 14.47 & 0.447 & 0.894 & 14.73 & 0.476 & 0.866\\
& DeepTensor \cite{saragadam2024deeptensor} & 36.81 & 0.943 & 0.099 & 38.40 & 0.947 & 0.088 & 39.99 & 0.951 & 0.077 & 41.20 & 0.965 & 0.066 & 42.41 & 0.979 & 0.055\\
& LRTFR \cite{luo2023low} & 39.35 & 0.921 & 0.068 & 42.23 & 0.947 & 0.047 & 43.40 & 0.954 & 0.041 & 44.42 & 0.957 & 0.036 & 45.14 & 0.961 & 0.034\\
& CNet \cite{wang2025tensor} & 34.00 & 0.928 & 0.106 & 36.32 & 0.935 & 0.091 & 38.64 & 0.942 & 0.076 & 40.19 & 0.955 & 0.067 & 41.74 & 0.968 & 0.058\\
& DTR \cite{zhou2025dtr} & 38.50 & 0.935 & 0.075 & 41.60 & 0.958 & 0.054 & 43.30 & 0.968 & 0.043 & 44.50 & 0.974 & 0.036 & 45.40 & 0.978 & 0.032\\
& FA-INR \cite{zhou2025frequency} & 40.87 & 0.952 & 0.054 & 44.17 & 0.967 & 0.036 & \underline{46.08} & 0.973 & \underline{0.028} & \underline{47.09} & 0.977 & \underline{0.024} & 47.81 & 0.979 & 0.022\\
& ScoreTR \cite{cheng2025score} & \underline{42.30} & \underline{0.960} & \underline{0.047} & \underline{44.72} & \underline{0.975} & \underline{0.032} & 44.22 & \underline{0.990} & 0.030 & 46.28 & \underline{0.993} & \underline{0.024} & \underline{48.34} & \underline{0.995} & \underline{0.021}\\
& \textbf{DiffTR} & \textbf{42.83} & \textbf{0.986} & \textbf{0.044} & \textbf{46.85} & \textbf{0.992} & \textbf{0.027} & \textbf{48.72} & \textbf{0.994} & \textbf{0.021} & \textbf{49.78} & \textbf{0.995} & \textbf{0.018} & \textbf{50.62} & \textbf{0.996} & \textbf{0.016}\\
\midrule
\multirow{8}{*}{\makecell[c]{Videos\\Foreman\\News\\Tempete\\Paris\\$(288\times 352\times 3\times 10)$}} & Observed & 5.299 & 0.010 & 0.976 & 5.548 & 0.017 & 0.949 & 5.797 & 0.024 & 0.922 & 6.059 & 0.031 & 0.894 & 6.321 & 0.038 & 0.866\\
& DeepTensor \cite{saragadam2024deeptensor} & 24.00 & 0.771 & 0.148 & 25.67 & 0.813 & 0.114 & 27.34 & 0.855 & 0.080 & 28.89 & 0.851 & 0.075 & 30.44 & 0.852 & 0.069\\
& LRTFR \cite{luo2023low} & 25.80 & 0.802 & 0.134 & 28.53 & 0.828 & 0.103 & 29.36 & 0.854 & 0.087 & 29.77 & 0.866 & 0.077 & 30.18 & 0.878 & 0.070\\
& CNet \cite{wang2025tensor} & 22.83 & 0.706 & 0.135 & 24.66 & 0.768 & 0.104 & 26.49 & 0.830 & 0.085 & 28.10 & 0.837 & 0.071 & 29.71 & 0.844 & 0.068\\
& DTR \cite{zhou2025dtr} & 24.80 & 0.790 & 0.135 & 27.80 & 0.850 & 0.105 & 29.40 & 0.880 & 0.087 & 30.40 & 0.900 & 0.077 & 31.20 & 0.914 & 0.070\\
& FA-INR \cite{zhou2025frequency} & 25.20 & 0.800 & 0.132 & 28.20 & 0.860 & 0.102 & 29.80 & 0.890 & 0.084 & 30.80 & 0.908 & 0.074 & 31.60 & 0.920 & 0.067\\
& ScoreTR \cite{cheng2025score} & \underline{26.17} & \underline{0.820} & \underline{0.128} & \underline{28.88} & \underline{0.875} & \underline{0.097} & \underline{30.56} & \underline{0.901} & \underline{0.079} & \underline{31.52} & \underline{0.914} & \underline{0.070} & \underline{32.26} & \underline{0.923} & \underline{0.065}\\
& \textbf{DiffTR} & \textbf{26.67} & \textbf{0.829} & \textbf{0.124} & \textbf{29.38} & \textbf{0.886} & \textbf{0.093} & \textbf{31.01} & \textbf{0.910} & \textbf{0.078} & \textbf{31.97} & \textbf{0.922} & \textbf{0.069} & \textbf{32.66} & \textbf{0.931} & \textbf{0.064}\\
\bottomrule
\end{tabular}}
\end{table*}

\begin{table*}[!h]
\centering
\small
\caption{\textbf{Quantitative comparison of tube-wise visual inpainting}. Results are averaged over the listed scenes. Bold and underlined values indicate the best and second-best results, respectively, including ties.}
\label{tab:imageInpainting2}
\fitTable{
\begin{tabular}{cc|ccc|ccc|ccc|ccc|ccc}
\toprule
\multicolumn{2}{c}{Sampling rate} & \multicolumn{3}{c}{0.05} & \multicolumn{3}{c}{0.1} & \multicolumn{3}{c}{0.15} & \multicolumn{3}{c}{0.2} & \multicolumn{3}{c}{0.25}\\
\midrule
Data & Method & PSNR & SSIM & NRMSE & PSNR & SSIM & NRMSE & PSNR & SSIM & NRMSE & PSNR & SSIM & NRMSE & PSNR & SSIM & NRMSE\\
\midrule
\multirow{8}{*}{\makecell[c]{Color images\\Sailboat\\House\\Peppers\\Plane\\$(512\times 512\times 3)$}}  &  Observed  & 4.613 & 0.015 & 0.975 & 4.848 & 0.023 & 0.949 & 5.095 & 0.030 & 0.922 & 5.360 & 0.038 & 0.894 & 5.641 & 0.046 & 0.866\\
 &  DeepTensor \cite{saragadam2024deeptensor}  & 16.60 & 0.350 & 0.250 & 20.60 & 0.470 & 0.159 & 23.70 & 0.670 & 0.111 & 25.70 & 0.710 & 0.090 & 27.10 & 0.750 & 0.076\\
 &  LRTFR \cite{luo2023low}  & 18.30 & 0.470 & 0.205 & 22.20 & 0.580 & 0.133 & 25.40 & 0.690 & 0.092 & 26.80 & 0.740 & 0.078 & 28.00 & 0.770 & 0.069\\
 &  CNet \cite{wang2025tensor}  & 18.70 & 0.420 & 0.196 & 21.60 & 0.530 & 0.142 & 23.60 & 0.670 & 0.113 & 24.70 & 0.710 & 0.100 & 25.80 & 0.740 & 0.089\\
 &  DTR \cite{zhou2025dtr}  & 19.60 & 0.560 & 0.177 & 23.20 & 0.680 & 0.117 & 25.30 & 0.740 & 0.093 & 26.70 & 0.790 & 0.079 & 27.90 & 0.820 & 0.070\\
 &  FA-INR \cite{zhou2025frequency}  & 20.30 & 0.590 & 0.163 & 23.90 & 0.710 & 0.108 & 25.90 & 0.760 & 0.087 & 27.10 & 0.800 & 0.076 & 28.20 & 0.830 & 0.067\\
 &  ScoreTR \cite{cheng2025score}  & \underline{21.70} & \underline{0.640} & \underline{0.139} & \underline{24.50} & \underline{0.740} & \underline{0.102} & \underline{26.40} & \underline{0.790} & \underline{0.082} & \underline{27.50} & \underline{0.820} & \underline{0.073} & \underline{28.50} & \underline{0.840} & \underline{0.065}\\
 &  \textbf{DiffTR}  & \textbf{23.82} & \textbf{0.688} & \textbf{0.109} & \textbf{26.05} & \textbf{0.765} & \textbf{0.085} & \textbf{27.38} & \textbf{0.802} & \textbf{0.073} & \textbf{28.34} & \textbf{0.827} & \textbf{0.066} & \textbf{29.20} & \textbf{0.848} & \textbf{0.060}\\
\midrule
\multirow{8}{*}{\makecell[c]{MSIs\\Beers\\Clay\\Flowers\\Balloons\\$(256\times 256\times 31)$}}  &  Observed  & 14.18 & 0.259 & 0.975 & 14.41 & 0.286 & 0.949 & 14.66 & 0.310 & 0.922 & 14.92 & 0.333 & 0.894 & 15.20 & 0.354 & 0.867\\
 &  DeepTensor \cite{saragadam2024deeptensor}  & 24.30 & 0.880 & 0.341 & 27.40 & 0.910 & 0.246 & 29.40 & 0.930 & 0.199 & 31.40 & 0.950 & 0.161 & 32.80 & 0.960 & 0.139\\
 &  LRTFR \cite{luo2023low}  & 28.30 & 0.880 & 0.215 & 31.20 & 0.910 & 0.158 & 32.80 & 0.930 & 0.135 & 34.60 & 0.940 & 0.111 & 35.80 & 0.950 & 0.099\\
 &  CNet \cite{wang2025tensor}  & 21.80 & 0.860 & 0.454 & 25.30 & 0.900 & 0.313 & 28.10 & 0.920 & 0.233 & 30.30 & 0.940 & 0.181 & 31.80 & 0.950 & 0.156\\
 &  DTR \cite{zhou2025dtr}  & 28.60 & 0.810 & 0.208 & 32.50 & 0.870 & 0.137 & 35.00 & 0.910 & 0.104 & 37.00 & 0.940 & 0.084 & 38.40 & 0.950 & 0.073\\
 &  FA-INR \cite{zhou2025frequency}  & \underline{29.80} & \underline{0.900} & \underline{0.181} & \underline{33.10} & \underline{0.930} & \underline{0.127} & \underline{35.50} & \underline{0.950} & \underline{0.099} & \underline{37.20} & \underline{0.960} & \underline{0.082} & \underline{38.50} & \underline{0.970} & \underline{0.072}\\
 &  ScoreTR \cite{cheng2025score}  & 28.60 & 0.890 & 0.208 & 31.60 & \underline{0.930} & 0.151 & 33.60 & \underline{0.950} & 0.123 & 36.40 & \underline{0.960} & 0.090 & 38.00 & \underline{0.970} & 0.077\\
 &  \textbf{DiffTR}  & \textbf{32.60} & \textbf{0.931} & \textbf{0.131} & \textbf{35.81} & \textbf{0.959} & \textbf{0.093} & \textbf{38.13} & \textbf{0.972} & \textbf{0.073} & \textbf{39.94} & \textbf{0.979} & \textbf{0.060} & \textbf{41.22} & \textbf{0.983} & \textbf{0.053}\\
\midrule
\multirow{8}{*}{\makecell[c]{Videos\\Foreman\\News\\Tempete\\Paris\\$(288\times 352\times 3\times 10)$}}  &  Observed  & 7.231 & 0.022 & 0.975 & 7.466 & 0.039 & 0.949 & 7.715 & 0.056 & 0.922 & 7.978 & 0.073 & 0.894 & 8.259 & 0.091 & 0.866\\
 &  DeepTensor \cite{saragadam2024deeptensor}  & 20.60 & 0.680 & 0.242 & 23.60 & 0.780 & 0.178 & 25.20 & 0.830 & 0.150 & 26.90 & 0.830 & 0.124 & 27.80 & 0.850 & 0.112\\
 &  LRTFR \cite{luo2023low}  & 23.50 & 0.710 & 0.174 & 26.40 & 0.790 & 0.128 & 27.30 & 0.830 & 0.119 & 27.80 & 0.840 & 0.112 & 28.50 & 0.860 & 0.103\\
 &  CNet \cite{wang2025tensor}  & 19.60 & 0.640 & 0.272 & 22.60 & 0.730 & 0.200 & 24.40 & 0.800 & 0.165 & 26.10 & 0.810 & 0.136 & 27.10 & 0.830 & 0.121\\
 &  DTR \cite{zhou2025dtr}  & 22.80 & 0.710 & 0.188 & 25.80 & 0.800 & 0.138 & 27.60 & 0.840 & 0.114 & 28.90 & 0.870 & 0.099 & 29.80 & 0.890 & 0.089\\
 &  FA-INR \cite{zhou2025frequency}  & 23.20 & 0.740 & 0.180 & 26.20 & 0.820 & 0.132 & 27.90 & 0.860 & 0.110 & 29.10 & 0.880 & 0.096 & 30.00 & 0.890 & 0.087\\
 &  ScoreTR \cite{cheng2025score}  & \underline{24.10} & \underline{0.760} & \underline{0.162} & \underline{26.80} & \underline{0.840} & \underline{0.123} & \underline{28.50} & \underline{0.870} & \underline{0.103} & \underline{29.60} & \underline{0.890} & \underline{0.092} & \underline{30.40} & \underline{0.900} & \underline{0.083}\\
 &  \textbf{DiffTR}  & \textbf{24.56} & \textbf{0.782} & \textbf{0.153} & \textbf{27.27} & \textbf{0.852} & \textbf{0.116} & \textbf{28.92} & \textbf{0.883} & \textbf{0.098} & \textbf{30.02} & \textbf{0.899} & \textbf{0.087} & \textbf{30.78} & \textbf{0.910} & \textbf{0.079}\\
\bottomrule
\end{tabular}}
\end{table*}

\subsubsection{Multidimensional Visual Data Inpainting}
We consider three visual modalities: RGB images from the USC-SIPI database\footnote{\url{https://sipi.usc.edu/database/database.php}}, MSIs from CAVE\footnote{\label{foot:cave}\url{https://cave.cs.columbia.edu/repository/Multispectral}}, and videos from the Xiph collection\footnote{\url{https://media.xiph.org/video/derf/}}. Random entry-wise observations are evaluated at sampling rates of $5\%$, $10\%$, $15\%$, $20\%$, and $25\%$. We report PSNR, SSIM, and normalized RMSE (NRMSE), averaged over the listed scenes. We also evaluate tube-wise sampling to test whether vector-valued atoms can be recovered without changing the conditional diffusion formulation.

\begin{figure*}[!htbp]
	\centering
	\captionsetup[subfloat]{labelsep=none,format=plain,labelformat=empty,font=footnotesize}
	\begin{minipage}{1.\linewidth}
\centering
\subfloat[PSNR 5.609]
{\includegraphics[width=0.102\linewidth]{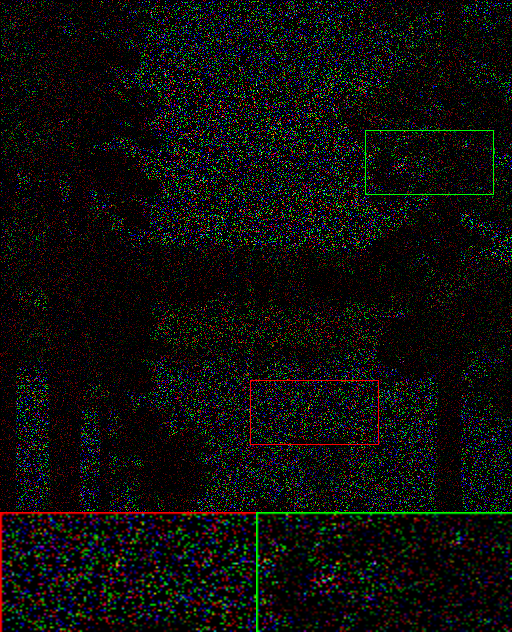}}
\hspace{0.01cm}
\subfloat[PSNR 19.61]
{\includegraphics[width=0.102\linewidth]{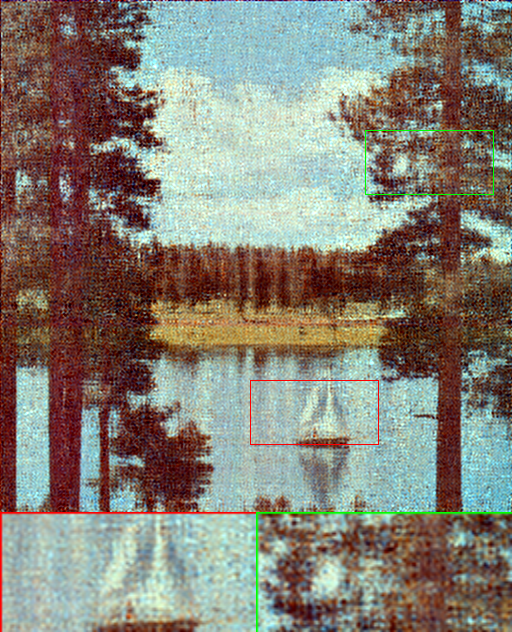}}
\hspace{0.01cm}
\subfloat[PSNR 20.40]
{\includegraphics[width=0.102\linewidth]{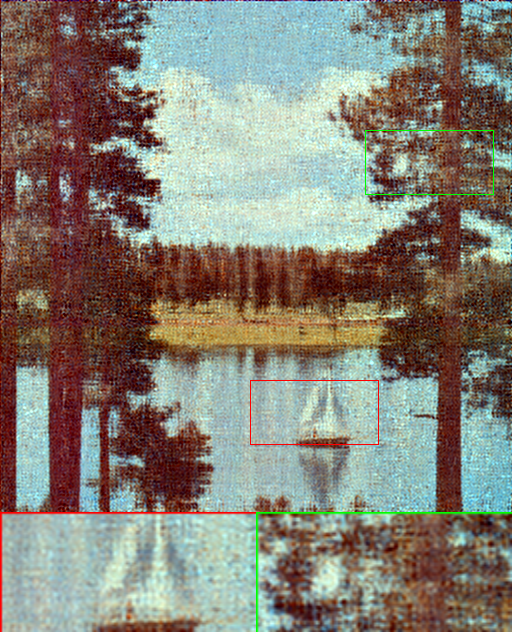}}
\hspace{0.01cm}
\subfloat[PSNR 21.21]
{\includegraphics[width=0.102\linewidth]{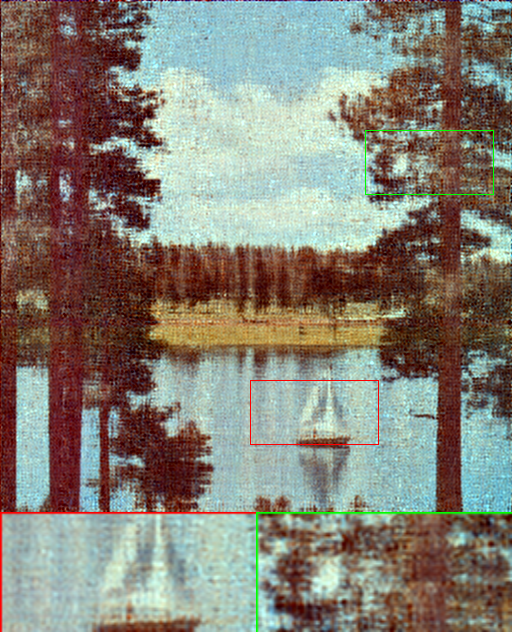}}
\hspace{0.01cm}
\subfloat[PSNR 21.63]
{\includegraphics[width=0.102\linewidth]{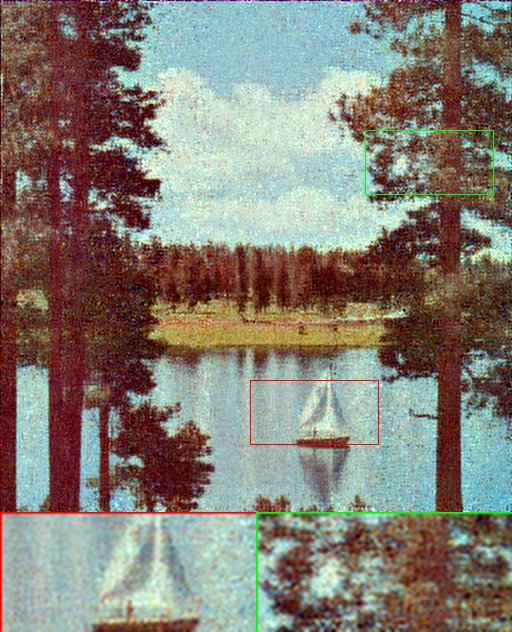}}
\hspace{0.01cm}
\subfloat[PSNR 21.86]
{\includegraphics[width=0.102\linewidth]{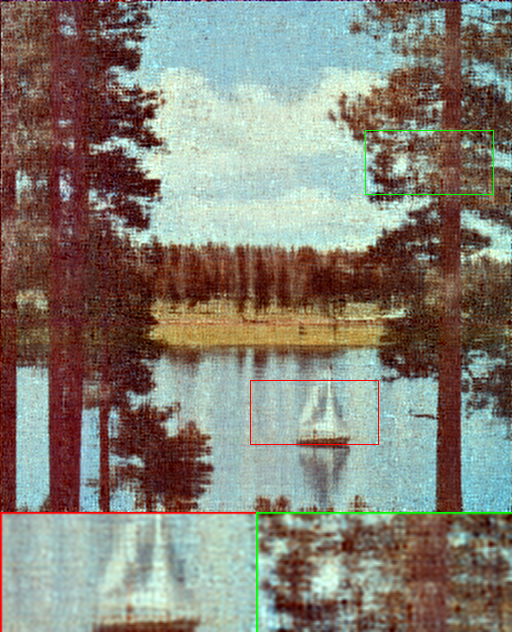}}
\hspace{0.01cm}
\subfloat[PSNR 23.52]
{\includegraphics[width=0.102\linewidth]{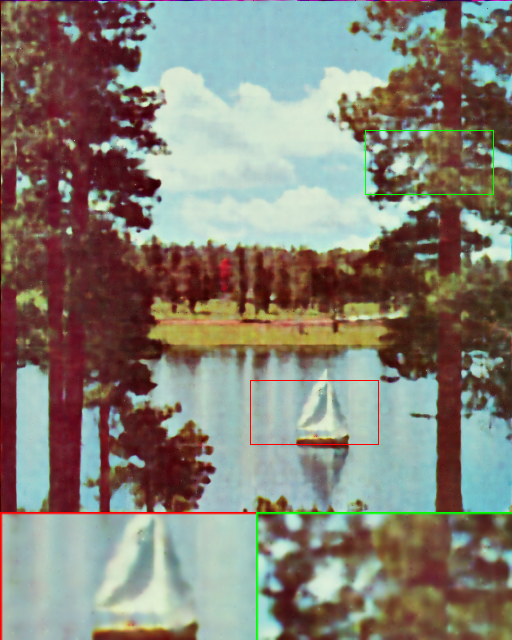}}
\hspace{0.01cm}
\subfloat[PSNR 24.65]
{\includegraphics[width=0.102\linewidth]{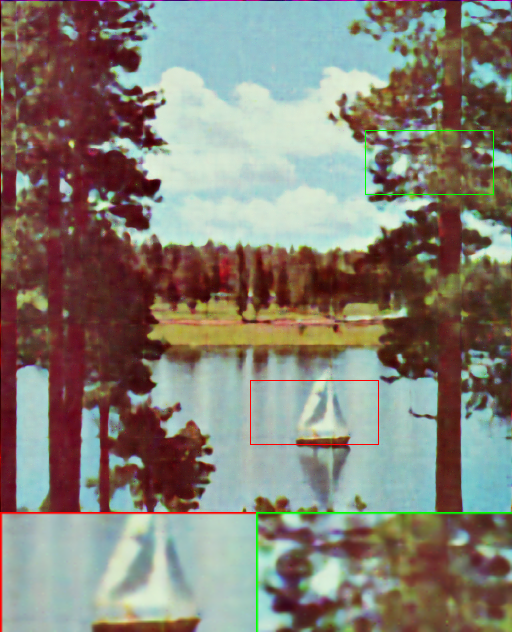}}
\hspace{0.01cm}
\subfloat[PSNR Inf]
{\includegraphics[width=0.102\linewidth]{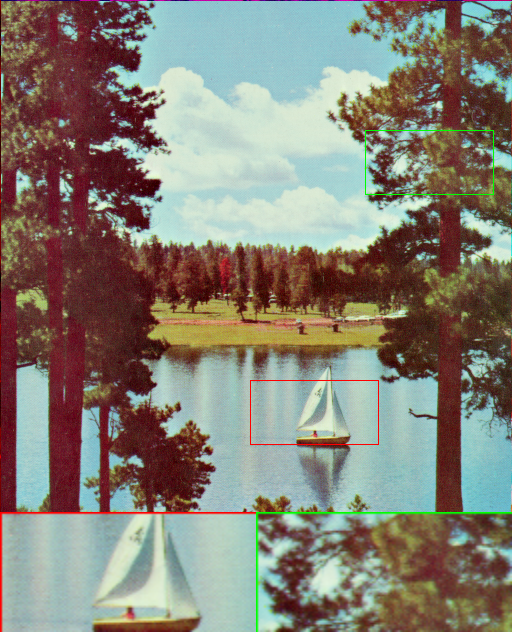}}
\end{minipage}
	\begin{minipage}{1.\linewidth}
\centering
\subfloat[PSNR 6.892]
{\includegraphics[width=0.102\linewidth]{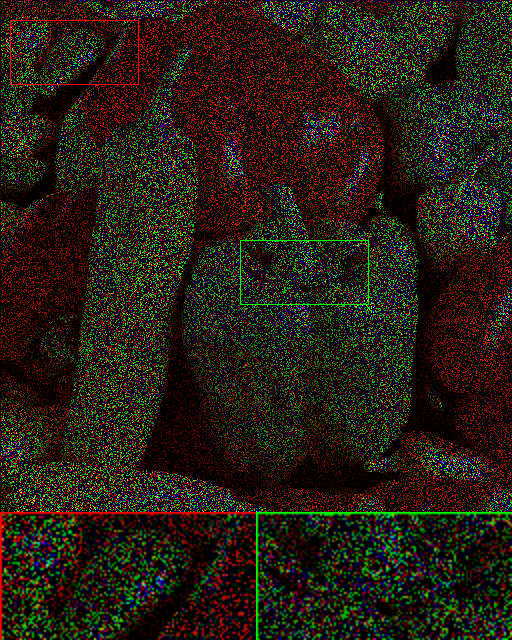}}
\hspace{0.01cm}
\subfloat[PSNR 23.70]
{\includegraphics[width=0.102\linewidth]{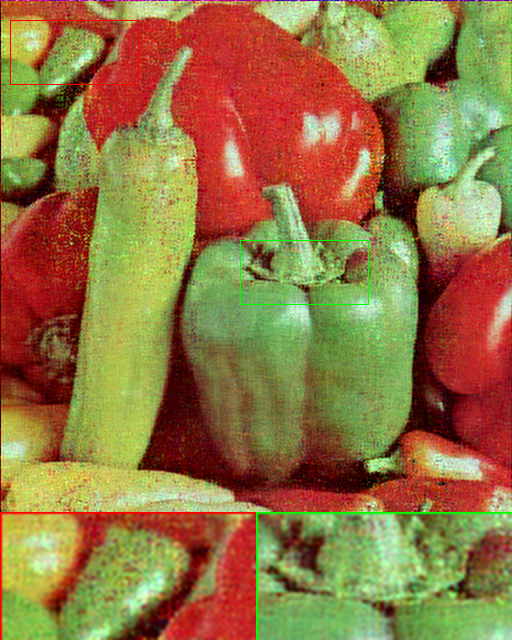}}
\hspace{0.01cm}
\subfloat[PSNR 26.65]
{\includegraphics[width=0.102\linewidth]{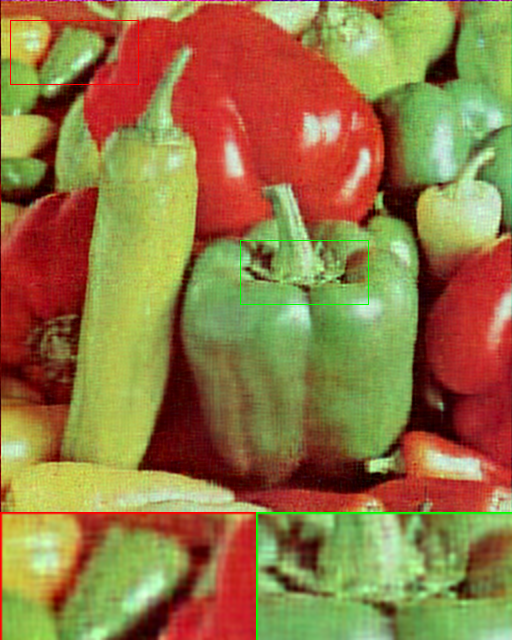}}
\hspace{0.01cm}
\subfloat[PSNR 26.93]
{\includegraphics[width=0.102\linewidth]{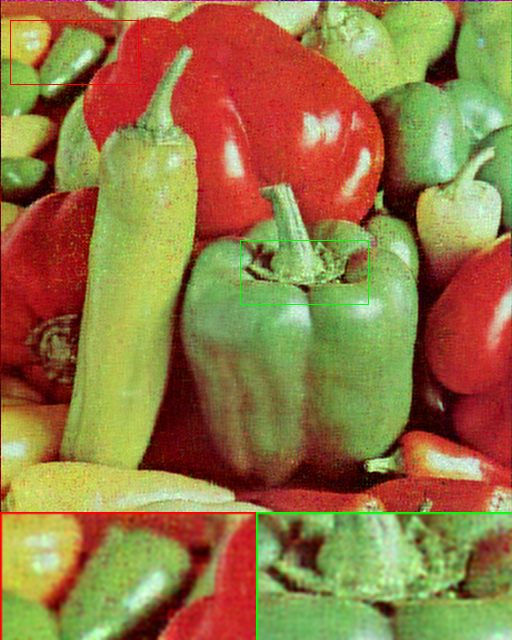}}
\hspace{0.01cm}
\subfloat[PSNR 25.96]
{\includegraphics[width=0.102\linewidth]{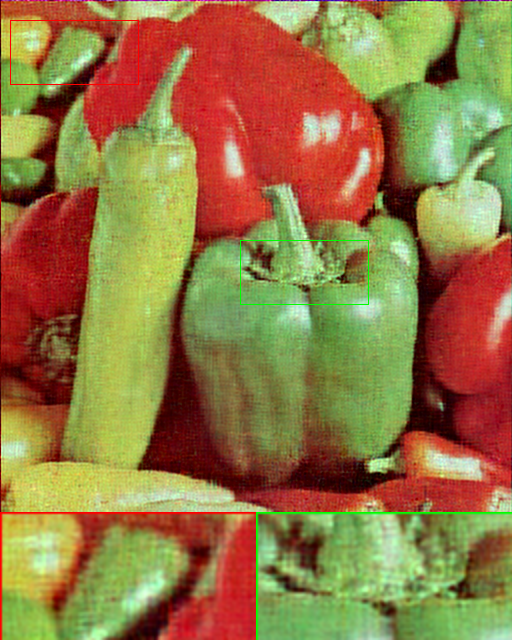}}
\hspace{0.01cm}
\subfloat[PSNR 28.24]
{\includegraphics[width=0.102\linewidth]{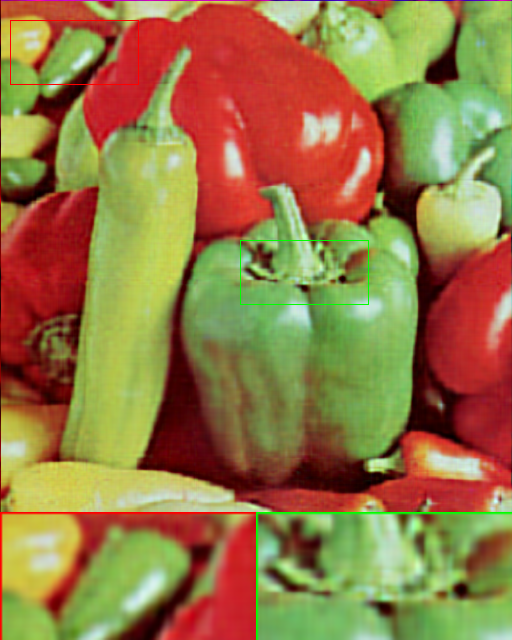}}
\hspace{0.01cm}
\subfloat[PSNR 30.09]
{\includegraphics[width=0.102\linewidth]{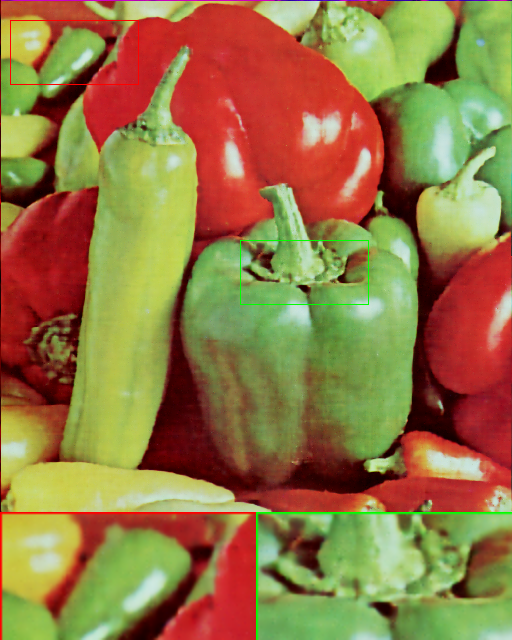}}
\hspace{0.01cm}
\subfloat[PSNR 30.56]
{\includegraphics[width=0.102\linewidth]{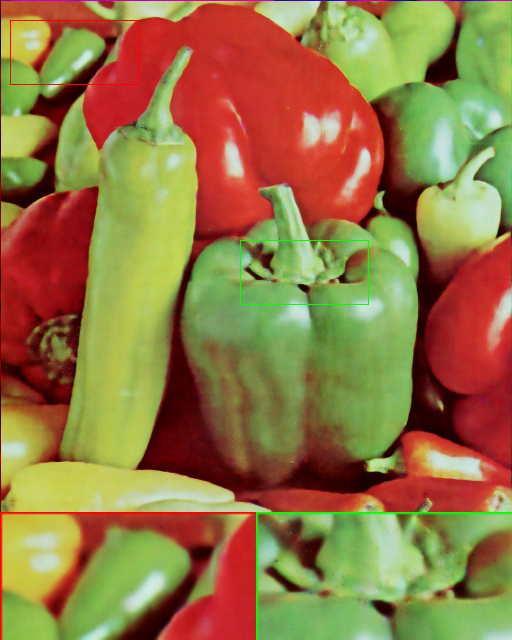}}
\hspace{0.01cm}
\subfloat[PSNR Inf]
{\includegraphics[width=0.102\linewidth]{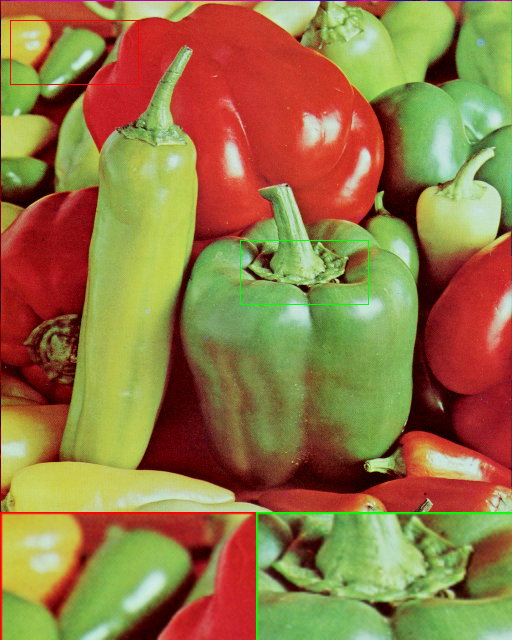}}
\end{minipage}
	\begin{minipage}{1.\linewidth}
\centering
\subfloat[PSNR 11.49]
{\includegraphics[width=0.102\linewidth]{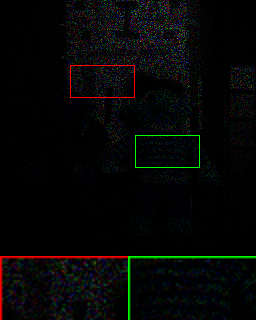}}
\hspace{0.01cm}
\subfloat[PSNR 34.59]
{\includegraphics[width=0.102\linewidth]{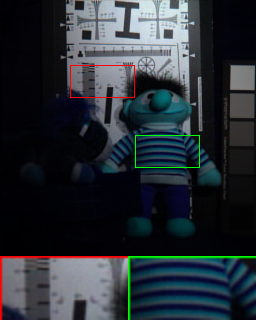}}
\hspace{0.01cm}
\subfloat[PSNR 36.87]
{\includegraphics[width=0.102\linewidth]{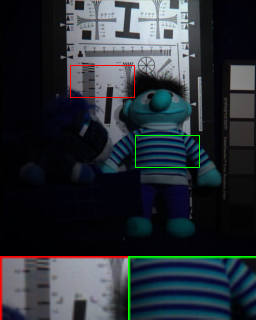}}
\hspace{0.01cm}
\subfloat[PSNR 38.24]
{\includegraphics[width=0.102\linewidth]{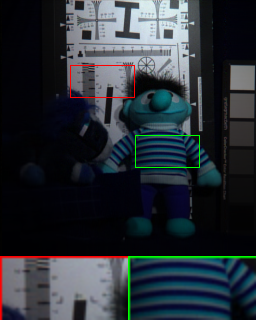}}
\hspace{0.01cm}
\subfloat[PSNR 39.17]
{\includegraphics[width=0.102\linewidth]{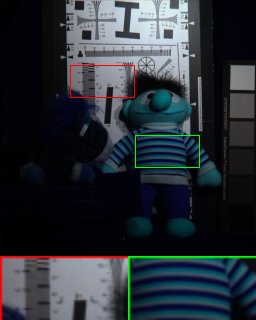}}
\hspace{0.01cm}
\subfloat[PSNR 40.39]
{\includegraphics[width=0.102\linewidth]{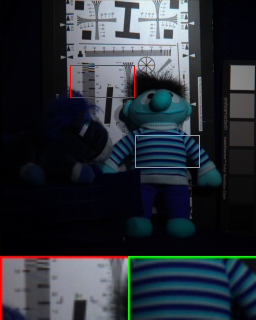}}
\hspace{0.01cm}
\subfloat[PSNR 41.07]
{\includegraphics[width=0.102\linewidth]{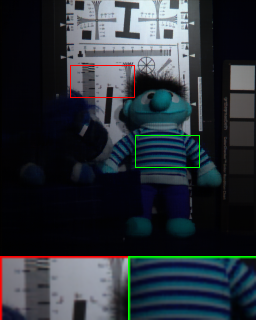}}
\hspace{0.01cm}
\subfloat[PSNR 41.47]
{\includegraphics[width=0.102\linewidth]{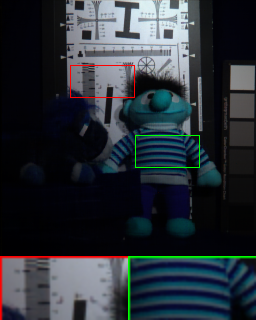}}
\hspace{0.01cm}
\subfloat[PSNR Inf]
{\includegraphics[width=0.102\linewidth]{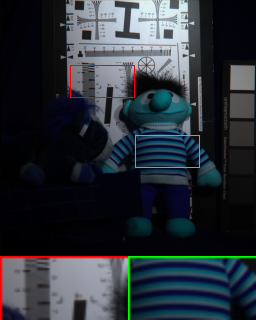}}
\end{minipage}
	\begin{minipage}{1.\linewidth}
\centering
\subfloat[PSNR 16.94]
{\includegraphics[width=0.102\linewidth]{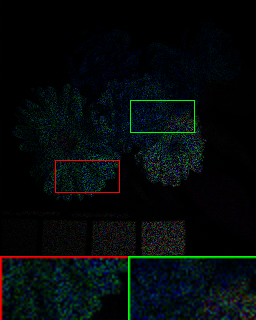}}
\hspace{0.01cm}
\subfloat[PSNR 39.52]
{\includegraphics[width=0.102\linewidth]{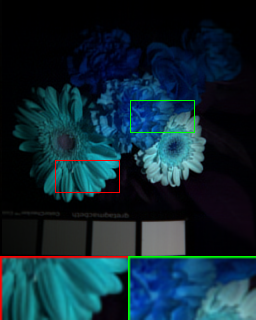}}
\hspace{0.01cm}
\subfloat[PSNR 41.00]
{\includegraphics[width=0.102\linewidth]{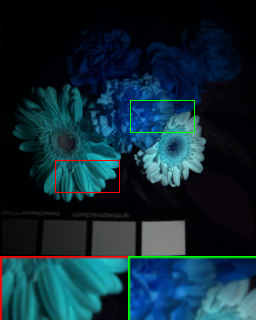}}
\hspace{0.01cm}
\subfloat[PSNR 44.08]
{\includegraphics[width=0.102\linewidth]{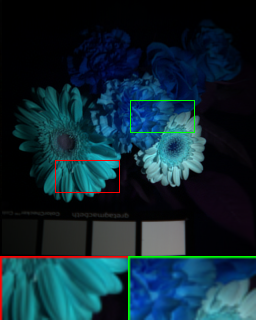}}
\hspace{0.01cm}
\subfloat[PSNR 42.36]
{\includegraphics[width=0.102\linewidth]{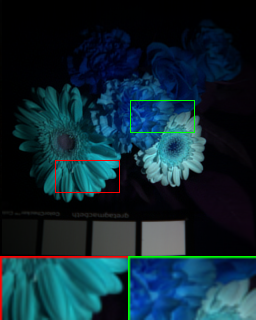}}
\hspace{0.01cm}
\subfloat[PSNR 45.30]
{\includegraphics[width=0.102\linewidth]{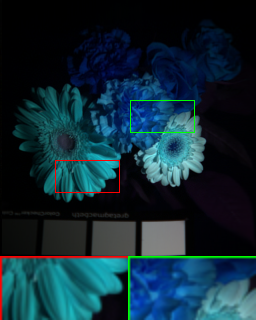}}
\hspace{0.01cm}
\subfloat[PSNR 46.07]
{\includegraphics[width=0.102\linewidth]{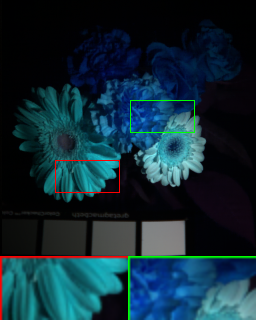}}
\hspace{0.01cm}
\subfloat[PSNR 45.38]
{\includegraphics[width=0.102\linewidth]{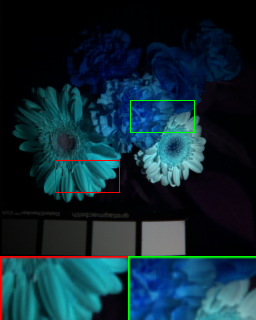}}
\hspace{0.01cm}
\subfloat[PSNR Inf]
{\includegraphics[width=0.102\linewidth]{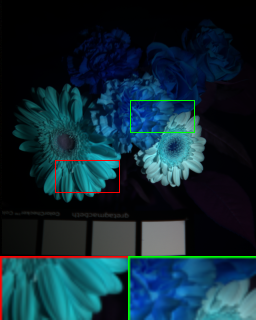}}
\end{minipage}
	\begin{minipage}{1.\linewidth}
\centering
\subfloat[PSNR 4.048]
{\includegraphics[width=0.102\linewidth]{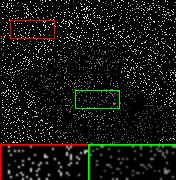}}
\hspace{0.01cm}
\subfloat[PSNR 24.46]
{\includegraphics[width=0.102\linewidth]{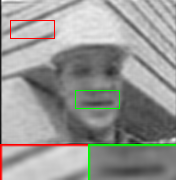}}
\hspace{0.01cm}
\subfloat[PSNR 25.60]
{\includegraphics[width=0.102\linewidth]{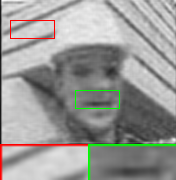}}
\hspace{0.01cm}
\subfloat[PSNR 25.91]
{\includegraphics[width=0.102\linewidth]{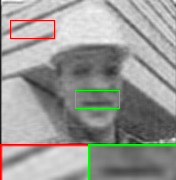}}
\hspace{0.01cm}
\subfloat[PSNR 27.05]
{\includegraphics[width=0.102\linewidth]{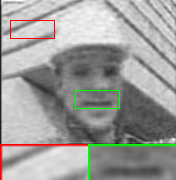}}
\hspace{0.01cm}
\subfloat[PSNR 27.14]
{\includegraphics[width=0.102\linewidth]{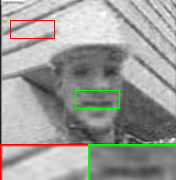}}
\hspace{0.01cm}
\subfloat[PSNR 32.92]
{\includegraphics[width=0.102\linewidth]{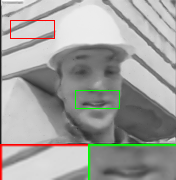}}
\hspace{0.01cm}
\subfloat[PSNR 32.62]
{\includegraphics[width=0.102\linewidth]{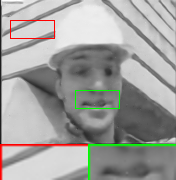}}
\hspace{0.01cm}
\subfloat[PSNR Inf]
{\includegraphics[width=0.102\linewidth]{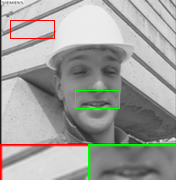}}
\end{minipage}
	\begin{minipage}{1.\linewidth}
\centering
\subfloat[\shortstack{ PSNR 7.559\\ Observed}]
{\includegraphics[width=0.102\linewidth]{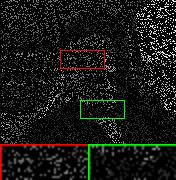}}
\hspace{0.01cm}
\subfloat[\shortstack{ PSNR 27.25\\ M\textsuperscript{2}DMT }]
{\includegraphics[width=0.102\linewidth]{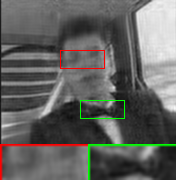}}
\hspace{0.01cm}
\subfloat[\shortstack{ PSNR 27.42\\ LRTC-ENR }]
{\includegraphics[width=0.102\linewidth]{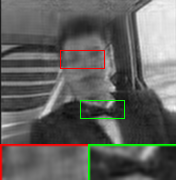}}
\hspace{0.01cm}
\subfloat[\shortstack{PSNR 28.85\\ HLRTF }]
{\includegraphics[width=0.102\linewidth]{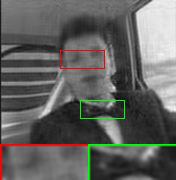}}
\hspace{0.01cm}
\subfloat[\shortstack{PSNR 29.81\\ DeepTensor }]
{\includegraphics[width=0.102\linewidth]{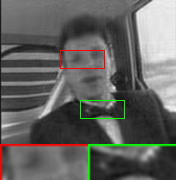}}
\hspace{0.01cm}
\subfloat[\shortstack{PSNR 31.31\\ LRTFR }]
{\includegraphics[width=0.102\linewidth]{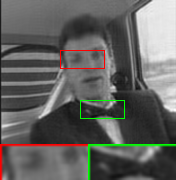}}
\hspace{0.01cm}
\subfloat[\shortstack{PSNR 36.41\\ ScoreTR }]
{\includegraphics[width=0.102\linewidth]{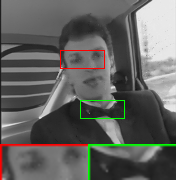}}
\hspace{0.01cm}
\subfloat[\shortstack{PSNR 35.25\\ DiffTR}]
{\includegraphics[width=0.102\linewidth]{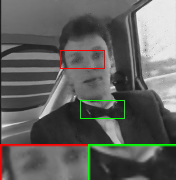}}
\hspace{0.01cm}
\subfloat[\shortstack{PSNR Inf\\ Original }]
{\includegraphics[width=0.102\linewidth]{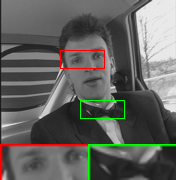}}
\end{minipage}
	\caption{\textbf{Qualitative comparison of visual inpainting}. RGB images, MSIs, and videos are reconstructed at sampling rates of 0.1 (odd rows) and 0.2 (even rows). Insets show enlarged details.}
	\label{fig:Demo4Inpainting}
\end{figure*}

Tables~\ref{tab:imageInpainting} and~\ref{tab:imageInpainting2} show that DiffTR achieves the best PSNR, SSIM, and NRMSE across all 30 combinations of modality, sampling rate, and observation pattern. At a 5\% entry-wise sampling rate, it exceeds the strongest baseline in PSNR by 1.28~dB on RGB images, 0.53~dB on MSIs, and 0.50~dB on videos. For tube-wise MSI recovery, it outperforms FA-INR by 2.63--2.80~dB across the tested sampling rates. The reconstructions in Figure~\ref{fig:Demo4Inpainting} further illustrate its preservation of object boundaries and local detail. Together, these results demonstrate that DiffTR effectively recovers both missing entries and tubes across diverse tensor modalities within a common conditional diffusion formulation.

\subsubsection{Image Denoising}
\begin{table*}[!h]
	\centering
	\caption{\textbf{Quantitative comparison of HSI and MSI denoising}. Entries report dataset-average PSNR/SSIM/NRMSE under five noise settings. Bold and underlined values indicate the best and second-best results, respectively, including ties.}
	\label{tab:compareDenoise}
	\fitTable{
		\begin{tabular}{cc|ccc|ccc|ccc|ccc|ccc}
			\toprule
			\multicolumn{2}{c}{Noise case} &\multicolumn{3}{c}{Case 1} & \multicolumn{3}{c}{Case 2} & \multicolumn{3}{c}{Case 3} &\multicolumn{3}{c}{Case 4} &\multicolumn{3}{c}{Case 5} \\
			\midrule
			Data &Method &PSNR &SSIM &NRMSE &PSNR &SSIM &NRMSE &PSNR &SSIM &NRMSE &PSNR &SSIM &NRMSE &PSNR &SSIM &NRMSE\\
			\midrule
			\multirow{7}{*}{\makecell[c]{HSIs\\WDC mall\\$4\times(320\times 307\times 191)$}}
			&Observed
			&15.44 &0.1560 &0.5890 &16.12 &0.2010 &0.5450 &15.20 &0.1710 &0.6060 &15.73 &0.1910 &0.5700 &14.91 &0.1650 &0.6260\\
			&LRTC-ENR\cite{fan2023euclideannorminduced}
			&28.07 &0.7250 &0.1570 &28.41 &0.8200 &0.1260 &28.79 &0.7920 &0.1100 &26.25 &0.7680 &0.1670 &25.21 &0.7460 &0.1510\\
			&HLRTF\cite{luo2022hlrtf}
			&27.76 &0.7360 &0.1440 &28.95 &0.8260 &0.1200 &30.87 &0.8390 &\underline{0.09300} &28.03 &0.7810 &0.1510 &27.69 &0.7420 &0.1440\\
			&DeepTensor\cite{saragadam2024deeptensor}
			&26.41 &0.7410 &0.1410 &27.88 &0.8090 &0.1610 &27.66 &0.7710 &0.1630 &25.81 &0.7540 &0.1580 &24.11 &0.7340 &0.1790\\
			&LRTFR\cite{luo2023low}
			&\underline{30.89} &\underline{0.8410} &\underline{0.1000} &33.35 &\underline{0.8980} &0.07500 &\underline{31.08} &\underline{0.8840} &0.09800 &28.11 &0.8120 &0.1370 &\textbf{27.96} &\underline{0.8080} &\textbf{0.1400}\\
			&ScoreTR\cite{cheng2025score}
			&\textbf{31.89} &\textbf{0.9133} &\textbf{0.08982} &\textbf{35.00} &\textbf{0.9362} &\textbf{0.06277} &\textbf{31.63} &\textbf{0.9268} &\textbf{0.09224} &\underline{28.67} &\textbf{0.8615} &\underline{0.1302} &27.70 &\textbf{0.8600} &0.1460\\
			&\textbf{DiffTR}
			&30.64 &0.8247 &0.1025 &\underline{33.38} &0.8943 &\underline{0.07480} &30.58 &0.8478 &0.1032 &\textbf{29.39} &\underline{0.8290} &\textbf{0.1180} &\underline{27.75} &0.7990 &\underline{0.1430}\\
			\midrule
			\multirow{7}{*}{\makecell[c]{MSIs\\PaviaU\\$2\times(305\times 340\times 103)$}}
			&Observed
			&15.54 &0.1500 &0.8150 &16.11 &0.1950 &0.7630 &15.81 &0.1720 &0.7900 &15.95 &0.1850 &0.7780 &15.67 &0.1650 &0.8030\\
			&LRTC-ENR\cite{fan2023euclideannorminduced}
			&28.58 &0.7740 &0.1810 &29.38 &0.8260 &0.1550 &28.23 &0.8310 &0.1690 &26.61 &0.7670 &0.1830 &\textbf{27.92} &\underline{0.8080} &\textbf{0.1890}\\
			&HLRTF\cite{luo2022hlrtf}
			&29.12 &0.7870 &0.1690 &30.08 &0.8420 &0.1320 &28.82 &0.8380 &0.1600 &27.89 &0.7720 &\underline{0.1760} &25.47 &0.7520 &0.2470\\
			&DeepTensor\cite{saragadam2024deeptensor}
			&26.73 &0.7200 &0.1900 &27.66 &0.8200 &0.1790 &27.52 &0.8200 &0.1670 &26.04 &0.7880 &0.1820 &24.83 &0.7350 &0.2630\\
			&LRTFR\cite{luo2023low}
			&\underline{29.52} &\underline{0.7940} &\underline{0.1630} &\underline{31.64} &\underline{0.8650} &\underline{0.1280} &\underline{29.84} &\underline{0.8410} &\underline{0.1570} &28.11 &0.7970 &0.1920 &26.44 &0.7620 &0.2320\\
			&ScoreTR\cite{cheng2025score}
			&\textbf{30.52} &\textbf{0.8876} &\textbf{0.1464} &\textbf{33.29} &\textbf{0.9156} &\textbf{0.1071} &\textbf{30.39} &\textbf{0.8997} &\textbf{0.1478} &\underline{28.67} &\textbf{0.8505} &0.1825 &26.18 &\textbf{0.8264} &0.2419\\
			&\textbf{DiffTR}
			&29.34 &0.7900 &0.1660 &31.06 &0.8328 &0.1367 &28.88 &0.8070 &0.1750 &\textbf{29.52} &\underline{0.8160} &\textbf{0.1630} &\underline{27.70} &0.7850 &\underline{0.2010}\\
			\midrule
			\multirow{7}{*}{\makecell[c]{MSIs\\Balloons\\Beads\\Flowers\\Fruits\\$(256\times 256\times 31)$}}
			&Observed
			&16.22 &0.08400 &0.9020 &16.26 &0.1090 &0.9000 &16.12 &0.1010 &0.9120 &16.20 &0.1070 &0.9060 &16.07 &0.1010 &0.9170\\
			&LRTC-ENR\cite{fan2023euclideannorminduced}
			&31.26 &0.7560 &\underline{0.1520} &33.88 &0.8450 &0.1270 &31.49 &0.7700 &0.1560 &28.96 &0.7530 &0.1740 &28.07 &0.7800 &0.2160\\
			&HLRTF\cite{luo2022hlrtf}
			&30.57 &0.7310 &0.1590 &32.75 &0.7810 &0.1520 &31.51 &0.7910 &0.1580 &29.53 &0.7560 &0.1730 &27.93 &0.7740 &0.2280\\
			&DeepTensor\cite{saragadam2024deeptensor}
			&29.97 &0.7250 &0.1550 &31.03 &0.7470 &0.1400 &30.79 &0.7860 &0.1690 &29.77 &0.7410 &0.1790 &27.89 &0.7720 &0.2210\\
			&LRTFR\cite{luo2023low}
			&31.32 &0.7360 &0.1670 &32.89 &0.7840 &0.1410 &31.96 &0.7940 &0.1530 &31.27 &0.7760 &0.1620 &\underline{29.97} &0.7820 &\underline{0.1870}\\
			&ScoreTR\cite{cheng2025score}
			&\textbf{32.32} &\textbf{0.8560} &\textbf{0.1500} &\textbf{34.54} &\textbf{0.8650} &\textbf{0.1180} &\underline{32.51} &\textbf{0.8700} &\underline{0.1440} &\underline{31.83} &\underline{0.8350} &\underline{0.1540} &29.71 &\textbf{0.8410} &0.1950\\
			&\textbf{DiffTR}
			&\underline{32.19} &\underline{0.8080} &0.1530 &\underline{34.52} &\underline{0.8590} &\underline{0.1200} &\textbf{32.70} &\underline{0.8600} &\textbf{0.1420} &\textbf{32.77} &\textbf{0.8470} &\textbf{0.1410} &\textbf{30.80} &\underline{0.8370} &\textbf{0.1730}\\
			\bottomrule
	\end{tabular}}
\end{table*}

We evaluate mixed-noise removal on WDC Mall and Pavia University HSIs and four CAVE MSIs. Following LRTFR \cite{luo2023low}, we use five noise settings. Case~1 adds Gaussian noise with standard deviation $0.2$. Case~2 combines Gaussian noise with $\sigma=0.1$ and $10\%$ sparse corruption. Case~3 adds zero-valued corruption to Case~2. Case~4 adds stripe noise to Case~2, affecting $10\%$ of pixels in $40\%$ of the bands. Case~5 adds zero-valued corruption to Case~4. DiffTR uses independent per-voxel zero masks in Cases~3 and~5. Training and configuration-selection settings are given in Appendix~\ref{app:denoise_settings}.

\begin{figure*}[!t]
	\centering
	\captionsetup[subfloat]{labelsep=none,format=plain,labelformat=empty,font=footnotesize}
	\begin{minipage}{1.\linewidth}
\centering
\subfloat[PSNR 15.91]
{\includegraphics[width=0.097\linewidth]{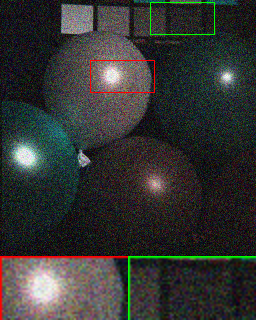}}
\hspace{0.01cm}
\subfloat[PSNR 33.13]
{\includegraphics[width=0.097\linewidth]{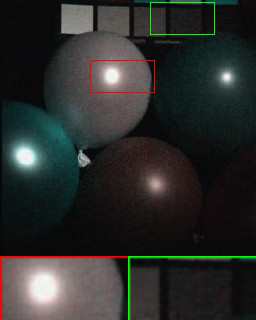}}
\hspace{0.01cm}
\subfloat[PSNR 36.11]
{\includegraphics[width=0.097\linewidth]{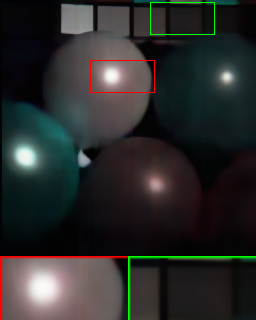}}
\hspace{0.01cm}
\subfloat[PSNR 35.45]
{\includegraphics[width=0.097\linewidth]{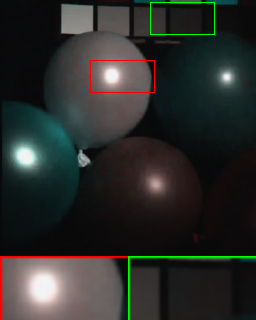}}
\hspace{0.01cm}
\subfloat[PSNR 35.19]
{\includegraphics[width=0.097\linewidth]{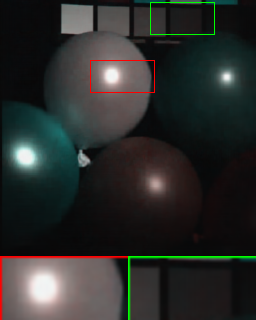}}
\hspace{0.01cm}
\subfloat[PSNR 36.62]
{\includegraphics[width=0.097\linewidth]{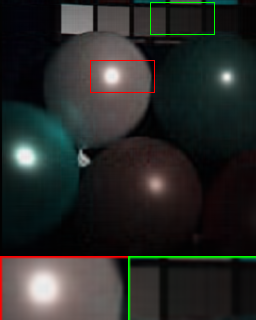}}
\hspace{0.01cm}
\subfloat[PSNR 37.73]
{\includegraphics[width=0.097\linewidth]{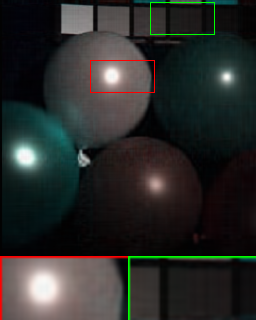}}
\hspace{0.01cm}
\subfloat[PSNR 37.91]
{\includegraphics[width=0.097\linewidth]{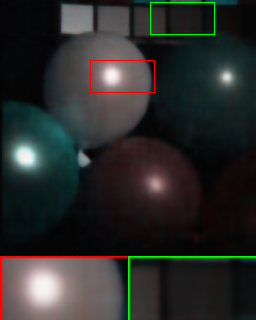}}
\hspace{0.01cm}
\subfloat[PSNR Inf]
{\includegraphics[width=0.097\linewidth]{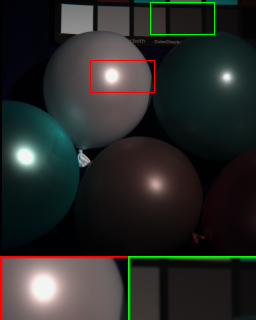}}
\end{minipage}
	\begin{minipage}{1.\linewidth}
\centering
\subfloat[PSNR 16.30]
{\includegraphics[width=0.097\linewidth]{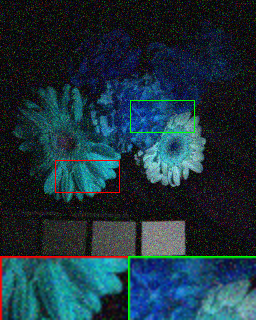}}
\hspace{0.01cm}
\subfloat[PSNR 32.68]
{\includegraphics[width=0.097\linewidth]{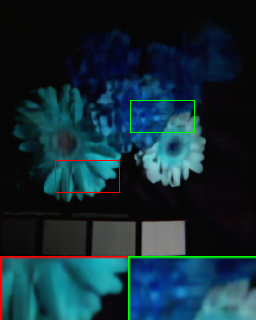}}
\hspace{0.01cm}
\subfloat[PSNR 36.17]
{\includegraphics[width=0.097\linewidth]{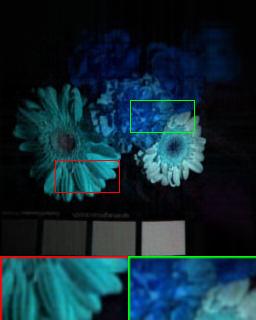}}
\hspace{0.01cm}
\subfloat[PSNR 34.49]
{\includegraphics[width=0.097\linewidth]{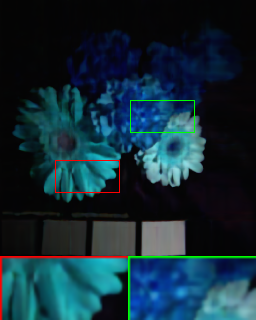}}
\hspace{0.01cm}
\subfloat[PSNR 34.07]
{\includegraphics[width=0.097\linewidth]{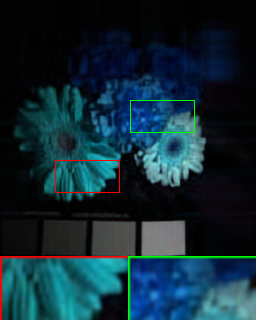}}
\hspace{0.01cm}
\subfloat[PSNR 35.20]
{\includegraphics[width=0.097\linewidth]{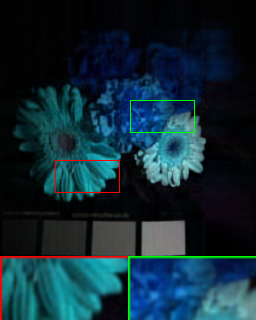}}
\hspace{0.01cm}
\subfloat[PSNR 37.31]
{\includegraphics[width=0.097\linewidth]{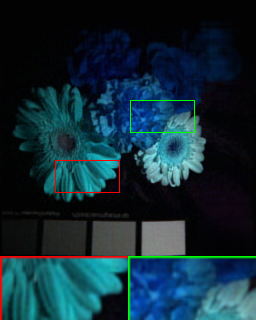}}
\hspace{0.01cm}
\subfloat[PSNR 37.94]
{\includegraphics[width=0.097\linewidth]{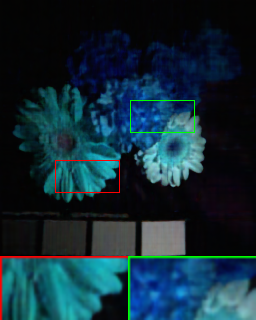}}
\hspace{0.01cm}
\subfloat[PSNR Inf]
{\includegraphics[width=0.097\linewidth]{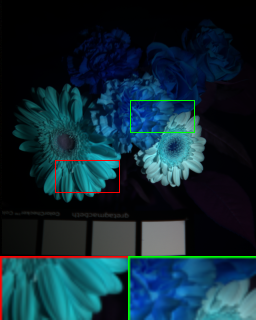}}
\end{minipage}
	\begin{minipage}{1.\linewidth}
\centering
\subfloat[PSNR 15.94]
{\includegraphics[width=0.097\linewidth]{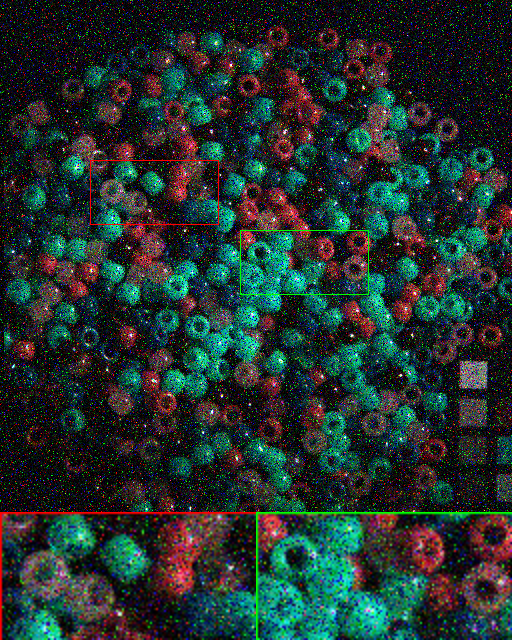}}
\hspace{0.01cm}
\subfloat[PSNR 26.68]
{\includegraphics[width=0.097\linewidth]{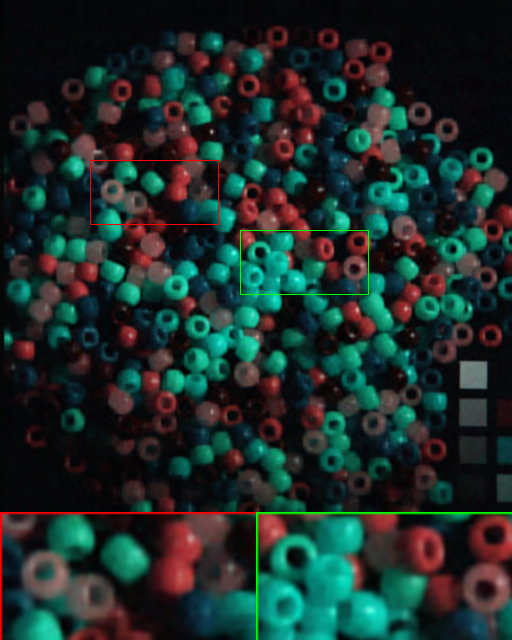}}
\hspace{0.01cm}
\subfloat[PSNR 27.18]
{\includegraphics[width=0.097\linewidth]{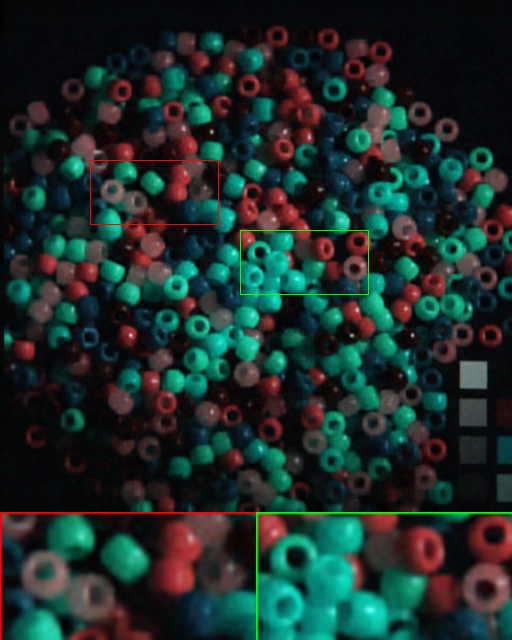}}
\hspace{0.01cm}
\subfloat[PSNR 27.20]
{\includegraphics[width=0.097\linewidth]{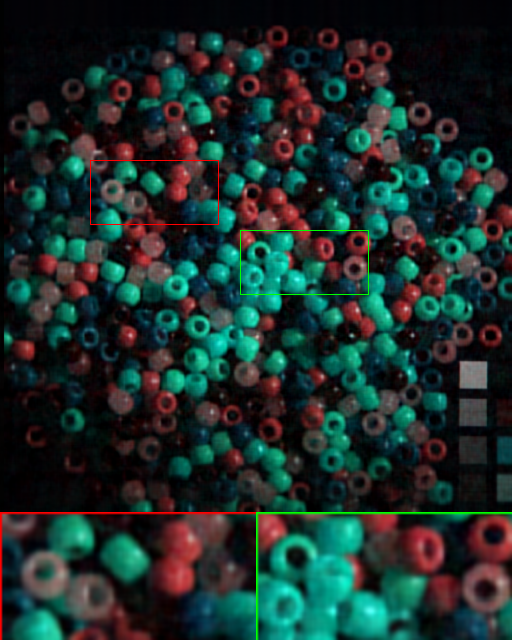}}
\hspace{0.01cm}
\subfloat[PSNR 26.08]
{\includegraphics[width=0.097\linewidth]{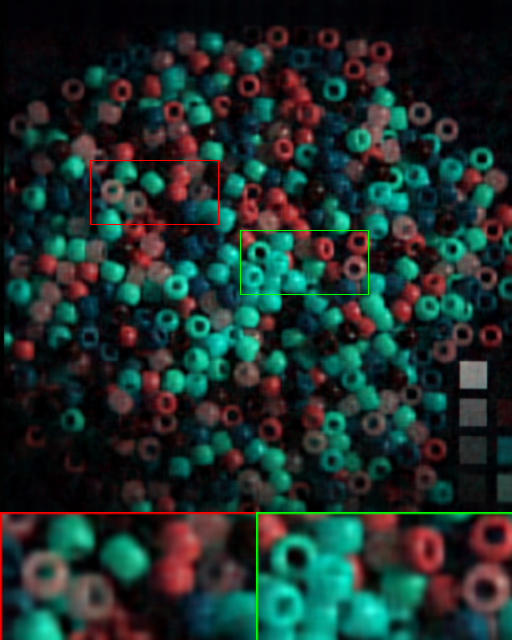}}
\hspace{0.01cm}
\subfloat[PSNR 27.93]
{\includegraphics[width=0.097\linewidth]{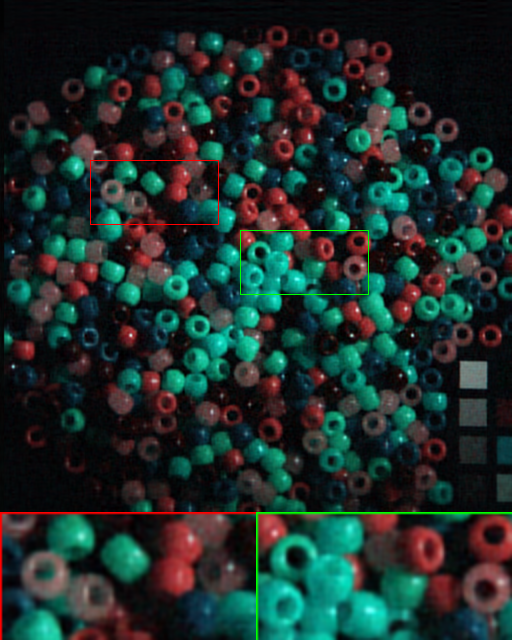}}
\hspace{0.01cm}
\subfloat[PSNR 28.52]
{\includegraphics[width=0.097\linewidth]{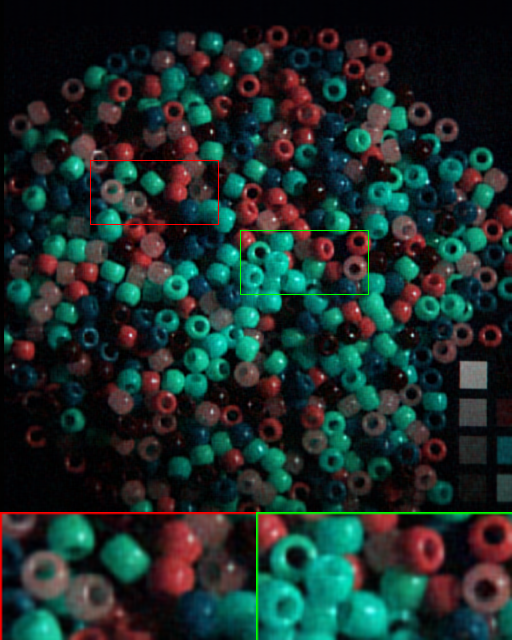}}
\hspace{0.01cm}
\subfloat[PSNR 28.07]
{\includegraphics[width=0.097\linewidth]{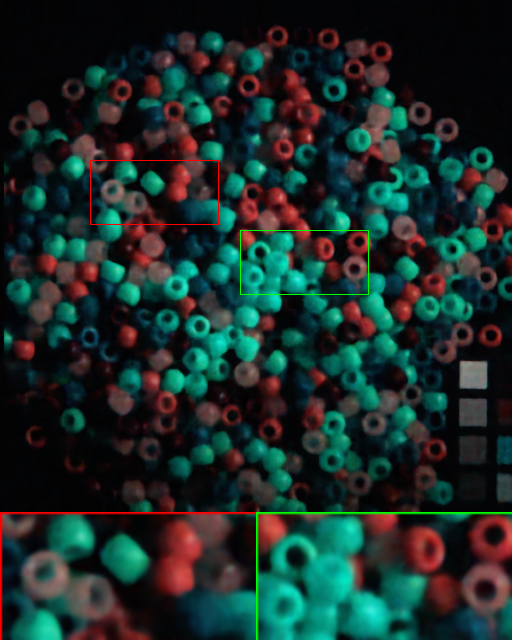}}
\hspace{0.01cm}
\subfloat[PSNR Inf]
{\includegraphics[width=0.097\linewidth]{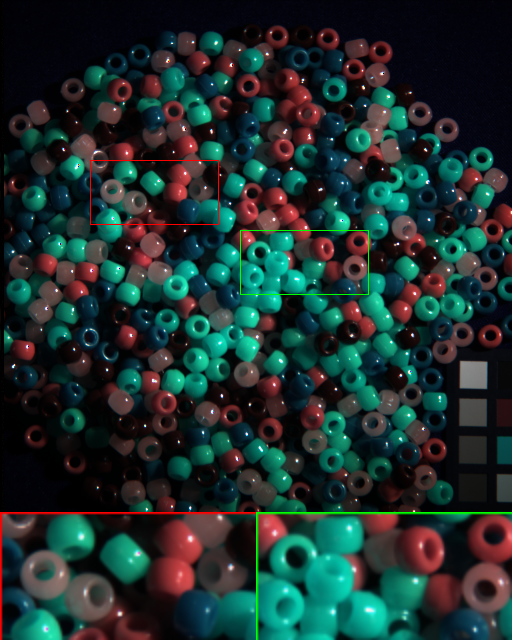}}
\end{minipage}
	\begin{minipage}{1.\linewidth}
\centering
\subfloat[PSNR 15.90]
{\includegraphics[width=0.097\linewidth]{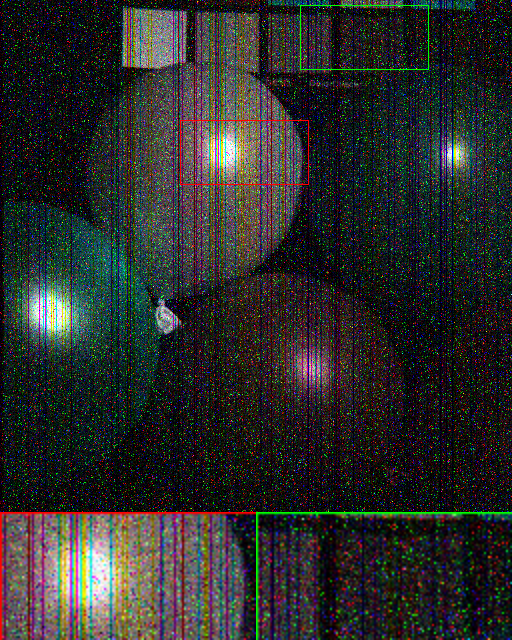}}
\hspace{0.01cm}
\subfloat[PSNR 29.89]
{\includegraphics[width=0.097\linewidth]{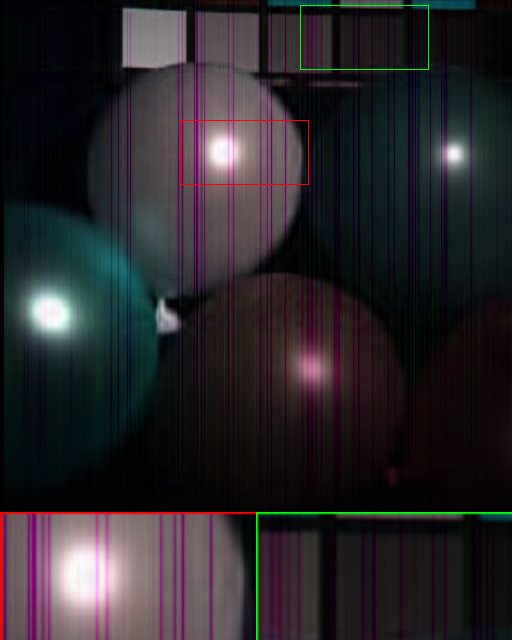}}
\hspace{0.01cm}
\subfloat[PSNR 32.06]
{\includegraphics[width=0.097\linewidth]{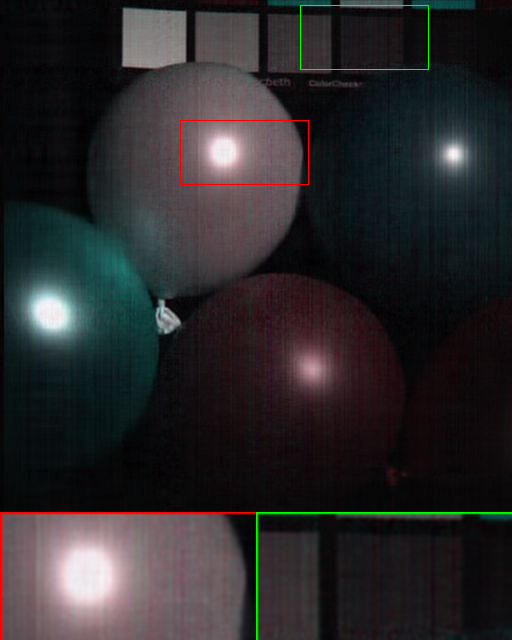}}
\hspace{0.01cm}
\subfloat[PSNR 32.41]
{\includegraphics[width=0.097\linewidth]{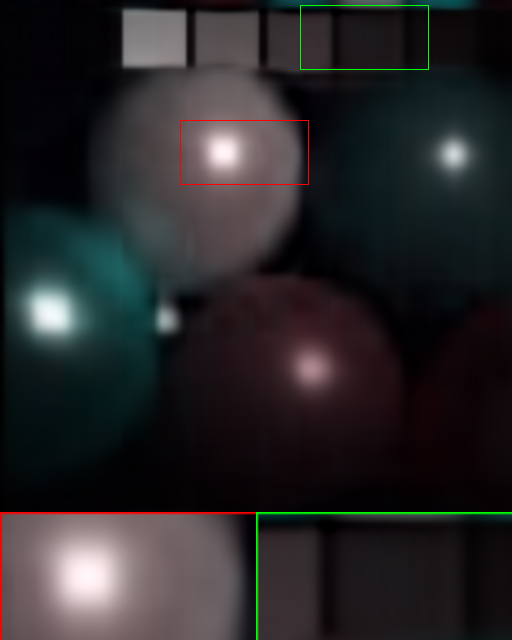}}
\hspace{0.01cm}
\subfloat[PSNR 30.75]
{\includegraphics[width=0.097\linewidth]{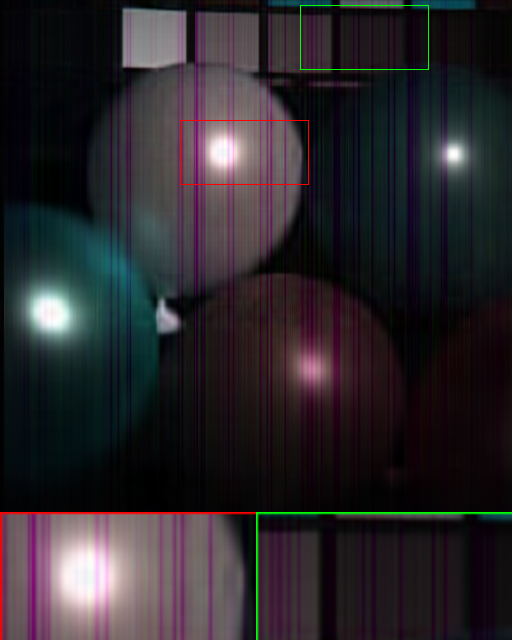}}
\hspace{0.01cm}
\subfloat[PSNR 33.80]
{\includegraphics[width=0.097\linewidth]{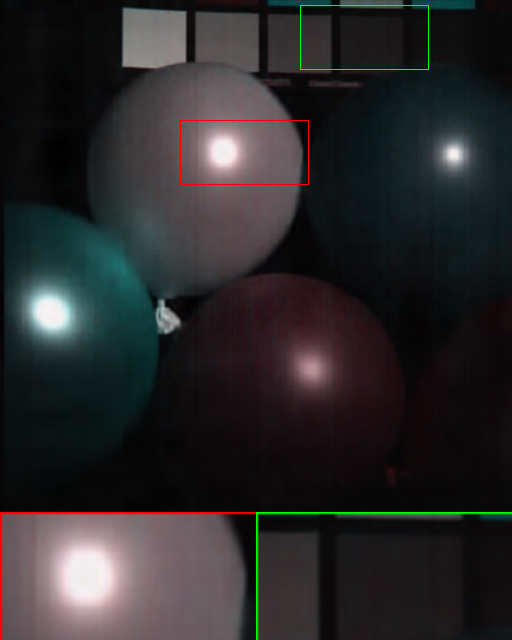}}
\hspace{0.01cm}
\subfloat[PSNR 34.43]
{\includegraphics[width=0.097\linewidth]{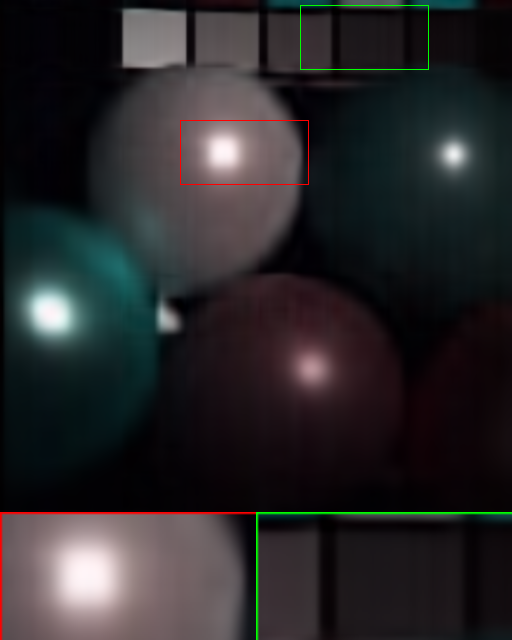}}
\hspace{0.01cm}
\subfloat[PSNR 35.09]
{\includegraphics[width=0.097\linewidth]{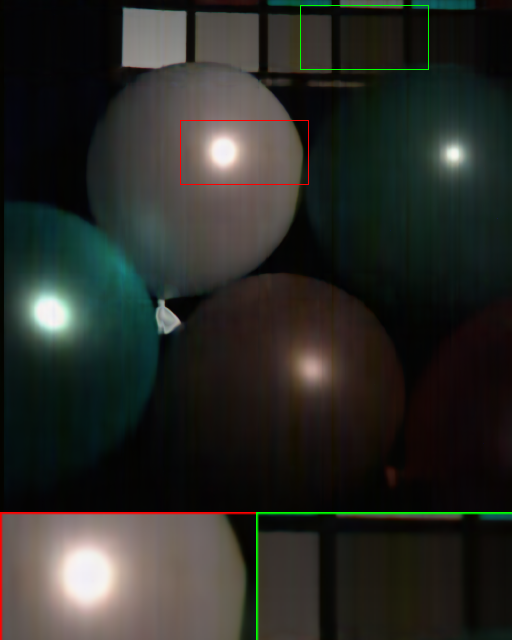}}
\hspace{0.01cm}
\subfloat[PSNR Inf]
{\includegraphics[width=0.097\linewidth]{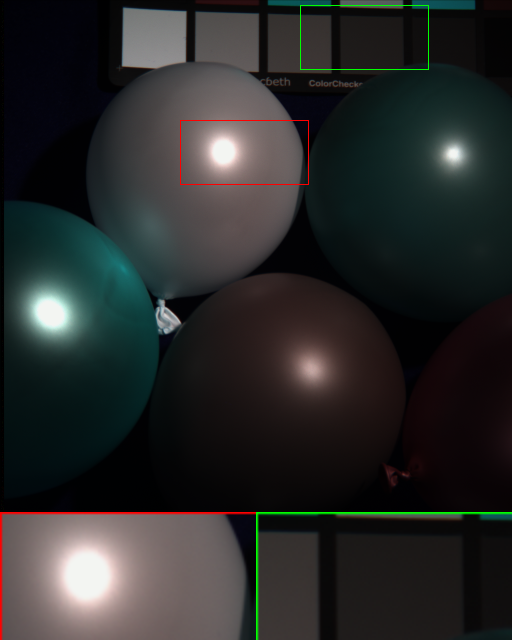}}
\end{minipage}
	\begin{minipage}{1.\linewidth}
\centering
\subfloat[\shortstack{ PSNR 16.49\\ Observed}]
{\includegraphics[width=0.097\linewidth]{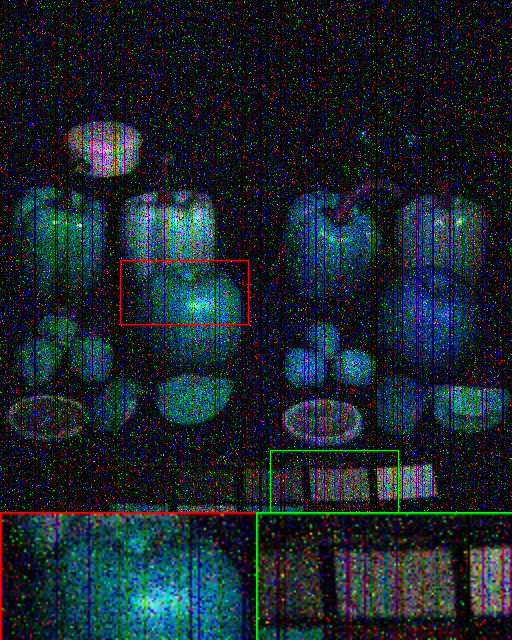}}
\hspace{0.01cm}
\subfloat[\shortstack{PSNR 28.55\\ M\textsuperscript{2}DMT}]
{\includegraphics[width=0.097\linewidth]{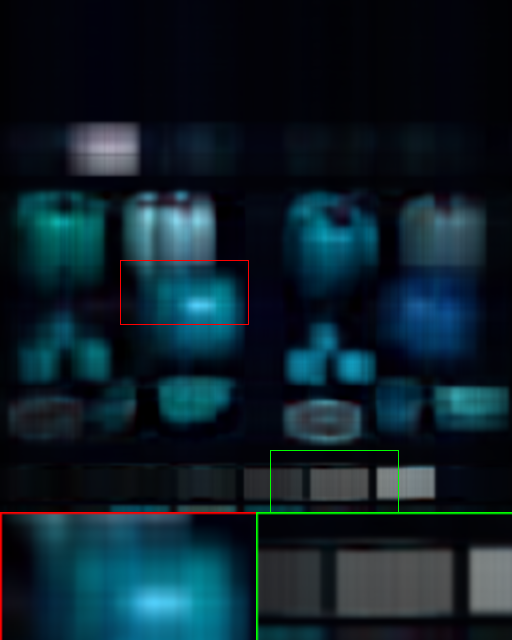}}
\hspace{0.01cm}
\subfloat[\shortstack{PSNR 29.93\\ LRTC-ENR}]
{\includegraphics[width=0.097\linewidth]{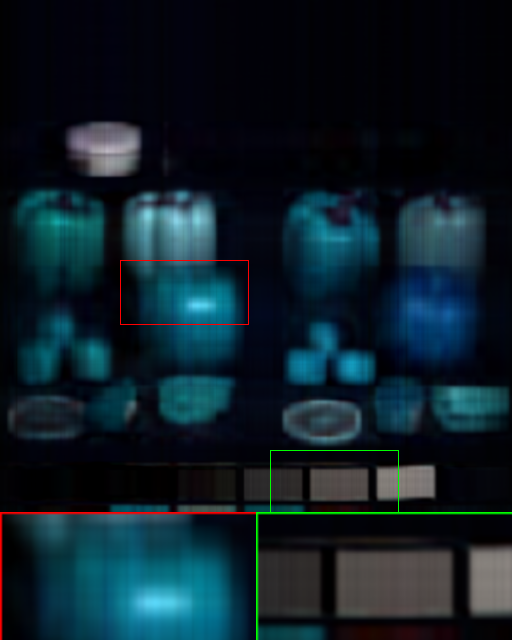}}
\hspace{0.01cm}
\subfloat[\shortstack{PSNR 30.65\\ HLRTF}]
{\includegraphics[width=0.097\linewidth]{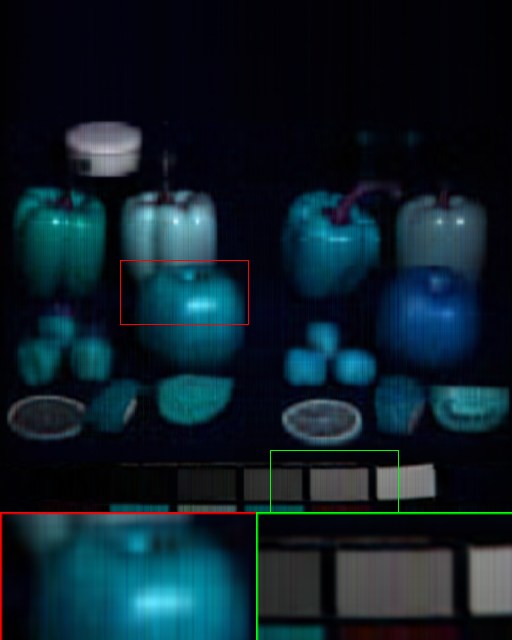}}
\hspace{0.01cm}
\subfloat[\shortstack{PSNR 29.51\\ DeepTensor}]
{\includegraphics[width=0.097\linewidth]{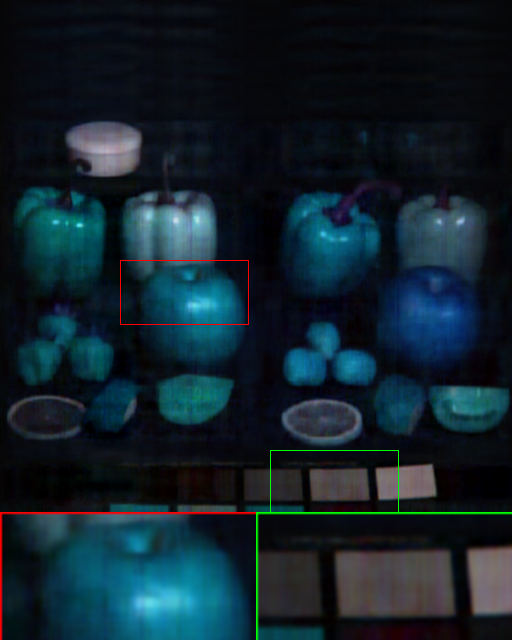}}
\hspace{0.01cm}
\subfloat[\shortstack{PSNR 31.42\\ LRTFR}]
{\includegraphics[width=0.097\linewidth]{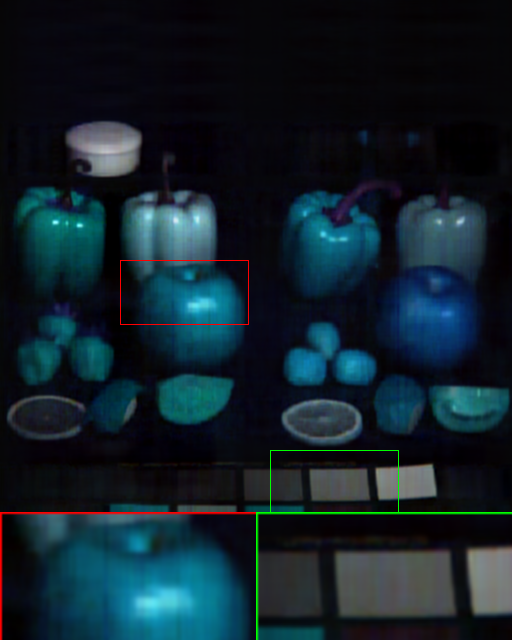}}
\hspace{0.01cm}
\subfloat[\shortstack{PSNR 31.85\\ ScoreTR}]
{\includegraphics[width=0.097\linewidth]{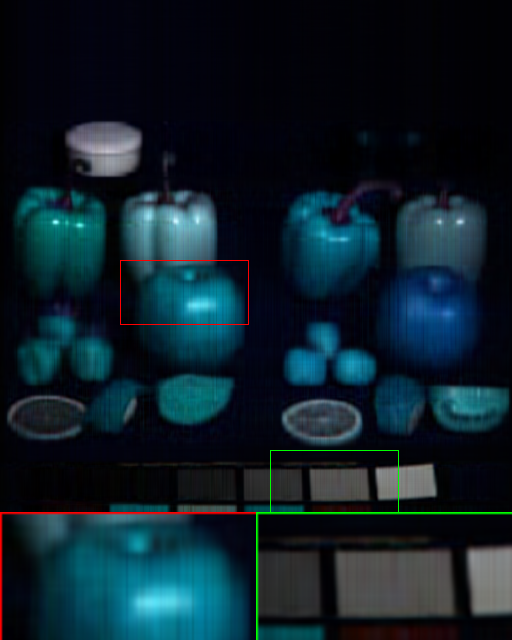}}
\hspace{0.01cm}
\subfloat[\shortstack{PSNR 33.64\\ DiffTR}]
{\includegraphics[width=0.097\linewidth]{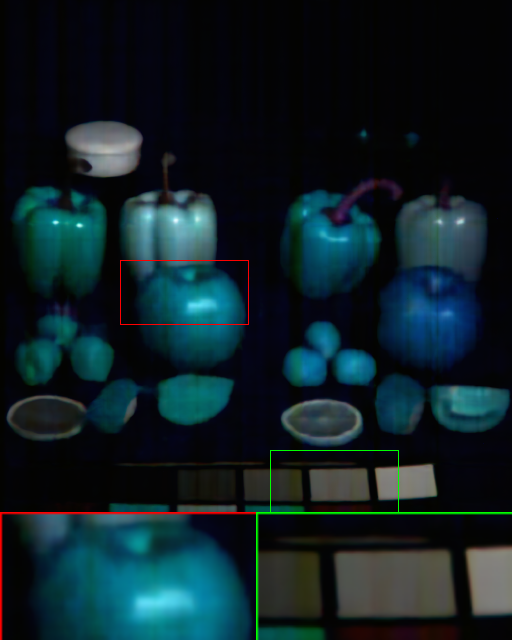}}
\hspace{0.01cm}
\subfloat[\shortstack{PSNR Inf\\ Original}]
{\includegraphics[width=0.097\linewidth]{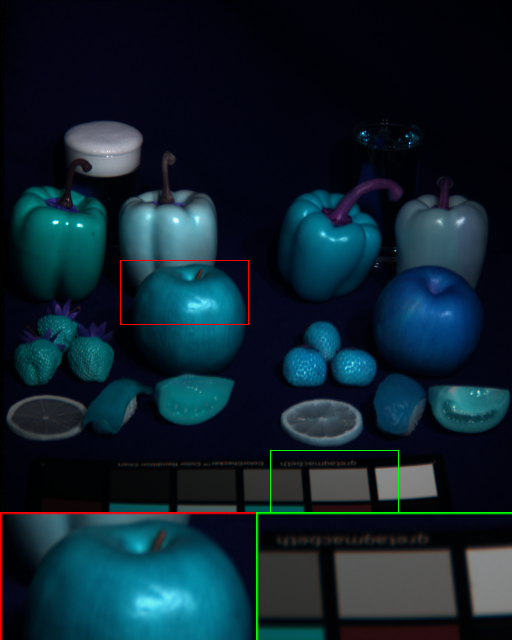}}
\end{minipage}
	
	\caption{\textbf{Qualitative comparison of MSI denoising}. Rows show Balloons, Flowers, Beads, Balloons, and Fruits under Cases 1--5, respectively. Insets show enlarged details.}
	\label{fig:Demo4Denoising}
\end{figure*}

Table~\ref{tab:compareDenoise} shows that DiffTR achieves the highest PSNR and lowest NRMSE in Case~4 across all three datasets, with PSNR gains of 0.72, 0.85, and 0.94~dB over the strongest baseline on WDC, PaviaU, and CAVE, respectively. It also leads both metrics on CAVE in Cases~3 and~5, while ScoreTR performs better on the HSI datasets in Cases~1--3. Figure~\ref{fig:Demo4Denoising} illustrates its preservation of object contours and local detail under independently generated noise realizations. These results demonstrate DiffTR's effectiveness for mixed-noise recovery, particularly in the tested stripe-noise setting.

\subsection{Hyperspectral Image Super-Resolution}
\label{sec:sr_experiment}
We evaluate unsupervised HSI super-resolution on WDC, PaviaU, and four CAVE scenes (Beers, Clay, Flowers, and Balloons), comparing DiffTR with CS2DIPs~\cite{fang2024cs2dips}, GTNN~\cite{dian2024hyperspectral}, GNLTN~\cite{xu2025nonlinear}, and EDIP-Net~\cite{li2025enhanced}. The observations are $\mathcal X_{\mathrm{LR}}=D_sG_\sigma(\mathcal X)$ and $\mathcal Y=G_\sigma(\mathcal X\times_3\mathbf R)$, where $G_\sigma$ applies a $7\times7$ Gaussian kernel with standard deviation $\sigma$, and $D_s$ performs block-average downsampling by a factor of $s$. Both degradation operators are known, and no measurement noise is added. We test $(s,\sigma)\in\{(4,0.5),(4,2),(8,0.5),(8,2),(16,0.5)\}$.
Each DiffTR model is trained on nearest-neighbor-upsampled LR-HSI. Reconstruction uses guided DDPM sampling followed by post-sampling refinement in Eq.~\eqref{eq:sr_terminal_proximal}, with $\tau=10^3$, $\gamma=10$, and a basis $B$ retaining 99.9\% of the uncentered LR-HSI spectral energy. These settings are fixed across all experiments in this subsection.

\begin{table*}[t]
\centering
\caption{\textbf{Quantitative comparison of HSI super-resolution}. Results are averaged over scenes under matched optimization budgets. Bold and underlined values indicate the best and second-best results, respectively, including ties.}
\label{tab:superRes}
{\small\setlength{\tabcolsep}{2pt}
\fitTable{\begin{tabular}{cc|ccc|ccc|ccc|ccc|ccc}
\toprule
\multicolumn{2}{c}{Setting} & \multicolumn{3}{c}{$s=4,\sigma=0.5$} & \multicolumn{3}{c}{$s=4,\sigma=2$} & \multicolumn{3}{c}{$s=8,\sigma=0.5$} & \multicolumn{3}{c}{$s=8,\sigma=2$} & \multicolumn{3}{c}{$s=16,\sigma=0.5$} \\
\midrule
Data & Method & \makecell[c]{PSNR$\uparrow$} & \makecell[c]{SSIM$\uparrow$} & \makecell[c]{NRMSE$\downarrow$} & \makecell[c]{PSNR$\uparrow$} & \makecell[c]{SSIM$\uparrow$} & \makecell[c]{NRMSE$\downarrow$} & \makecell[c]{PSNR$\uparrow$} & \makecell[c]{SSIM$\uparrow$} & \makecell[c]{NRMSE$\downarrow$} & \makecell[c]{PSNR$\uparrow$} & \makecell[c]{SSIM$\uparrow$} & \makecell[c]{NRMSE$\downarrow$} & \makecell[c]{PSNR$\uparrow$} & \makecell[c]{SSIM$\uparrow$} & \makecell[c]{NRMSE$\downarrow$} \\
\midrule
\multirow{6}{*}{\makecell[c]{WDC mall\\4 patches\\$320\times256\times191$}} & \textit{LR-HSI} & 26.03 & 0.716 & 0.174 & 25.52 & 0.665 & 0.185 & 23.74 & 0.549 & 0.227 & 23.69 & 0.539 & 0.228 & 22.43 & 0.479 & 0.264 \\
 & CS2DIPs\cite{fang2024cs2dips} & 41.39 & 0.988 & 0.030 & 28.57 & 0.822 & 0.130 & 40.40 & 0.981 & 0.033 & \underline{28.94} & \underline{0.830} & \underline{0.125} & \underline{37.83} & \underline{0.967} & \underline{0.045} \\
 & GTNN\cite{dian2024hyperspectral} & \textbf{45.91} & \textbf{0.995} & \textbf{0.018} & \textbf{29.98} & \textbf{0.876} & \textbf{0.111} & \textbf{42.63} & \textbf{0.990} & \textbf{0.026} & \textbf{29.84} & \textbf{0.870} & \textbf{0.112} & 34.60 & 0.946 & 0.065 \\
 & GNLTN\cite{xu2025nonlinear} & 34.35 & 0.943 & 0.067 & 28.71 & 0.831 & 0.128 & 30.68 & 0.877 & 0.102 & 27.37 & 0.760 & 0.149 & 29.08 & 0.848 & 0.123 \\
 & EDIP-Net\cite{li2025enhanced} & 33.28 & 0.930 & 0.076 & 27.95 & 0.796 & 0.140 & 31.88 & 0.909 & 0.089 & 27.75 & 0.782 & 0.143 & 29.78 & 0.869 & 0.113 \\
 & \textbf{DiffTR} & \underline{42.78} & \underline{0.991} & \underline{0.025} & \underline{28.93} & \underline{0.845} & \underline{0.125} & \underline{41.00} & \underline{0.985} & \underline{0.031} & 28.62 & \underline{0.830} & 0.129 & \textbf{39.37} & \textbf{0.980} & \textbf{0.038} \\
\midrule
\multirow{6}{*}{\makecell[c]{PaviaU\\2 patches\\$320\times320\times103$}} & \textit{LR-HSI} & 26.80 & 0.718 & 0.224 & 26.27 & 0.667 & 0.238 & 24.39 & 0.533 & 0.295 & 24.34 & 0.525 & 0.297 & 22.85 & 0.438 & 0.353 \\
 & CS2DIPs\cite{fang2024cs2dips} & 39.97 & 0.977 & 0.049 & 30.06 & 0.848 & 0.154 & 39.39 & 0.975 & \underline{0.052} & \underline{30.11} & \underline{0.850} & \underline{0.153} & 37.36 & 0.969 & 0.066 \\
 & GTNN\cite{dian2024hyperspectral} & \textbf{41.95} & \underline{0.981} & \textbf{0.039} & \underline{30.12} & \underline{0.851} & \underline{0.153} & \underline{40.88} & \underline{0.978} & \textbf{0.044} & 30.08 & 0.849 & \underline{0.153} & \underline{39.82} & \underline{0.976} & \underline{0.050} \\
 & GNLTN\cite{xu2025nonlinear} & 40.53 & 0.977 & 0.046 & 29.95 & 0.848 & 0.156 & 37.86 & 0.966 & 0.063 & 29.66 & 0.836 & 0.161 & 34.95 & 0.949 & 0.088 \\
 & EDIP-Net\cite{li2025enhanced} & 32.92 & 0.927 & 0.111 & 28.60 & 0.800 & 0.182 & 32.80 & 0.927 & 0.112 & 28.39 & 0.790 & 0.186 & 32.48 & 0.921 & 0.116 \\
 & \textbf{DiffTR} & \underline{41.77} & \textbf{0.982} & \underline{0.040} & \textbf{31.55} & \textbf{0.894} & \textbf{0.129} & \textbf{41.02} & \textbf{0.980} & \textbf{0.044} & \textbf{31.46} & \textbf{0.892} & \textbf{0.131} & \textbf{40.46} & \textbf{0.979} & \textbf{0.046} \\
\midrule
\multirow{6}{*}{\makecell[c]{CAVE\\4 images\\$512\times512\times31$}} & \textit{LR-HSI} & 34.87 & 0.954 & 0.095 & 34.28 & 0.945 & 0.102 & 30.59 & 0.894 & 0.154 & 30.54 & 0.892 & 0.155 & 27.25 & 0.842 & 0.226 \\
 & CS2DIPs\cite{fang2024cs2dips} & 42.75 & 0.979 & 0.037 & 37.93 & 0.961 & 0.066 & 42.26 & 0.966 & 0.040 & 38.86 & 0.950 & 0.060 & 39.68 & 0.970 & 0.053 \\
 & GTNN\cite{dian2024hyperspectral} & 49.40 & \underline{0.995} & \underline{0.017} & 39.88 & 0.975 & 0.054 & \textbf{47.95} & \textbf{0.994} & \textbf{0.021} & 39.75 & 0.973 & 0.054 & \textbf{46.13} & \textbf{0.992} & \textbf{0.026} \\
 & GNLTN\cite{xu2025nonlinear} & \underline{49.55} & \textbf{0.996} & \underline{0.017} & \underline{40.57} & \underline{0.981} & \underline{0.049} & \underline{47.24} & \underline{0.993} & \underline{0.023} & \underline{40.15} & \textbf{0.979} & \underline{0.052} & \underline{44.75} & \underline{0.990} & \underline{0.031} \\
 & EDIP-Net\cite{li2025enhanced} & 35.01 & 0.940 & 0.089 & 32.66 & 0.907 & 0.119 & 33.92 & 0.924 & 0.103 & 32.85 & 0.905 & 0.117 & 33.98 & 0.930 & 0.101 \\
 & \textbf{DiffTR} & \textbf{50.18} & \textbf{0.996} & \textbf{0.016} & \textbf{42.49} & \textbf{0.986} & \textbf{0.040} & 46.73 & \underline{0.993} & 0.025 & \textbf{40.86} & \underline{0.978} & \textbf{0.048} & 43.78 & 0.988 & 0.035 \\
\bottomrule
\end{tabular}}}
\end{table*}

Table~\ref{tab:superRes} covers all 50 combinations of scenes and degradation settings. DiffTR scores average three sampling runs per fixed final checkpoint and then average over scenes. Baseline optimization budgets match DiffTR's initialization and training time; its additional sampling and refinement costs enter the total reconstruction time in Table~\ref{tab:sr_resources}. Implementation and refinement-selection details appear in Appendix~\ref{app:sr_details}.
DiffTR achieves the highest PSNR in 8 of 15 dataset--setting comparisons. At $(s,\sigma)=(4,2)$, it exceeds the best baseline by 1.43~dB on PaviaU and 1.92~dB on CAVE; at $(16,0.5)$, it leads all three metrics on WDC and PaviaU. GTNN leads PSNR on the remaining WDC settings and on CAVE at $s=8,16$ with weak blur.

\begin{figure}[!h]
\centering
\includegraphics[width=\linewidth]{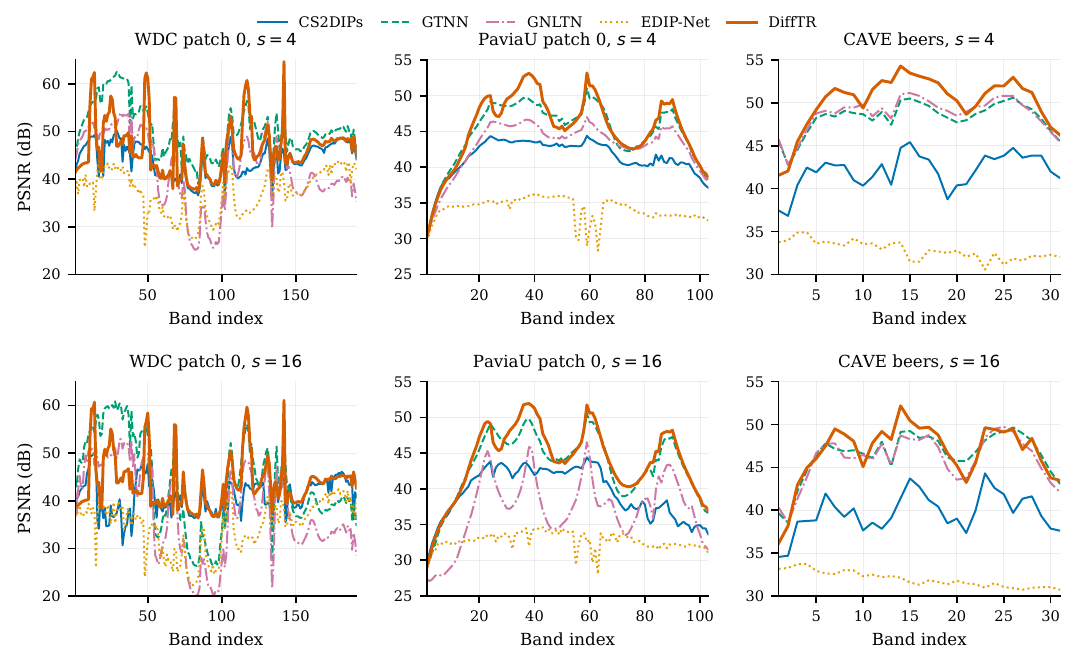}
\caption{\textbf{Bandwise PSNR comparison for HSI super-resolution}. Columns show WDC patch~0, PaviaU patch~0, and CAVE Beers. The top and bottom rows use $s=4$ and $s=16$, respectively, with $\sigma=0.5$. Unsmoothed curves follow the budget protocol in Table~\ref{tab:superRes}; DiffTR uses one reconstruction per setting.}
\label{fig:sr_bandwise}
\end{figure}

\begin{figure*}[!htbp]
\centering
\includegraphics[width=\textwidth]{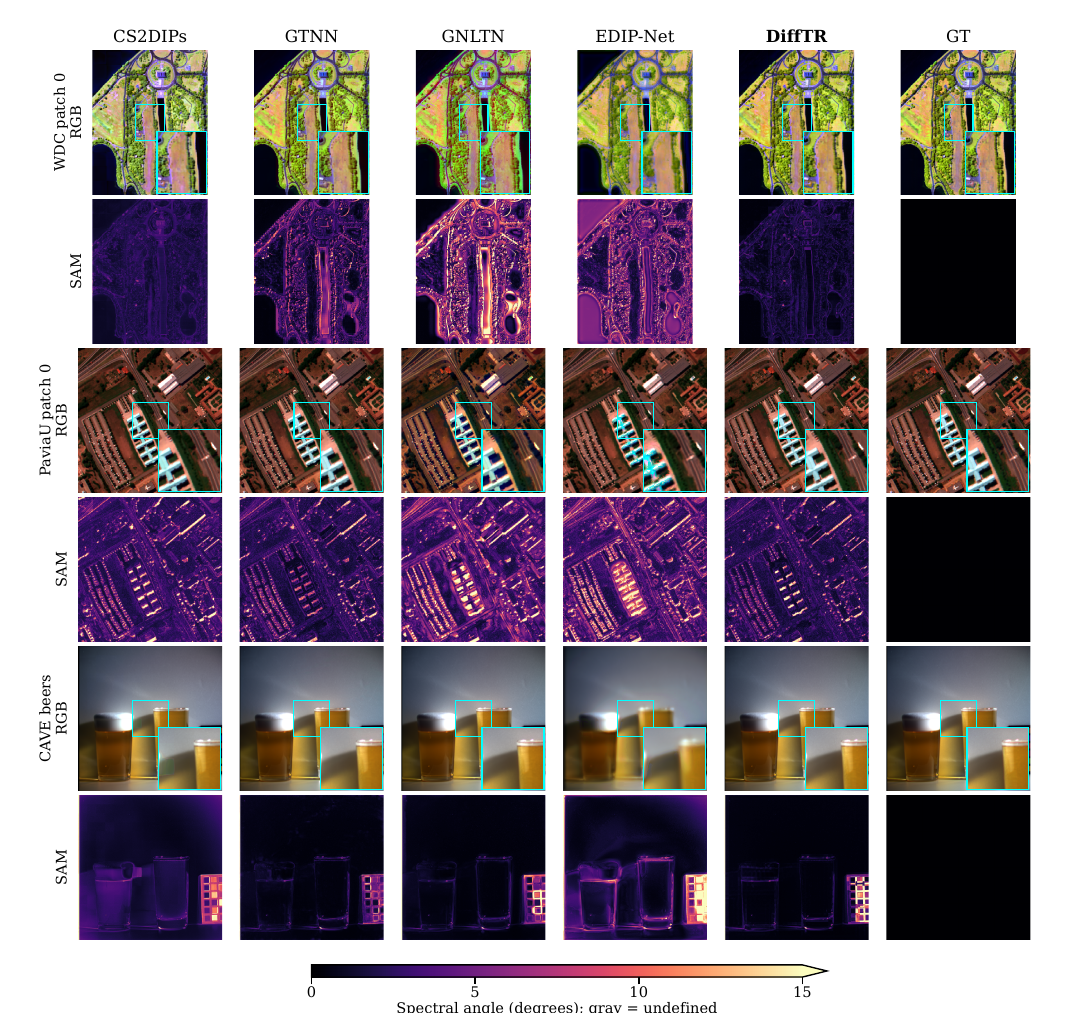}
\caption{\textbf{Qualitative comparison of HSI super-resolution}. RGB reconstructions, enlarged regions, and spectral angle maps are shown at $s=16$ and $\sigma=0.5$.}
\label{fig:sr_visual}
\end{figure*}

Figure~\ref{fig:sr_bandwise} shows that DiffTR matches or exceeds GTNN on most bands of PaviaU patch~0 and CAVE Beers at $s=4$, and avoids several pronounced PSNR drops observed for GTNN on WDC patch~0 at $s=16$. Figure~\ref{fig:sr_visual} further illustrates spatial boundary preservation at $s=16$, with fewer regions of large spectral-angle error than GNLTN and EDIP-Net on WDC and PaviaU. Together, these comparisons show that DiffTR recovers both spatial detail and spectral structure.

\begin{table*}[!h]
\centering
\caption{\textbf{Computational cost comparison for HSI super-resolution}. Parameter counts, runtime, and peak GPU memory follow the protocol in Table~\ref{tab:superRes}. Bold values indicate the lowest cost per dataset; n/a denotes an inapplicable parameter count.}
\label{tab:sr_resources}
{\small\setlength{\tabcolsep}{3pt}
\fitTable{\begin{tabular}{l|ccc|ccc|ccc|ccc}
\toprule
 & \multicolumn{3}{c|}{Parameters (M)} & \multicolumn{3}{c|}{$T_{\rm opt}$ (s)} & \multicolumn{3}{c|}{$T_{\rm rec}$ (s)} & \multicolumn{3}{c}{Peak GPU memory (GiB)} \\
Method & WDC & PaviaU & CAVE & WDC & PaviaU & CAVE & WDC & PaviaU & CAVE & WDC & PaviaU & CAVE \\
\midrule
CS2DIPs & 8.16--9.39 & 8.18--9.72 & 8.34--12.27 & 118.61 & 119.62 & 119.49 & 118.61 & 119.62 & 119.49 & 4.10 & 4.82 & 11.45 \\
GTNN & n/a & n/a & n/a & \textbf{92.16} & 117.99 & 117.18 & \textbf{92.16} & 117.99 & 117.18 & \textbf{0.28} & \textbf{0.24} & \textbf{0.42} \\
GNLTN & n/a & n/a & n/a & 116.32 & \textbf{115.77} & \textbf{111.33} & 116.32 & \textbf{115.77} & \textbf{111.33} & 0.33 & 0.31 & 0.68 \\
EDIP-Net & 14.06 & 12.92 & 11.98 & 118.53 & 119.51 & 119.30 & 118.53 & 119.51 & 119.30 & 6.81 & 8.09 & 19.40 \\
DiffTR & \textbf{7.82} & \textbf{4.65} & \textbf{3.12} & 118.67 & 119.67 & 119.65 & 135.00 & 133.00 & 127.68 & 0.97 & 0.66 & 0.54 \\
\bottomrule
\end{tabular}}}

\end{table*}

Table~\ref{tab:sr_resources} reports mean optimization time ($T_{\rm opt}$), mean total reconstruction time ($T_{\rm rec}$), and peak GPU memory for each dataset. DiffTR uses 3.12--7.82 million parameters and 0.54--0.97~GiB of GPU memory, both lower than CS2DIPs and EDIP-Net on every dataset. On CAVE, it uses 0.54~GiB, compared with 11.45~GiB and 19.40~GiB, respectively. Sampling and refinement add 8.03--16.33~s, giving total reconstruction times of 127.68--135.00~s. Under this optimization-budget protocol, DiffTR combines competitive reconstruction quality with low memory use among the neural methods, while GTNN and GNLTN achieve shorter total runtimes.

\section{Ablation and Discussion}
\label{sec:ablation}
We examine DiffTR's representation and task-specific observation models through denoising component ablations, noise-aware recovery, and studies of sampling depth, measurement guidance, and post-sampling refinement. Evaluation protocols are provided in Appendix~\ref{app:denoise_settings} for denoising and Appendix~\ref{app:sr_ablation} for super-resolution.

\subsection{Denoising Component Ablation}
\label{sec:denoise_component_ablation}
We evaluate coordinate conditioning, sparse-noise separation, and spatial consistency on top-left $128\times128$ crops of PaviaU and WDC, retaining 103 and 191 bands, respectively, after bandwise normalization of each full scene. Case~3 combines Gaussian noise ($\sigma=0.1$), sparse corruption (density $0.1$), and 12 dead columns per band. All variants use one-step diffusion with clean-sample prediction, rank 8, batch size 1,024, and 3,000 training updates with a cosine learning-rate schedule starting at $10^{-3}$. We vary one component at a time, match corruptions and training samples across variants, and evaluate final exponential moving average (EMA) weights. Table~\ref{tab:ablation_denoise} reports means and sample standard deviations over three runs.

\begin{table*}[t]
\centering
\caption{\textbf{Component ablation study for tensor denoising}. Entries report mean $\pm$ standard deviation of PSNR and SSIM on PaviaU and WDC crops under Case~3. Bold values indicate the highest mean in each column.}
\label{tab:ablation_denoise}
\small
\fitTable{\begin{tabular}{lcccc}
\toprule
& \multicolumn{2}{c}{PaviaU} & \multicolumn{2}{c}{WDC} \\
Variant & PSNR (dB) & SSIM & PSNR (dB) & SSIM \\
\midrule
Full implementation & $27.086\pm0.128$ & $0.6325\pm0.0085$ & $26.803\pm0.159$ & $0.7017\pm0.0061$ \\
Constant condition & $20.624\pm0.007$ & $0.4370\pm0.0002$ & $13.809\pm0.009$ & $0.3299\pm0.0003$ \\
Without sparse separation & $27.007\pm0.127$ & $0.6068\pm0.0076$ & $24.812\pm0.030$ & $0.5569\pm0.0022$ \\
Without spatial consistency & $27.110\pm0.123$ & $0.6346\pm0.0089$ & $\mathbf{26.851}\pm0.169$ & $\mathbf{0.7039}\pm0.0050$ \\
Stop-gradient target & $\mathbf{27.114}\pm0.120$ & $\mathbf{0.6352}\pm0.0078$ & $26.825\pm0.205$ & $0.7024\pm0.0071$ \\
\bottomrule
\end{tabular}
}
\end{table*}

Replacing all input coordinates with zero makes the condition $\gamma_\phi(\mathbf i)$ in Section~\ref{sec:shared_factor_diffusion} constant while preserving the architecture and parameter count. PSNR drops by 6.46~dB on PaviaU and 12.99~dB on WDC, highlighting the importance of coordinate conditioning for the visual tensor representation.
Disabling the sparse update in Eq.~\eqref{eq:sparse_noise_update} reduces PSNR by 0.08~dB on PaviaU and 1.99~dB on WDC, and SSIM by 0.0257 and 0.1448, respectively. Explicit separation improves mixed-noise recovery on both datasets, with a larger effect on WDC.
Setting $\lambda_D=0$ removes the spatial-consistency loss in Eq.~\eqref{equ:smoothReg} and increases PSNR by only 0.024~dB on PaviaU and 0.048~dB on WDC. Stopping gradients through the perturbed-coordinate branch while retaining the loss produces similarly small changes. These changes are smaller than the reported standard deviations, indicating limited sensitivity to spatial consistency and its gradient routing in this setting.

\subsection{Noise-Aware Denoising and Proximal Guidance}
\label{sec:noise_aware_ablation}
We compare two fitting procedures on matched $96\times96$ crops of WDC and PaviaU: BCD alternates sparse separation and model updates (Algorithm~\ref{algo:tensorDenoising}), whereas Likelihood fits a deterministic estimate using the noise model in Section~\ref{sec:denoising_method}. Proximal then refines the Likelihood output with fixed model parameters; $g$ controls the correction strength. Two-stage diffusion uses the fitted estimate as its training target and applies the correction during sampling. Each configuration uses one training run; Table~\ref{tab:denoise_repeat} reports an additional independent Case~2 run. Objectives and protocols are specified in Appendix~\ref{app:noise_aware_optimization}.

\begin{table*}[t]
\centering
\caption{\textbf{Comparison of BCD, likelihood fitting, and post-fit proximal refinement}. Likelihood denotes the deterministic initial estimate. Entries report PSNR/SSIM/NRMSE; bold indicates the best result for each row and metric, including ties.}
\label{tab:denoise_guidance}
\fitTable{
\begin{tabular}{lccccc}
\toprule
Data & Case & BCD & Likelihood & Proximal ($g=1$) & Proximal ($g=4$) \\
\midrule
WDC & 1 & 30.94/0.751/0.120 & \textbf{33.81}/\textbf{0.903}/\textbf{0.086} & 33.29/0.796/0.091 & 30.94/0.671/0.120 \\
WDC & 2 & 32.73/0.795/0.098 & 35.84/\textbf{0.932}/0.068 & \textbf{36.96}/0.927/\textbf{0.060} & 35.80/0.878/0.069 \\
WDC & 3 & 31.17/0.805/0.117 & 32.63/0.838/0.099 & \textbf{33.24}/\textbf{0.857}/\textbf{0.092} & 32.78/0.833/0.097 \\
PaviaU & 1 & 30.17/0.715/0.181 & \textbf{31.52}/\textbf{0.793}/\textbf{0.155} & 30.24/0.737/0.179 & 27.68/0.632/0.241 \\
PaviaU & 2 & 31.93/0.806/0.147 & 33.44/0.865/0.124 & \textbf{34.03}/\textbf{0.878}/\textbf{0.116} & 32.43/0.834/0.139 \\
PaviaU & 3 & \textbf{30.21}/0.778/\textbf{0.180} & 29.65/0.817/0.192 & 29.79/\textbf{0.824}/0.189 & 28.90/0.771/0.209 \\
\bottomrule
\end{tabular}
}
\end{table*}

\begin{table*}[t]
\centering
\caption{\textbf{Effect of proximal guidance in two-stage diffusion}. The initial estimate is obtained by likelihood fitting and provides fixed diffusion training targets. Entries report PSNR/SSIM/NRMSE; bold indicates the best result for each row and metric, including ties.}
\label{tab:denoise_staged}
\fitTable{
\begin{tabular}{lccccc}
\toprule
Data & Case & \makecell{Initial estimate\\(3,000 updates)} & DDIM ($g=0$) & DDIM ($g=1$) & DDIM ($g=4$) \\
\midrule
WDC & 1 & 33.03/\textbf{0.895}/\textbf{0.094} & 32.90/\textbf{0.895}/0.096 & \textbf{33.06}/0.792/\textbf{0.094} & 30.88/0.667/0.121 \\
WDC & 2 & 34.80/0.920/0.077 & 34.14/\textbf{0.922}/0.083 & \textbf{36.17}/0.919/\textbf{0.066} & 35.34/0.868/0.072 \\
WDC & 3 & 32.07/0.830/0.105 & 31.76/0.846/0.109 & \textbf{32.86}/\textbf{0.855}/\textbf{0.096} & 32.49/0.830/0.100 \\
PaviaU & 1 & \textbf{31.11}/\textbf{0.769}/\textbf{0.162} & 31.09/0.768/\textbf{0.162} & 30.24/0.733/0.179 & 27.72/0.631/0.239 \\
PaviaU & 2 & 32.97/0.849/0.131 & 32.94/0.848/0.131 & \textbf{33.89}/\textbf{0.874}/\textbf{0.118} & 32.42/0.832/0.139 \\
PaviaU & 3 & 29.24/0.795/0.201 & 29.25/0.794/0.201 & \textbf{29.56}/\textbf{0.815}/\textbf{0.194} & 28.74/0.765/0.213 \\
\bottomrule
\end{tabular}
}
\end{table*}

Table~\ref{tab:denoise_guidance} shows that likelihood fitting improves PSNR over BCD in five of six cases. Post-fit proximal refinement with $g=1$ further raises WDC Case~2 PSNR from 35.84 to 36.96~dB, while the initial estimate remains best in Case~1 on both datasets. In two-stage diffusion (Table~\ref{tab:denoise_staged}), changing $g$ from 0 to 1 raises WDC Case~2 PSNR from 34.14 to 36.17~dB. Increasing $g$ to 4 reduces PSNR throughout both tables, showing that stronger proximal correction is not consistently beneficial.

\begin{table*}[t]
\centering
\caption{\textbf{Controls for fitting loss and diffusion training targets}. MSE and Huber directly fit the coordinate-conditioned network to observations; noisy-target DDIM trains diffusion on those observations. Entries report PSNR/SSIM/NRMSE; bold indicates the best result for each row and metric, including ties.}
\label{tab:denoise_observation_controls}
\fitTable{
\begin{tabular}{lcccccc}
\toprule
Data & Case & MSE & Huber & Noisy-target DDIM ($g=0$) & Noisy-target DDIM ($g=1$) & Noisy-target DDIM ($g=4$) \\
\midrule
WDC & 1 & 24.53/0.565/0.251 & \textbf{32.60}/\textbf{0.780}/\textbf{0.099} & 16.45/0.076/0.636 & 17.11/0.112/0.589 & 17.28/0.124/0.578 \\
WDC & 2 & 26.55/0.603/0.199 & \textbf{34.61}/\textbf{0.818}/\textbf{0.079} & 16.85/0.107/0.607 & 17.74/0.160/0.548 & 18.03/0.195/0.530 \\
\bottomrule
\end{tabular}
}
\end{table*}

\begin{table*}[t]
\centering
\caption{\textbf{Additional quantitative comparison under Case~2}. An additional run compares likelihood fitting and proximal refinement. Entries report PSNR/SSIM/NRMSE; bold values indicate the best result for each row and metric, including ties.}
\label{tab:denoise_repeat}
\fitTable{
\begin{tabular}{lccccc}
\toprule
Data & Case & BCD & Likelihood & Proximal ($g=1$) & Proximal ($g=4$) \\
\midrule
WDC & 2 & 32.24/0.788/0.103 & 35.42/0.924/0.072 & \textbf{36.72}/\textbf{0.925}/\textbf{0.062} & 35.71/0.877/0.069 \\
PaviaU & 2 & 32.01/0.809/0.146 & 33.53/0.869/0.123 & \textbf{34.01}/\textbf{0.879}/\textbf{0.116} & 32.40/0.834/0.140 \\
\bottomrule
\end{tabular}
}
\end{table*}

\begin{table*}[h]
	\centering
	\caption{\textbf{Ablation study of post-sampling refinement}. Comparisons cover data consistency, spectral subspace regularization, and the reference reconstruction used in the proximity term. Entries report PSNR/SSIM/NRMSE; bold values indicate the best results.}
	\label{tab:sr_subspace_ablation}
	\fitTable{\begin{tabular}{lcc|cccc}
\toprule
Scene & $s$ & $\sigma$ & DC-10 & DC-100 & Subspace & NN + subspace \\
\midrule
WDC patch 0 & 4 & 0.5 & 40.573/0.9853/0.0342 & 40.761/0.9860/0.0334 & \textbf{42.731}/\textbf{0.9883}/\textbf{0.0266} & 32.411/0.9397/0.0874 \\
PaviaU patch 0 & 4 & 0.5 & 41.229/0.9796/0.0422 & 41.887/0.9811/0.0391 & \textbf{41.960}/\textbf{0.9822}/\textbf{0.0388} & 41.674/0.9816/0.0401 \\
CAVE beers & 4 & 0.5 & 47.993/0.9942/0.0124 & \textbf{48.924}/\textbf{0.9950}/\textbf{0.0112} & 48.602/0.9949/0.0116 & 48.666/0.9949/0.0115 \\
WDC patch 0 & 4 & 2 & 29.134/0.8374/0.1275 & 29.482/0.8520/0.1225 & \textbf{29.592}/\textbf{0.8562}/\textbf{0.1209} & 28.207/0.8157/0.1419 \\
PaviaU patch 0 & 4 & 2 & 29.519/0.8141/0.1625 & 30.481/0.8509/0.1455 & \textbf{30.570}/\textbf{0.8550}/\textbf{0.1440} & 29.141/0.8149/0.1697 \\
CAVE beers & 16 & 0.5 & 40.238/0.9834/0.0303 & 42.245/0.9882/0.0241 & 44.857/\textbf{0.9922}/\textbf{0.0178} & \textbf{44.859}/\textbf{0.9922}/\textbf{0.0178} \\
\bottomrule
\end{tabular}
}
\end{table*}

\begin{table}[h]
	\centering
	\caption{\textbf{Sensitivity analysis of spectral subspace regularization}. Entries report paired PSNR changes (dB) relative to the subspace configuration when varying the spectral penalty or retained energy. Positive values indicate improvement.}
	\label{tab:sr_subspace_sensitivity}
	\fontsize{10}{12}\selectfont
	\fitTable{\begin{tabular}{lcc|rrc}
\toprule
Scene & $s$ & $\sigma$ & $\gamma=1$ & $\gamma=100$ & Energy $99.99\%$ \\
\midrule
WDC patch 0 & 4 & 0.5 & -1.374 & -0.572 & +1.094 \\
PaviaU patch 0 & 4 & 0.5 & +0.317 & -0.963 & +0.485 \\
CAVE beers & 4 & 0.5 & +0.758 & -4.053 & +0.731 \\
WDC patch 0 & 4 & 2 & -0.058 & -0.075 & +0.018 \\
PaviaU patch 0 & 4 & 2 & -0.019 & -0.089 & -0.037 \\
CAVE beers & 16 & 0.5 & -0.316 & -2.184 & -1.496 \\
\bottomrule
\end{tabular}
}
\end{table}

Table~\ref{tab:denoise_observation_controls} compares MSE and Huber fitting with diffusion trained directly on noisy observations. Huber fitting performs best on both WDC cases, highlighting the importance of handling corruption in the fitting objective and training targets. The additional Case~2 run in Table~\ref{tab:denoise_repeat} reproduces the gains from likelihood fitting over BCD and from proximal refinement with $g=1$ over the initial estimate on both datasets.

\subsection{Sampling Depth}
\label{sec:sampling_depth}
We examine the reverse-sampling stage of Algorithm~\ref{algo:superResolution} using deterministic DDIM updates with the same measurement correction and no post-sampling refinement. Fixed checkpoints are evaluated on PaviaU patch~0 ($320\times320\times103$) and WDC patch~0 ($320\times256\times191$) at $s=4$ and $\sigma=0.5$, with blur also applied to the HR-MSI. Each setting uses three initial-noise seeds. Varying the number of DDIM steps at fixed weights isolates inference depth from model training.

\begin{figure}[t]
\centering
\includegraphics[width=\linewidth]{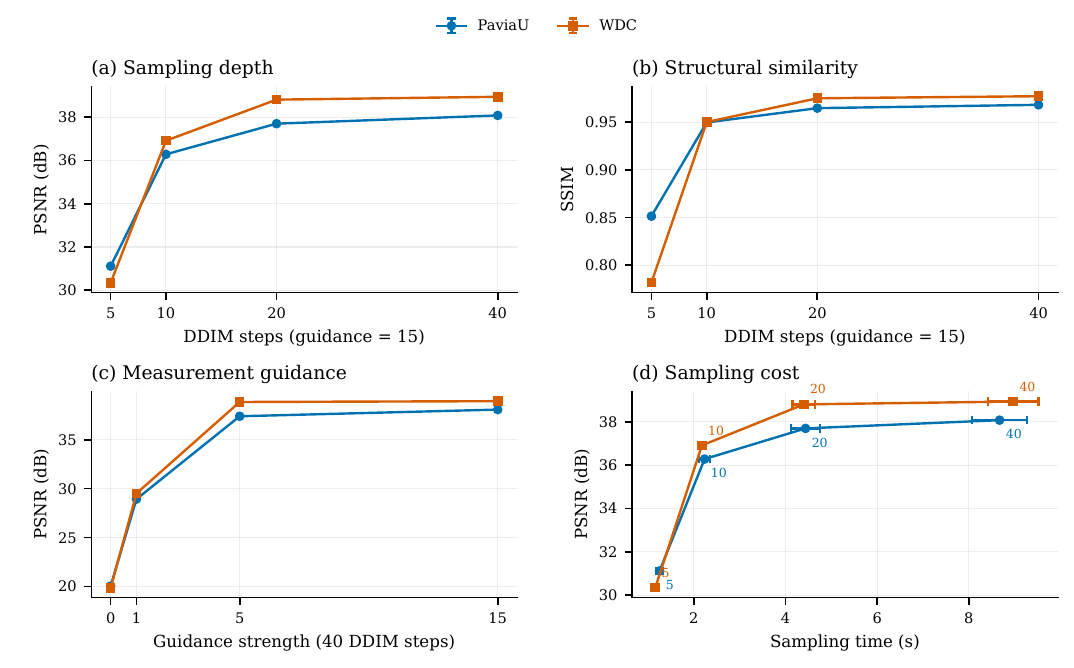}
\caption{\textbf{Ablation study of DDIM sampling}. PaviaU and WDC patch~0 are evaluated at $s=4$, $\sigma=0.5$, and $\eta=0$, without post-sampling refinement. Panels (a,b) vary sampling depth, (c) varies guidance strength, and (d) compares PSNR with RTX~3090 sampling time. Error bars indicate sample standard deviations over three runs.}
\label{fig:sr_sampling_ablation}
\end{figure}

Figure~\ref{fig:sr_sampling_ablation}(a) shows that, with $g=15$, increasing the number of reverse steps from 5 to 20 raises PSNR from 31.11 to 37.70~dB on PaviaU and from 30.35 to 38.81~dB on WDC. Increasing from 20 to 40 steps adds only 0.38 and 0.13~dB, while mean RTX~3090 sampling time rises from 4.44 to 8.67~s and from 4.40 to 8.97~s, respectively (Figure~\ref{fig:sr_sampling_ablation}(d)). Thus, 20 steps offer a useful balance between speed and quality for these DDIM reconstructions, while 40 steps yield the highest tested PSNR.

\subsection{Measurement Guidance}
We vary $g$ in the guidance schedule $\zeta_k=g\bar\alpha_k$ of Algorithm~\ref{algo:superResolution}, using the same checkpoints and observations as Section~\ref{sec:sampling_depth}, 40 DDIM steps, and no post-sampling refinement. Setting $g=0$ removes the HR-MSI likelihood correction while retaining the LR-HSI-trained model. Figure~\ref{fig:sr_sampling_ablation}(c) shows that increasing $g$ from 0 to 15 raises PSNR from 20.03 to 38.08~dB on PaviaU and from 19.84 to 38.94~dB on WDC. The gains from $g=5$ to $g=15$ are only 0.68 and 0.08~dB, respectively, indicating diminishing returns from stronger measurement guidance.

\subsection{Data Consistency and Spectral Subspace Regularization}
\label{sec:sr_terminal_ablation}
Post-sampling refinement optimizes the reconstruction after diffusion sampling using Eq.~\eqref{eq:sr_terminal_proximal}, with all learned parameters fixed. DC-10, DC-100, and Subspace use $(\tau,\gamma)=(10,0)$, $(100,0)$, and $(100,10)$, respectively, on the same diffusion outputs. NN + subspace instead uses nearest-neighbor-upsampled LR-HSI as the reference in the proximity term. The spectral-penalty comparison fixes $\tau=100$; the main results use $\tau=1000$. Other settings are given in Appendix~\ref{app:sr_ablation}.

In Table~\ref{tab:sr_subspace_ablation}, Subspace improves PSNR over DC-100 in five of six configurations, including gains of 1.97~dB on WDC with weak blur and 2.61~dB on CAVE Beers at $s=16$, with a 0.32~dB decrease on Beers at $s=4$. Using the diffusion estimate rather than upsampled LR-HSI as the reference adds 10.32~dB on WDC with weak blur, while the CAVE results differ by less than 0.07~dB.
Table~\ref{tab:sr_subspace_sensitivity} shows that increasing the spectral penalty or retained energy does not consistently improve PSNR. The benefits of both spectral subspace regularization and the diffusion reference therefore depend on the scene and degradation.

The ablations demonstrate how the shared representation accommodates coordinate information, corruption models, spatial constraints, and spectral measurements. Their task-dependent effects motivate task-specific observation models within a common recovery framework.

\FloatBarrier
\section{Conclusion}
We introduced DiffTR, a unified tensor recovery framework built on shared-factor-conditioned score learning. The same representation supports scalar entries and vector-valued tubes, while task-specific observation models extend recovery from completion to mixed-noise denoising and HSI super-resolution. Likelihood fitting and proximal correction accommodate corruption models, and optional spatial consistency provides a coordinate-based alternative to TV. Experiments across tabular tensors, images, and videos demonstrate the framework's broad applicability, competitive recovery quality, and efficient neural representation. This separation of representation learning and observation modeling provides a foundation for extending DiffTR to additional tensor structures and inverse problems.
\section*{Acknowledgments}
This research was supported by the National Science and Technology Major Project (2025ZD1401102), and the Innovation Foundation for Doctoral Dissertation of Northwestern Polytechnical University (CX2025074).

\FloatBarrier
\bibliographystyle{IEEEtran}
\bibliography{reference} 
\vfill

\appendices
\renewcommand{\thesection}{\Alph{section}}
\renewcommand{\thesubsection}{\thesection.\arabic{subsection}}
\onecolumn
\setcounter{table}{0}
\section{General Implementation Settings}
\label{app:implementation}
Experiments use Ubuntu 20.04, two Intel Xeon Silver 4214 CPUs, NVIDIA RTX~3090 GPUs, and 128~GB of RAM, with PyTorch 2.1 (2.4.1 for HSI super-resolution). Table~\ref{tab:implementation} collects training settings for the main comparisons and denoising ablations. Each embedding or score subnetwork is a two-layer multilayer perceptron (MLP) with equal hidden widths unless specified otherwise. Task-specific optimization and evaluation protocols follow in Appendix~\ref{app:noise_aware_optimization} and Appendix~\ref{app:sr_details}.
\begin{table*}[h]
	\centering
	\small
	\caption{\textbf{Implementation settings for tensor recovery}. $K$: diffusion horizon; $R$: factor dimension; LR: initial learning rate. A dash denotes no diffusion horizon.}
	\label{tab:implementation}
	\fitTable{\begin{tabular}{cccccccc}
		\toprule
		Task &Dataset &Iterations &Batch &$K$ &Hidden widths &LR &$R$ \\
		\midrule
		\multirow{2}{*}{Sparse tensor completion}
		&Alog &40000 &256 &1 &[256, 256] &5e-3 &\{3,5,8,10\} \\
		&ACC &10000 &8192 &3 &[128, 128] &1e-2 &\{3,5,8,10\} \\
		\midrule
		\multirow{2}{*}{Continuous-time tensor completion}
		&Air &40000 &1024  &1 &[64, 64] &1e-2 &\{3,5,8,10\} \\
		&Click &200 &2048  &1 &[64, 64] &1e-2 &\{3,5,8,10\} \\
		\midrule
		\multirow{3}{*}{\makecell[c]{Visual data inpainting\\\textit{Entry-wise sampling}}}
		&RGB &20000 &256  &1 &[512, 512] &1e-3 &64 \\
		&MSI &40000 &256  &1 &[64, 64] &1e-3 &32 \\
		&Video &40000 &256 &1 &[512, 512] &1e-3 &32 \\
		\midrule
		\multirow{3}{*}{\makecell[c]{Visual data inpainting\\\textit{Tube-wise sampling}}}
		&RGB &20000 &256  &1 &[512, 512] &1e-3 &64 \\
		&MSI &40000 &256  &10 &[64, 64] &1e-3 &128 \\
		&Video &40000 &512 &1 &[512, 512] &2e-3 &64 \\
		\midrule
		Denoising (BCD, main) &\makecell{WDC/PaviaU\\CAVE} &40000 &8192 &1 &[128, 128] &1e-3 &8/16\\
		Denoising (likelihood, main) &WDC/PaviaU &10000 &8192 &--- &[128, 128] &1e-3 &8/16\\
		Denoising (component ablations) &WDC/PaviaU &3000 &1024 &1 &[128, 128] &1e-3 &8\\
		Denoising (BCD, crops) &WDC/PaviaU &4000 &2048 &1 &[128, 128] &1e-3 &8\\
		Denoising (likelihood, crops) &WDC/PaviaU &4000 &2048 &--- &[128, 128] &1e-3 &8\\
		Denoising (two-stage, crops) &WDC/PaviaU &3000+1000 &2048 &100 &[128, 128] &1e-3 &8\\
		\midrule
		\multirow{2}{*}{\makecell[c]{Hyperspectral\\super-resolution}}
		&CAVE &10000 &256 &10 &[512, 512] &1e-3 &128\\
		&WDC/PaviaU &10000 &256 &40 &[512, 512] &2e-4 &128\\
		\bottomrule
	\end{tabular}}
\end{table*}

\section{Denoising: Optimization and Evaluation}
\label{app:noise_aware_optimization}
BCD and Likelihood provide alternative fitting procedures; Proximal refines a fitted estimate. For an observed voxel $y_{\mathbf i}\in[0,1]$ with band index $b$, let $F_\psi(v,\mathbf i,k)$ predict its clean value in normalized $[-1,1]$ units, with $\psi$ collecting network and coordinate-factor parameters. The corresponding score is $s_\psi=(\sqrt{\bar\alpha_k}F_\psi-v)/(1-\bar\alpha_k)$. We write $Q_k(u,\epsilon)=\sqrt{\bar\alpha_k}u+\sqrt{1-\bar\alpha_k}\epsilon$ for forward perturbation and $\Pi$ for clipping to $[0,1]$.

\subsection{BCD: Sparse-Noise Separation and Model Fitting}
The BCD implementation of Algorithm~\ref{algo:tensorDenoising} refreshes the sparse estimate on each sampled mini-batch. With $y'_{\mathbf i}=2y_{\mathbf i}-1$, it first predicts $\widetilde u_{\mathbf i}=F_{\psi^{(t)}}(Q_k(y'_{\mathbf i},\epsilon),\mathbf i,k)$ without tracking gradients, then computes
\begin{equation}
\begin{aligned}
s_{\mathbf i}&=\arg\min_s\left\{\tfrac12(y'_{\mathbf i}-\widetilde u_{\mathbf i}-s)^2+\delta|s|\right\}
=\operatorname{Soft}_{\delta}(y'_{\mathbf i}-\widetilde u_{\mathbf i}),\\
u_{\mathbf i}&=y'_{\mathbf i}-s_{\mathbf i},\qquad v_{\mathbf i}=Q_k(u_{\mathbf i},\epsilon).
\end{aligned}
\label{eq:app_bcd_sparse}
\end{equation}
Here, $\delta=0.04$ corresponds to $\lambda/2$ in Eq.~\eqref{eq:sparse_noise_update}. Holding $u_{\mathbf i}$ fixed, one Adam step updates $\psi$ using the clean-sample prediction loss
\begin{equation}
\mathcal L_{\mathrm{BCD}}=\mathbb E\!\Big[w_k\Big\{\big(F_\psi(v_{\mathbf i},\mathbf i,k)-u_{\mathbf i}\big)^2
+\kappa\big(F_\psi(v_{\mathbf i},\mathbf i',k)-F_\psi(v_{\mathbf i},\mathbf i,k)\big)^2\Big\}\Big],
\label{eq:app_bcd_loss}
\end{equation}
where $w_k=\bar\alpha_k/(1-\bar\alpha_k)$ and $\kappa=0.01$. The perturbed coordinate $\mathbf i'$ retains the band index; spatial offsets are rounded unit-standard-deviation Gaussian draws clipped to image bounds. Both branches share $v_{\mathbf i}$. BCD uses one diffusion level and reuses $\epsilon$ within each update. Checkpoint and evaluation rules are specified in Appendix~\ref{app:denoise_settings}.

\subsection{Likelihood: Fitting a Deterministic Estimate}
The deterministic prediction is $f_\psi(\mathbf i)=[F_\psi(0,\mathbf i,0)+1]/2$, using the implementation's zero time index. For band $b$, we model Gaussian inliers with scale $\sigma_b$ and outliers with probability $\pi_b$. Clipping gives the likelihood contribution
\begin{equation}
\begin{aligned}
h_{\sigma_b}(y\mid x)&=\begin{cases}
\Phi(-x/\sigma_b),&y=0,\\
\sigma_b^{-1}\varphi((y-x)/\sigma_b),&0<y<1,\\
\Phi((x-1)/\sigma_b),&y=1,
\end{cases}\\
\ell_b(x;y)&=-\log\!\left[(1-\pi_b)h_{\sigma_b}(y\mid x)+\pi_b c(y)\right].
\end{aligned}
\label{eq:app_censored_likelihood}
\end{equation}
Here, $\varphi$ and $\Phi$ are the standard Gaussian density and distribution functions. Clipping a uniform distribution on $[-0.5,1.5]$ gives the outlier density $c(y)=1/2$ in the interior and mass $1/4$ at each endpoint. This mixture approximates the experimental corruption.

Let $m_{\mathbf i}=0$ for a spatial column whose observed values are at least 95\% zero in band $b$, and $m_{\mathbf i}=1$ otherwise. This mask is inferred from observations. For a mini-batch $\mathcal B$, fitting minimizes
\begin{equation}
\mathcal L_{\mathrm{like}}=\frac{\sum_{\mathbf i\in\mathcal B}m_{\mathbf i}\ell_b(f_\psi(\mathbf i);y_{\mathbf i})}{\max(1,\sum_{\mathbf i\in\mathcal B}m_{\mathbf i})}+\frac{10}{|\mathcal B|}\sum_{\mathbf i\in\mathcal B}\big[f_\psi(\mathbf i)-\Pi(f_\psi(\mathbf i))\big]^2.
\label{eq:app_likelihood_fit}
\end{equation}
Adam optimizes $\psi$, $\log\sigma_b$, and the logits of $\pi_b$, with the resulting noise scales and probabilities constrained to $[0.02,0.5]$ and $[0.001,0.4]$. Scales start from the median absolute deviation of valid $2\times2$ Haar differences divided by $0.67448975$, clipped to $[0.03,0.4]$; $\pi_b$ starts at $0.05$. Noise-parameter updates begin after 500 steps at learning rate $2\times10^{-3}$. The clipped final EMA prediction is the Likelihood output. MSE and Huber controls replace $\ell_b$ with squared error and Huber loss (threshold $0.02$), retaining the mask and range penalty.

\subsection{Proximal Refinement and Two-Stage Diffusion}
Let $H\in\mathbb R^{N\times I_C}$ contain the reference spectra of $N$ pixels and $V\in\mathbb R^{I_C\times r}$ contain the leading $r=8$ orthonormal eigenvectors of the centered covariance of the Likelihood output. Corrections take the form $CV^{\mathsf T}$, where $C\in\mathbb R^{N\times r}$. With the network, noise parameters, and basis fixed, the refinement objective is
\begin{equation}
\mathcal E_g(C;H)=\tfrac12\|C\|_F^2+g\sum_{p=1}^{N}\sum_{b=1}^{I_C}m_{pb}\sigma_b^2\ell_b\!\left((H+CV^{\mathsf T})_{pb};y_{pb}\right).
\label{eq:app_denoise_proximal}
\end{equation}
The quadratic term limits the change from $H$, and $\sigma_b^2$ scales the likelihood contribution of each band. We approximate the solution with ten gradient steps:
\begin{equation}
C^{(0)}=0,\qquad C^{(j+1)}=C^{(j)}-\frac{0.2}{1+g}\nabla_C\mathcal E_g(C^{(j)};H),\qquad j=0,\ldots,9.
\label{eq:app_denoise_proximal_update}
\end{equation}
The output is $\Pi(H+C^{(10)}V^{\mathsf T})$. The post-fit Proximal variant uses the Likelihood output as $H$; $g=0$ leaves it unchanged.

Two-stage diffusion trains on frozen Likelihood predictions, starting from the fitted EMA weights with a new optimizer and EMA. The clean-sample prediction loss is weighted by $\min(w_k,5)$. During 20-step deterministic DDIM sampling, $H$ is the current clean estimate and the correction strength is $g\bar\alpha_k$. The corrected estimate is clipped, mapped to $[-1,1]$, and inserted into the DDIM update while retaining the preceding noise prediction; the final step returns the corrected estimate. The noisy-target control trains diffusion and estimates its spectral basis directly from observations, keeping noise parameters at their initial values.

\subsection{Experimental Protocols}
\label{app:denoise_settings}
Training budgets, batch sizes, and ranks are listed in Table~\ref{tab:implementation}. In the main comparison (Table~\ref{tab:compareDenoise}), $R=16$ for WDC Case~3 and $R=8$ otherwise. Likelihood with proximal refinement is used for WDC Case~1 ($g=1$), WDC Cases~2--3, and PaviaU Case~2 ($g=4$); other entries use BCD. BCD checkpoints are evaluated every 1,000 updates and selected by clean-reference PSNR. Configurations are selected by dataset-average PSNR within each noise case, with all metrics computed from the same reconstructions.

The component study (Table~\ref{tab:ablation_denoise}) uses $128\times128$ crops and Case~3 dead columns. Corruptions are regenerated across three runs and matched across variants; entries report means and sample standard deviations from final EMA models.

The fitting and guidance comparisons (Tables~\ref{tab:denoise_guidance}--\ref{tab:denoise_repeat}) use $96\times96$ crops with per-voxel zero masks. Network parameters use Adam at initial learning rate $10^{-3}$, cosine decay to $10^{-6}$, and infinity-norm gradient clipping at two. Each entry reports a single training run evaluated with final EMA weights; Table~\ref{tab:denoise_repeat} adds an independent Case~2 run. For two-stage training, the two update counts in Table~\ref{tab:implementation} refer to likelihood fitting and diffusion training, respectively.

\section{HSI Super-Resolution: Refinement and Evaluation}
\label{app:sr_details}

\subsection{Data and Observation Operators}
WDC uses four $320\times256\times191$ patches starting at rows 0, 320, 640, and 960 and columns 0--255. PaviaU uses two $320\times320\times103$ patches starting at rows 0 and 290 and columns 0--319, with a 30-row overlap. CAVE uses the full $512\times512\times31$ Beers, Clay, Flowers, and Balloons images. WDC/PaviaU are normalized bandwise by full-scene extrema; CAVE is divided by 65535. The five degradation settings yield 20 WDC, 10 PaviaU, and 20 CAVE combinations.

Spatial degradation uses a separable $7\times7$ Gaussian kernel with replicate padding and non-overlapping $s\times s$ block averaging. The MSI receives the same blur, without added measurement noise. PaviaU uses zero-based columns 4--106 of the IKONOS-like response, normalized per output channel; CAVE uses the three-channel Nikon D700 response. WDC uses the transposed $191\times8$ \emph{SRF\_matrix.mat} from \href{https://arxiv.org/abs/2603.11530}{TenFuse}, preserving weights and band order. All output-channel weights sum to one; correspondence with the stored HSI band order is assumed because wavelength metadata are unavailable. Earlier WDC sensitivity experiments with five-channel responses use a separate protocol.

\subsection{Post-Sampling Refinement}
\label{app:sr_refinement}
The basis $B$ contains the smallest set of leading eigenvectors of $\mathbf L_{\mathrm{sp}}^{\mathsf T}\mathbf L_{\mathrm{sp}}$ retaining 99.9\% of its eigenvalue sum, where rows of $\mathbf L_{\mathrm{sp}}$ are uncentered LR-HSI spectra. The normalization $b_q=\max(A_q^*A_q\mathbf1)$ bounds $\|A_q\|_2^2$ for these nonnegative observation operators. Here, $\mathbf1$ is an all-ones tensor and $A_q^*$ is the adjoint, including the transpose of replicate padding.

Define $[\mathcal P_\perp(\mathcal U)]_{\mathrm{sp}}=\mathbf U_{\mathrm{sp}}(I-BB^{\mathsf T})$. This orthogonal projection is self-adjoint and idempotent. Setting the gradient of Eq.~\eqref{eq:sr_terminal_proximal} to zero gives
\begin{equation}
\begin{aligned}
\mathcal H&=I+\tau\sum_{q\in\{L,M\}}\frac{A_q^*A_q}{b_q}+\gamma\mathcal P_\perp,\\
\mathcal H(\mathcal U)&=\widehat{\mathcal X}+\tau\sum_{q\in\{L,M\}}\frac{A_q^*\mathcal Y_q}{b_q}.
\end{aligned}
\label{eq:sr_refinement_system}
\end{equation}
For $\tau,\gamma\geq0$, the identity term makes $\mathcal H$ positive definite, so the minimizer is unique. To reduce cancellation in single precision, we solve for the correction $\Delta=\mathcal U-\widehat{\mathcal X}$:
\begin{equation}
\begin{aligned}
\mathcal H(\Delta)&=\mathcal R_0,\\
\mathcal R_0&=\tau\sum_{q\in\{L,M\}}\frac{A_q^*(\mathcal Y_q-A_q\widehat{\mathcal X})}{b_q}-\gamma\mathcal P_\perp(\widehat{\mathcal X}).
\end{aligned}
\label{eq:sr_refinement_correction}
\end{equation}
Conjugate gradients start at $\Delta=0$ and apply $\mathcal H$ without forming a matrix. The main comparison uses $(\tau,\gamma)=(1000,10)$, at most 400 iterations, and stopping rule $\|\mathcal R_0-\mathcal H(\Delta)\|_F\leq10^{-5}\|\mathcal R_0\|_F$. The true residual is recomputed every 25 iterations and before convergence is accepted, with restarts as needed. All 150 reconstructions meet this tolerance. The output is $\operatorname{clip}_{[0,1]}(\widehat{\mathcal X}+\Delta)$.

\subsection{Training, Selection, and Evaluation}
Training follows Table~\ref{tab:implementation}, with latent widths 256 for WDC and 128 for PaviaU/CAVE and cosine learning-rate decay. Targets are nearest-neighbor-upsampled LR-HSI values. DDPM sampling uses final non-EMA weights and $\zeta_k=15\bar\alpha_k$. Network fitting and checkpoint selection use no clean HR-HSI. Three reconstructions per checkpoint vary only the sampling seed, measuring sampling variability at fixed trained weights.

Refinement settings were selected by mean PSNR on eight diagnostic combinations: WDC/PaviaU patch~0 and CAVE Beers at $s\in\{4,16\}$, $\sigma=0.5$, plus WDC/PaviaU patch~0 at $s=4$, $\sigma=2$. Seven candidates covered sampling corrections, spatial regularization, residual reinjection, and data consistency. Checkpoints and sampling noise were fixed during selection. The chosen settings replaced the earlier $(\tau,\gamma)=(100,10)$ configuration and were applied to the remaining 42 combinations, some sharing scenes with the diagnostic set. Thus, refinement selection is benchmark-informed; all 50 combinations enter the final comparison.

PSNR uses whole-cube MSE and unit data range; SSIM averages bandwise scores with a $7\times7$ window; NRMSE is $\|\widehat{\mathcal X}-\mathcal X\|_F/\|\mathcal X\|_F$. Scores are averaged over reconstructions and then scenes. Figures~\ref{fig:sr_bandwise} and~\ref{fig:sr_visual} use the same fixed DiffTR reconstruction and budget-matched baselines. One-based RGB bands are 101/61/31 (WDC), 61/31/11 (PaviaU), and 31/17/6 (CAVE). Display limits use each reference channel's 1st--99th percentiles for all methods; spectral angles span $0$--$15^\circ$, with undefined values in gray. Display normalization does not affect metrics.

\subsection{Ablation Protocols}
\label{app:sr_ablation}
The sampling-depth and guidance studies in Figure~\ref{fig:sr_sampling_ablation} use the fixed checkpoints and observations described in Section~\ref{sec:sampling_depth}, three matched initial-noise seeds, and deterministic DDIM without post-sampling refinement. Only the reverse-step count or guidance strength changes between variants.

Post-sampling refinement ablations use WDC/PaviaU patch~0 at $s=4$, $\sigma\in\{0.5,2\}$, and CAVE Beers at $s\in\{4,16\}$, $\sigma=0.5$. DDPM step counts follow Table~\ref{tab:implementation}, with $g=15$. The DC-10, DC-100, and Subspace settings are defined in Section~\ref{sec:sr_terminal_ablation}; their CG limits are 20, 100, and 100 iterations, respectively, with the stopping rule above. The 99.9\% energy threshold gives a three-dimensional spectral basis in these cases. Table~\ref{tab:sr_subspace_ablation} averages three reconstructions per diffusion-based variant and evaluates the deterministic NN + subspace control once. Table~\ref{tab:sr_subspace_sensitivity} varies one parameter at a time using a single shared diffusion sample and reports paired PSNR differences.

\subsection{Baseline Budgets and Resource Measurement}
Baselines receive DiffTR's measured initialization-and-training time for each scene and setting. CS2DIPs selects the lowest observed-data Huber loss within 5,000 updates; GTNN and GNLTN allow up to 100 and 50 iterations. Their implementations use Lloyd clustering, bicubic initialization, matched degradation operators, and CPU LAPACK SVD for selected operations. EDIP-Net allocates $2/9$ of its budget to stage two (at most 4,000 updates) and the remainder to stage three; learning rate 0.004 is constant for half of each phase, then decays linearly. Allocation rules and architectures are fixed, and retained outputs satisfy both time and iteration limits.

Profiling uses RTX~3090 GPUs, two CPU threads, one process per device, a three-second warm-up, and CUDA-synchronized timing. Baselines have $T_{\rm rec}=T_{\rm opt}$; DiffTR also includes basis estimation, operator setup, sampling, and refinement. Training and inference may use different GPUs of the same model. Loading, evaluation, and file I/O are excluded. Parameter counts cover network weights, including both EDIP-Net networks, but exclude optimized inputs, EMA copies, optimizer states, and buffers. Peak GPU memory is the largest allocated value per dataset; DiffTR uses the larger training or standalone-inference peak after releasing the optimizer. CUDA context memory, unused allocator reservations, and CPU SVD memory are excluded.

\end{document}